\pdfoutput=1
\documentclass[manuscript]{acmart}
\AtBeginDocument{%
  }

\usepackage{latexsym}
\usepackage{amsmath}
\usepackage{amsthm}
\usepackage{booktabs}
\usepackage{enumitem}
\usepackage{algorithm}
\usepackage[indLines=true,noEnd=false]{algpseudocodex}
\usepackage{color}
\usepackage{svg}
\usepackage{graphicx} %
\usepackage{caption}
\usepackage{subcaption}
\usepackage{multirow}
\usepackage{comment}
\usepackage{tablefootnote}
\usepackage[flushleft]{threeparttable}

\setcopyright{cc}
\acmJournal{TOMPECS}
\acmYear{2024} \acmVolume{1} \acmNumber{1} \acmArticle{1}
\acmMonth{1} \acmPrice{}\acmDOI{10.1145/3708984}

\begin{document}

\title{FedPref: Federated Learning Across Heterogeneous Multi-objective Preferences}

\author{Maria Hartmann}
\email{maria.hartmann@uni.lu}
\orcid{0000-0002-2179-3703}
\affiliation{%
  \institution{SnT, University of Luxembourg}
  \streetaddress{6, avenue de la Fonte}
  \city{Esch-sur-Alzette}
  \country{Luxembourg}
  \postcode{L-4364}
}

\author{Grégoire Danoy}
\email{gregoire.danoy@uni.lu}
\orcid{0000-0001-9419-4210}
\affiliation{%
  \institution{SnT, University of Luxembourg}
  \streetaddress{6, avenue de la Fonte}
  \city{Esch-sur-Alzette}
  \country{Luxembourg}
  \postcode{L-4364}
}
\affiliation{%
  \institution{FSTM/DCS, University of Luxembourg}
  \streetaddress{2, place de l’Université}
  \city{Esch-sur-Alzette}
  \country{Luxembourg}
  \postcode{L-4365}
}

\author{Pascal Bouvry}
\email{pascal.bouvry@uni.lu}
\orcid{0000-0001-9338-2834}
\affiliation{%
  \institution{FSTM/DCS, University of Luxembourg}
  \streetaddress{2, place de l’Université}
  \city{Esch-sur-Alzette}
  \country{Luxembourg}
  \postcode{L-4365}
}

\begin{abstract}
  The Federated Learning paradigm is a distributed machine learning strategy, developed for settings where training data is owned by distributed devices and cannot be shared with others. Federated Learning circumvents this constraint by carrying out model training in distribution, so that each participant, or client, trains a local model only on its own data. The parameters of these local models are shared intermittently among participants and aggregated to enhance model accuracy. 
This strategy has shown impressive success, and has been rapidly adopted by the industry in efforts to overcome confidentiality and resource constraints in model training. 

However, the application of FL to real-world settings brings additional challenges, many associated with heterogeneity between participants. Research into mitigating these difficulties in Federated Learning has largely focused on only two particular types of heterogeneity: the unbalanced distribution of training data, and differences in client resources. Yet many more types of heterogeneity exist, and some are becoming increasingly relevant as the capability of FL expands to cover more and more complex real-world problems, from the tuning of large language models to enabling machine learning on edge devices. In this work, we discuss a novel type of heterogeneity that is likely to become increasingly relevant in future applications: this is \textit{preference heterogeneity}, emerging when clients learn under multiple objectives, with different importance assigned to each objective on different clients. \\
In this work, we discuss the implications of this type of heterogeneity and propose a FedPref, a first algorithm designed to facilitate personalised federated learning in this setting. We demonstrate the effectiveness of the algorithm across several different problems, preference distributions and model architectures. In addition, we introduce a new analytical point of view, based on multi-objective metrics, for evaluating the performance of federated algorithms in this setting beyond the traditional client-focused metrics. We perform a second experimental analysis based in this view, and show that FedPref outperforms compared algorithms. 
\end{abstract}



\received{29 April 2024}

\maketitle

\section{Introduction}
Federated Learning is a distributed machine learning paradigm suited to problems where training data originates in distribution and cannot be shared. Barriers to the sharing of training data include, for example, privacy concerns, confidentiality, and technological constraints such as limited communication resources. Under the Federated Learning paradigm, learning is shifted to the distributed participants, with each participant training a separate local model only on the locally available dataset. The resulting local models are shared periodically with a central server or directly with other participants, where multiple such models are aggregated to enhance accuracy. This approach allows participants to implicitly share information contained in their training data without exposing the data itself.\\
This approach has demonstrated great potential and has already seen substantial uptake by the industry, e.g. for privacy-preserving healthcare applications \cite{Yuan2020AFL}, federated prompt tuning of large language models \cite{che-etal-2023-federated}, and on unmanned aerial vehicles \cite{Brik2020}; yet conceptual challenges remain. In particular, the different types of client heterogeneity are the subject of extensive study in the FL community, as heterogeneity is considered one of the main challenges standing in the way of realising the full potential of the FL paradigm \cite{kairouz_advances_2021}. To date, much of the focus in this field is devoted to mitigating the problem of data heterogeneity, where client datasets are not independent and data is distributed non-identically (non-iid) \cite{zhu_federated_2021}. The remaining focus is largely given to solving system heterogeneity%
, the type of heterogeneity that arises when clients have different functional capabilities, leading to challenges such as delayed convergence times, asynchronicity, imbalanced participation and reduced model quality \cite{abdelmoniem_comprehensive_2023}. This is also known as device heterogeneity in some parts of the literature.
\begin{figure}[t]
     \centering
     \begin{subfigure}[b]{0.3\columnwidth}
         \centering
         \includegraphics[width=\columnwidth]{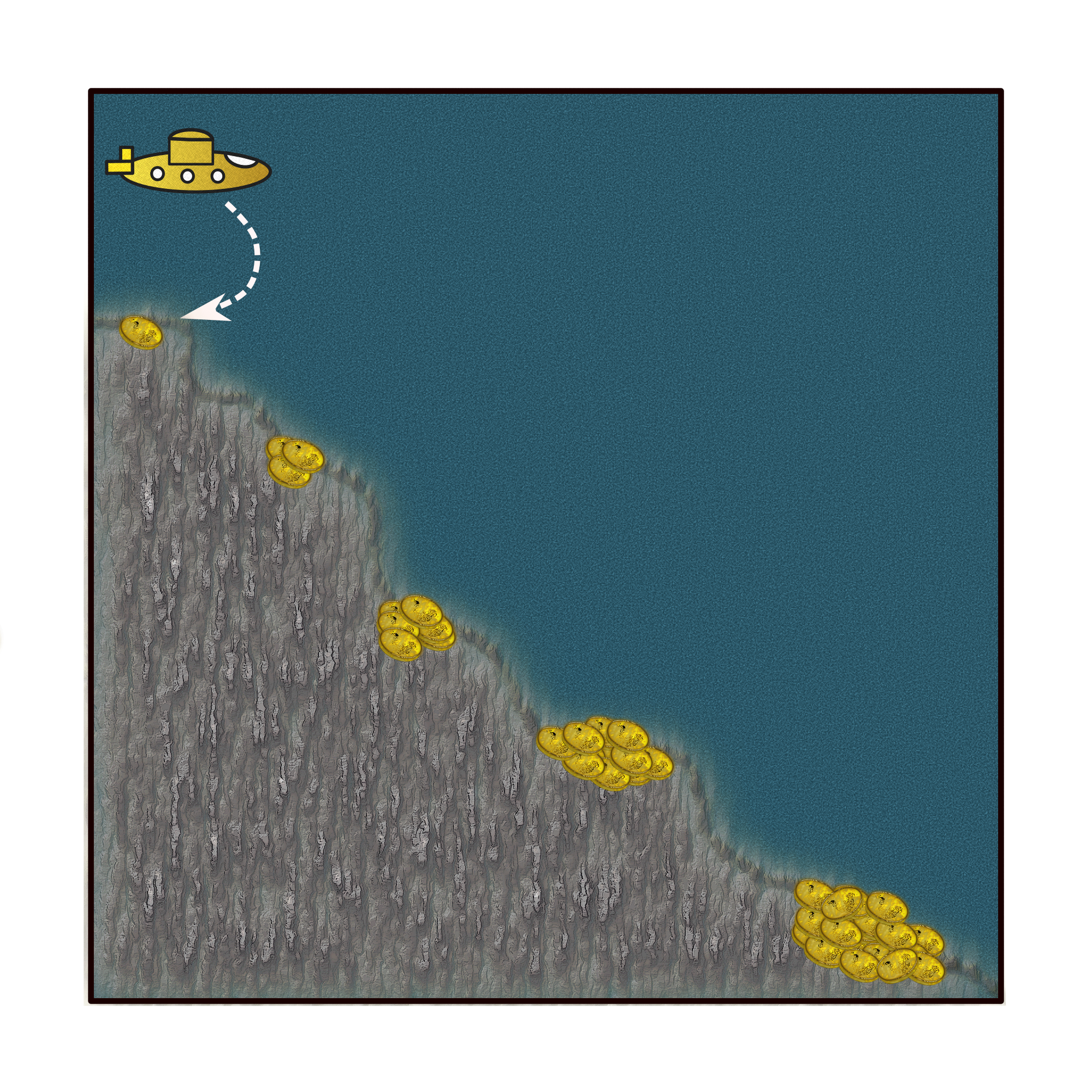}
     \end{subfigure}
     \hfill
     \begin{subfigure}[b]{0.3\columnwidth}
         \centering
         \includegraphics[width=\columnwidth]{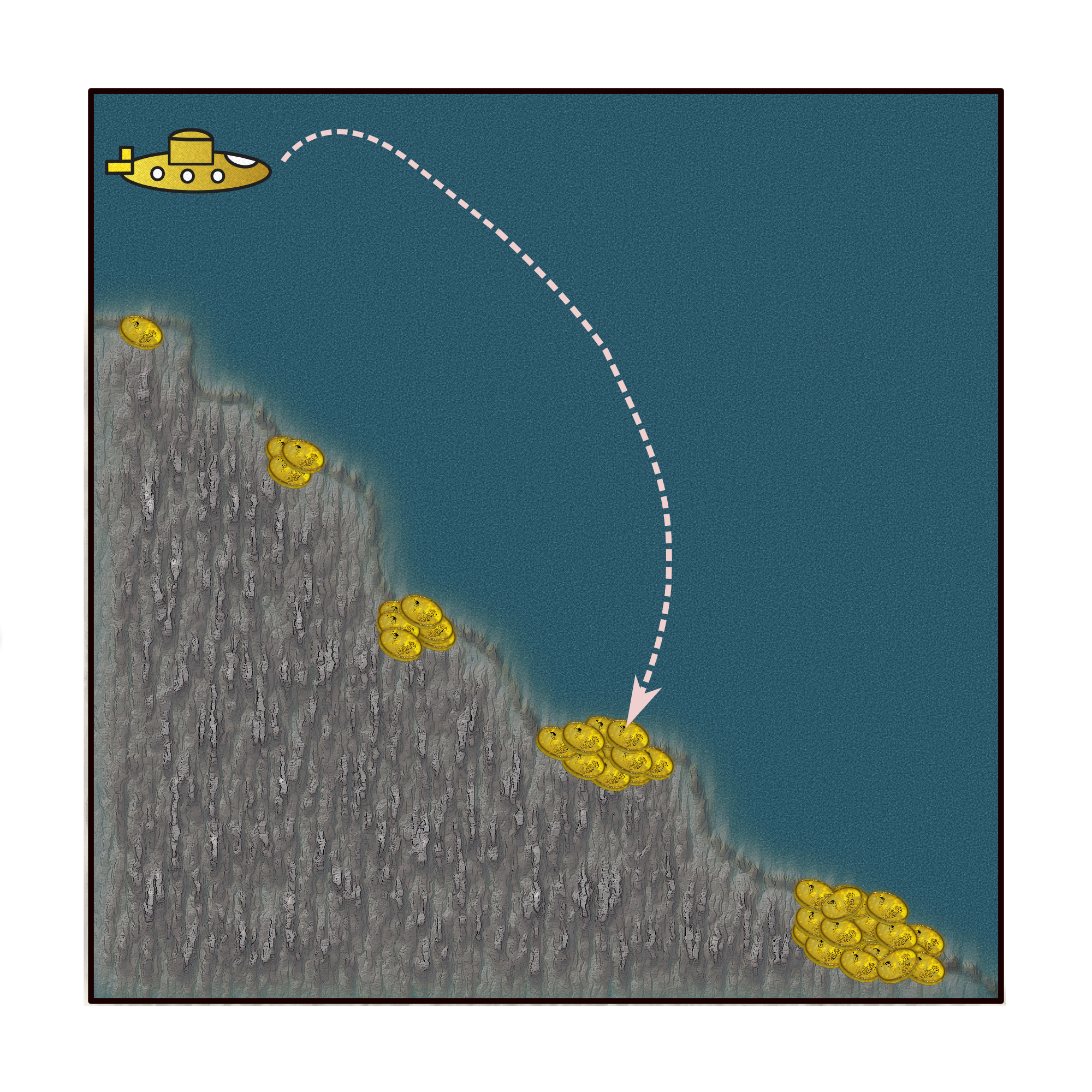}
     \end{subfigure}
     \hfill
     \begin{subfigure}[b]{0.3\columnwidth}
         \centering
         \includegraphics[width=\columnwidth]{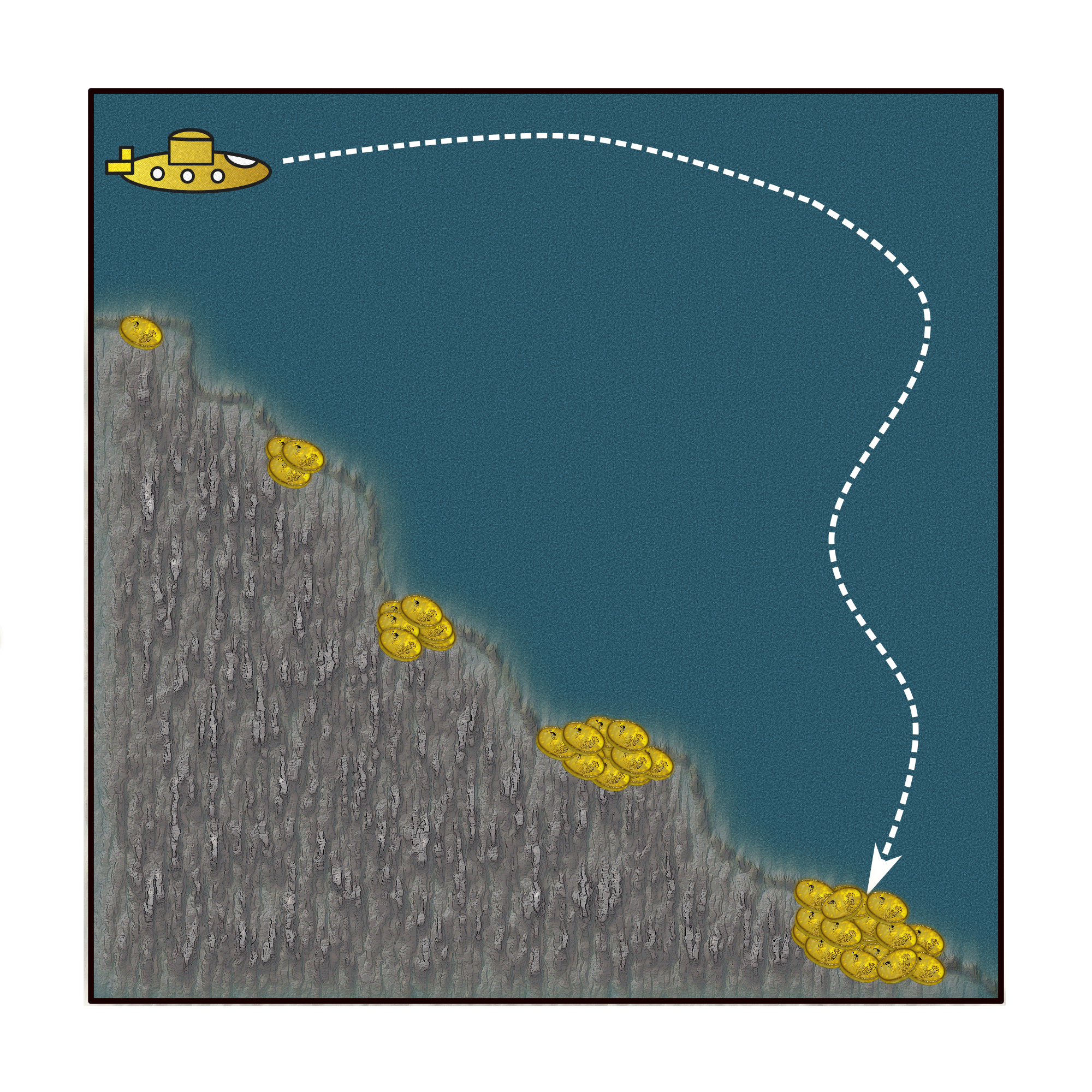}
     \end{subfigure}
        \caption{Different preferences lead to different solutions in a yellow submarine searching for underwater treasure. Left: A strong preference for minimising travel distance. Centre: Balanced preferences. Right: A strong preference for maximising the value of the treasure reward. The goal of our work is to allow clients with problems like these to perform FL effectively, despite their heterogeneous objective preferences.} 
        \Description{Three illustrations, each showing a yellow submarine diving for piles of underwater treasure of varying value distributed at different depths across the seabed. Left: The submarine dives for the closest available treasure, which has a low value. Centre: The submarine dives for a treasure of medium depth and medium value. Right: The submarine dives for the treasure of the greatest depth and value.}
        \label{fig:illustration-submarine}
        \vspace{2em}
\end{figure}
However, we argue that there is another type of heterogeneity that has so far received little attention, but is indeed highly relevant to many use case scenarios. This is the problem of \textit{(multi-)objective preference heterogeneity}, where clients have multiple objectives each, with different importance given to each objective on different clients.\\
Such problems arise naturally: many real-life problems inherently involve multiple underlying objectives, and often these objectives can be assigned different importance, or preferences, for different users or situations. Tuning large language models, managing the routing and charging of electrical vehicles, and planning the trajectories of unmanned aerial vehicles (UAVs) during search-and-rescue operations are just some examples of problems that could be modelled more accurately using multiple objectives.\\ 
However, performing federated aggregation with such clients can be challenging, as different objectives may lead to substantially different models, and objectives can be conflicting. The classical approach of training a single global model would likely struggle to satisfy each of these different preference distributions in the general case. Furthermore, the underlying idea behind modelling a problem with multiple objectives is often to allow the finding of multiple diverse solutions, with each representing a different trade-off between the various objectives. So a single global solution, even one delivering high objective values for all involved clients, would not necessarily be a satisfying solution in this scenario. For an intuitive example, consider the scenario illustrated in Figure~\ref{fig:illustration-submarine}. In this toy scenario, several submarines are searching for underwater treasure. Each has two separate objectives: minimising the diving distance, and maximising the haul of treasure recovered. We observe that these objectives are conflicting, as more valuable treasure is located deeper down on the seabed, necessitating a longer dive. In assigning different preference weights to the two objectives when solving this problem, we expect to recover different behaviours, as illustrated in the different panels of Figure~\ref{fig:illustration-submarine}. In the left-most image, the submarine has a high preference weight placed on the travel-distance objective, and so travels to the closest treasure. The right-most image shows the reverse: the submarine has a high preference for finding treasure, and so dives as far as necessary to reach the most valuable location. This is the learning outcome a user might expect in assigning preference weights; yet a non-personalised federated learning algorithm might instead converge to the same solution for all clients, shown in the central illustration. This represents a ``middle ground'' between the preference distributions of the two others, potentially leading to comparable scalarised results for both. However, the different preference weights have essentially lost their expected meaning: the user has no perceived control over the learning outcome.\\
To avoid this outcome, we propose FedPref, an algorithmic approach based on Personalised Federated Learning (PFL), where each federated client learns an individual model tailored to its needs, and different objective preferences lead to different solutions. The goal of our algorithm is twofold: first, to optimise individual client performance %
, as is common for other PFL approaches. However, we also want our algorithm to perform well under a multi-objective view of the federated system as a whole, conforming to the common expectations of multi-objective problem solving. In order to measure our success at this secondary goal, we introduce a novel analytical view of the federated system as a whole, using metrics common in the fields of multi-objective optimisation and multi-objective learning to assess the diversity and convergence of the set of solutions found by all clients.

Our contributions can be summarised as follows:

 \begin{itemize}
    \item We describe and formalise a new type of heterogeneity problem that occurs naturally in the federated setting.
    \item We propose a new algorithm to efficiently perform personalised Federated Learning in this setting, adaptively aggregating similar models based on a modified version of the cosine similarity metric. This algorithm does not require any knowledge about client preferences.
    \item We demonstrate the successful performance of our algorithm compared to several baselines from the state-of-the-art, and show that existing algorithms designed for data heterogeneity do not easily transfer to the preference-heterogeneous setting without loss of performance.
    \item We provide extensive additional validation experiments, studying the performance of the different components of our algorithm and the impact of different parameter choices.
    \item Finally, we introduce a novel analytical view of the federated system as a set of multi-objective solutions. We discuss the performance of our algorithm from a multi-objective point of view, exploring several multi-objective metrics and discussing the implications of these results.
\end{itemize}

\section{Related work}
To the best of our knowledge, there are no previous works in the literature addressing this continuous objective-heterogeneous setting in detail. However, several existing areas of research are related to our problem, most importantly in the fields of federated multi-objective learning, federated multi-task learning and mitigation of other types of heterogeneity in the federated setting. We will discuss the state of the art in these fields, insofar as it pertains to our problem setting, in the remainder of this section. 
\subsection{Federated multi-objective learning}
Two recent works \cite{yang_federated_2023}\cite{hartmann2023mofld} have addressed the federated multi-objective setting, where each client solves a problem with several objectives.
The former work \cite{yang_federated_2023} extends multi-gradient descent to the federated setting, showing that the algorithm can converge to a Pareto-stationary solution. This approach is said to tolerate the existence of different preference distributions across clients; however, clients ultimately train only a single global model, with no special regard for potentially conflicting objective distributions. To the best of our understanding, the objective-heterogeneous case is not examined in further detail.
In contrast, our algorithm trains a personalised model for each client, grouping clients with similar models for aggregation to exploit commonalities while separating clients whose models prove incompatible during the training process. Furthermore, our algorithm assumes a setting where client preferences remain private and are applied purely on the client side, mapping the multi-objective problem to a single-objective problem using a linear combination. The gradient update received by the server is computed with respect to this linearised objective, unlike in \cite{yang_federated_2023}.\\
The latter work \cite{hartmann2023mofld} considers the case where clients do not have fixed personalised preferences and instead collaborate under the direction of a central server to explore the space of objective preferences in search of different trade-off solutions. In our work, we assume -- arguably more realistically -- that each client does have preferences describing the individual importance of each objective to the client, and that the server has no knowledge of or control over these preferences. \\ %
In addition, we note that several other works in the literature, e.g.~\cite{hu_federated_2022}, \cite{zhu_multi-objective_2020}, \cite{mehrabi_towards_2022} , apply multi-objective methods to Federated Learning problems. These works are not closely related to the setting discussed here, as they apply multi-objective methods to aspects of the federated learning system itself rather than solving inherently multi-objective problems on the clients.

 \subsection{Federated multi-task learning}\label{subsec:related-work-multi-task}
We observe that the notion of Federated Multi-Task Learning (FMTL) appears to be not yet well-defined in the Federated Learning community; different definitions are given or implied in different works in the literature. This likely stems from the fact that the underlying idea of multi-task learning -- using commonalities between different but related learning tasks to enhance model training for each task \cite{zhang_overview_2017} -- can be transferred to the federated learning setting in different ways:
In some works, the label is applied exclusively to the strategy of using personalised federated learning to mitigate data heterogeneity \cite{Sattler2019ClusteredFL}, or other types of heterogeneity as well \cite{smith_federated_2017}, across clients. 
In \cite{yang_federated_2023}, the scenario where each federated client is solving a multi-objective problem is considered as a variant of FMTL, with each objective corresponding to a task. In other works, e.g.~\cite{cai_many_2023}\cite{ghosh_efficient_2020}\cite{huang_federated_2023}, the multi-task label is used for a scenario where federated clients are solving different learning problems (i.e.,~each client has a single objective, and objectives may differ across clients).

The common description of all these variants as federated multi-task learning, or multi-task federated learning, could be reconciled by defining FMTL as a broad federated learning strategy training individual client models to handle any type of heterogeneity, including data heterogeneity, hardware heterogeneity, objective heterogeneity and preference heterogeneity. However, this description hardly seems useful in its generality, and indeed shares a significant overlap with the strategy commonly labelled as Personalised Federated Learning (PFL) -- see the following section for a further discussion of PFL. In the remainder of this work, we refer to FMTL only in the context of single-objective heterogeneity.\\
 This is arguably an edge case of our problem, recoverable from the general formulation (where each task corresponds to one objective) by assigning a preference of zero to all but one objective. Most of the works addressing this variant, e.g. \cite{ghosh_efficient_2020}, focus on attempting to cluster clients that solve the same task, excluding others from aggregation. Others propose methods that allow for the aggregation of models trained for different tasks; for example, Cai et al.~\cite{cai_many_2023} propose to perform weighted aggregation across clients based on a model similarity metric.

 \subsection{Other types of federated heterogeneity}
Our problem is also related to other types of heterogeneity problems that present themselves in the FL setting, where the FL algorithm must account for differences between clients, such as data heterogeneity or hardware heterogeneity. Particularly in the case of data heterogeneity, client models also tend to develop in different directions -- as is to be expected for clients in our objective-heterogeneous setting -- making the comparison with our problem setting an interesting one. Many varied approaches have been proposed to address this problem \cite{ye_heterogeneous_2023}; these can be broadly divided by their approach to model aggregation \cite{tan_towards_2023}.\\
Some works follow the more classical approach of producing a single generalised global model, with the goal of adapting this model as well as possible to all individual client datasets simultaneously. One of the first such algorithms was the FedProx framework \cite{li_federated_2018}, which relies on regularisation to encourage model adaptation. To accomplish this, a new proximal term is added to the loss function of each client, penalising divergence of the local model from the global model. Other regularisation-based algorithms have since followed, e.g.~\cite{karimireddy_scaffold_2020} and ~\cite{li_model_2021}, introducing variance reduction and model-contrastive learning, respectively. 

In contrast, the goal of the federated aggregation in the second approach is to learn an individual model tailored to each client \cite{tan_towards_2023}. This strategy is known as Personalised Federated Learning (PFL).
Variants of PFL, in turn, may be separated into those based on a modified model architecture, such as parameter decoupling or knowledge distillation, and those based on model similarity-guided aggregation. Of the former approaches, knowledge distillation strategies can be costly, and the parameter-decoupling strategy may struggle to adapt to training with different objective functions. Therefore, we choose to place our focus on the latter approaches, as model similarity-based methods may be light-weight, appear to have the potential to adapt well to different types of heterogeneity, and require no additional information about clients. A number of such approaches have been proposed in the literature in recent years, e.g.~\cite{Long2022}, \cite{ghosh_efficient_2020}, \cite{Duan2021}.
In this work, we focus on two recent methods that appear most flexible, and so most likely to transfer well to the preference heterogeneous setting: the Clustered Federated Learning \cite{Sattler2019ClusteredFL} (CFL) algorithm, and Many-Task Federated Learning \cite{cai_many_2023} (MaTFL) -- the latter has already been discussed in Section~\ref{subsec:related-work-multi-task}. %
The former work, proposing the Clustered FL [12] (CFL) algorithm, is of particular interest here. It deals with settings where the underlying data distributions known to participants are not fully compatible, leading to conflicts in the training of a joint model. To solve this, the idea of CFL is to train clients together in a classical federation until the global model converges to a stationary point, allowing clients to learn from each other until mutual conflicts stall the training process. Then clients are permanently separated into clusters based on the similarity of model gradients in the stationary point. Our multi-objective preference-heterogeneous setting is related to the data-incongruity problem tackled by CFL,
in that we expect clients with preferences for conflicting objectives to also produce incompatible models during training. However, the heterogeneity of clients shall be more complex, given the number of potential objectives and different preference distributions.
Therefore, we take inspiration from the clustering strategy of CFL for our approach, but additionally introduce the idea of personalising learning inside each cluster. Our aim is to allow a higher degree of individual exploration for clients at an earlier stage in the training,
without cutting off cooperation earlier than necessary.

\section{The FedPref algorithm}
In this section, we define the FedPref algorithm and relevant concepts. We begin by formally defining the problem setting in Section~\ref{subsec:theory-problem}, followed by an initial sketch of the algorithm and a definition of the underlying similarity metric in Section~\ref{subsec:theory-sketch-definitions}. Finally, we discuss the components of the algorithm in more detail: the weighted aggregation strategy is described in Section~\ref{subsec:theory-weighted-aggregation}, the clustering strategy in Section~\ref{subsec:theory-clustering}, and the full FedPref algorithm in Section~\ref{subsec:theory-full-algorithm}.

\subsection{Problem formulation}\label{subsec:theory-problem}
\begin{figure}
    \centering
    \includegraphics[width=.75\columnwidth]{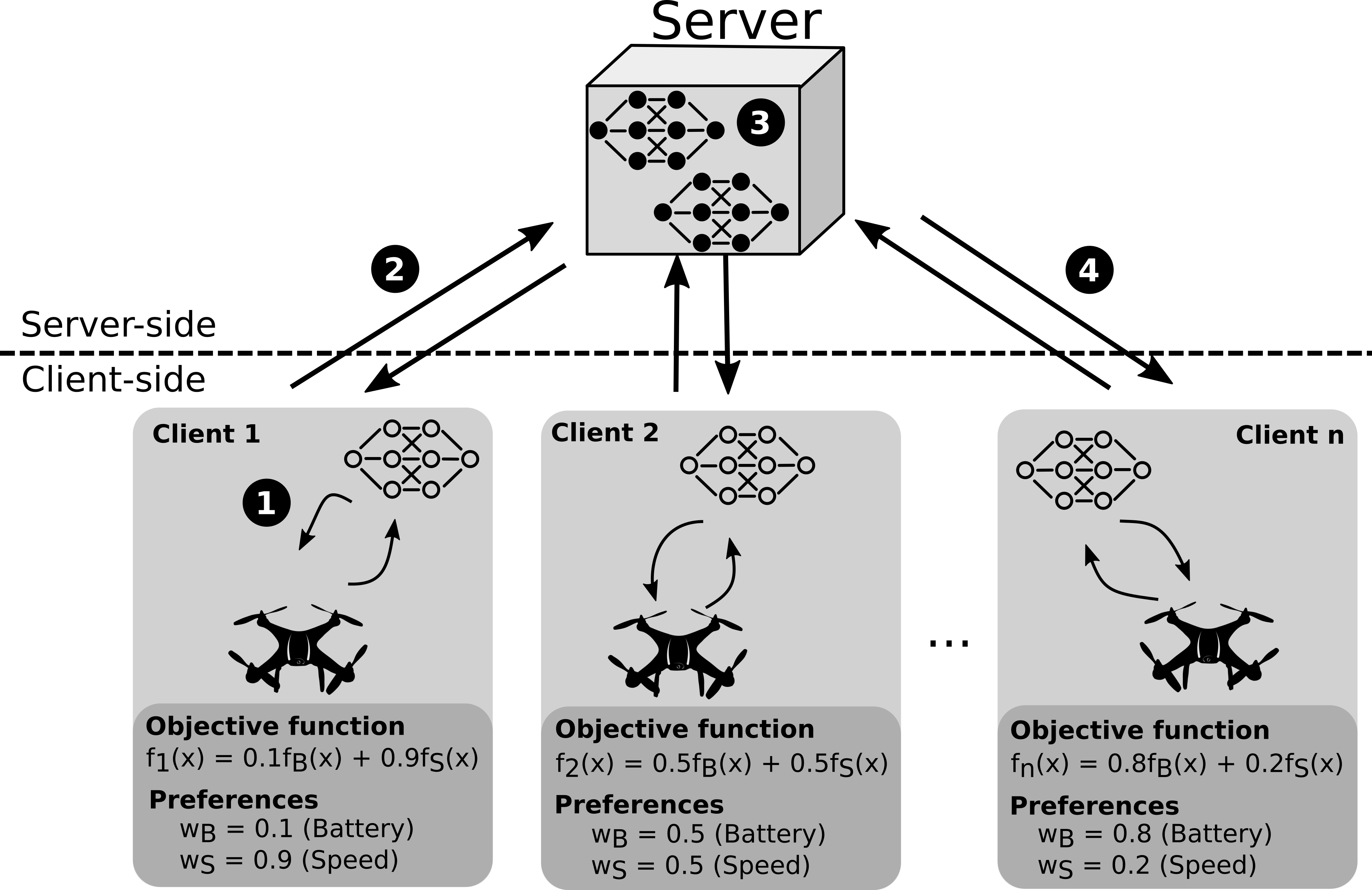}
    \caption{An illustration of the federated system solving a multi-objective problem. In this instance, we want to learn to plan trajectories for drones, under two potentially conflicting objectives: conserving energy and maximising speed. Each drone assigns different importance (preference weights) to these objectives. Federated Learning takes place as follows: (1) Clients (drones) perform local training, using the objective function defined by their preferences. (2) Clients submit model updates to the server. (3) The server aggregates these model updates, obtaining personalised models. (4) The server returns the respective personalised models to the clients.}
    \label{fig:fl-mo-problem}
    \Description{See caption.}
\end{figure}
We want to perform personalised Federated Learning across $n$ clients, each of which has a learning problem with $m$ distinct objectives $f_1,\cdots ,f_m$. There is no general importance order assigned between objectives, but each client has a personal fixed preference weight vector across all objectives. See also Figure~\ref{fig:fl-mo-problem} for an illustration of the problem and the federated learning process.\\
Following a classical approach in multi-objective optimisation\cite{sharma_comprehensive_2022}, we map this multi-objective problem to a single-objective problem in order to solve it, so that all clients learn a linear combination of these same objectives, with the preference weights assigned as scalars. So client $i$, with preference distribution $\vec{w^i}=(w^i_1,\dots, w^i_m)^T$, is optimising the objective function \begin{equation}
f^i(\theta)=f(\vec{w^i}, \theta) = \vec{w^i}\vec{f}(\theta) = \sum_{j}^m w^i_j f_j(\theta).
\end{equation}
The preference distribution of each client is unknown to all other participants, including the federated server. (We can assume without loss of generality that all single-objective components $f_j$ are known to all clients.) Each client $i$ trains a \textit{personalised} model $\theta_i$ using its personal preference weights. A major challenge in this scenario is that the objectives of federated clients may conflict, and these conflicts can lead to the divergence of client models at any stage of the training.

\subsection{Concept sketch and definitions}\label{subsec:theory-sketch-definitions}
\begin{figure}
    \centering
    \includegraphics[width=.5\columnwidth]{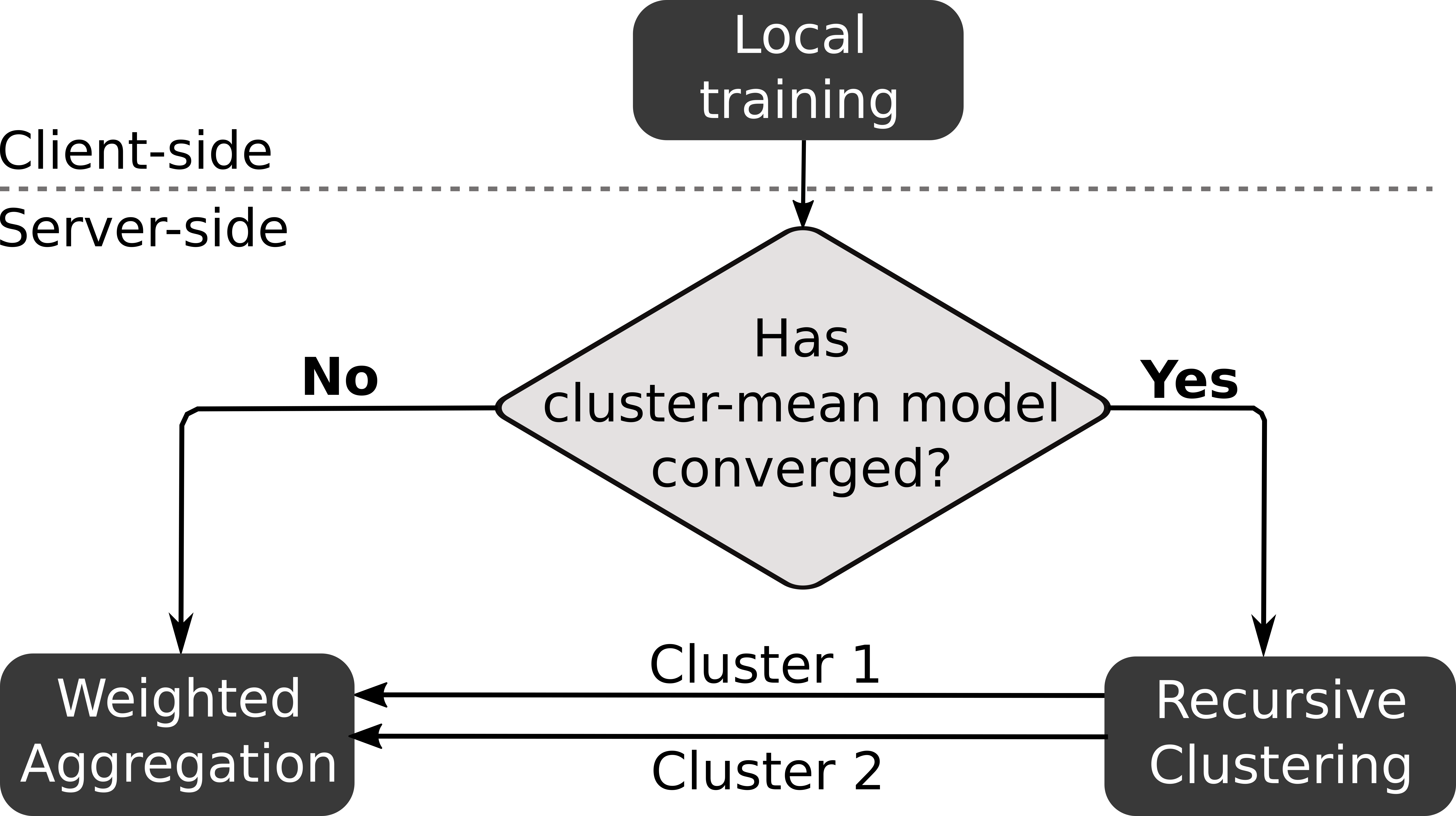}
    \caption{A schematic representation of the flow between components of the algorithm.}
    \label{fig:alg-schematic}
    \Description{Following local training on the client-side, clustering is performed only if the cluster-mean model has converged. If clustering is performed, it is followed by weighted aggregation on both clusters; otherwise, the cluster remains unchanged and only weighted aggregation is performed.}
\end{figure}
The fundamental idea of FedPref is to combine a recursive clustering mechanism, similar to \cite{Sattler2019ClusteredFL}, and an adaptive weighted aggregation scheme, both based on a model similarity metric. The underlying idea behind this combination is to enable effective grouping and aggregation of clients whose preferences are compatible during the learning process (provided by the clustering component), while also maintaining the flexibility of training a personalised model for each client using weighted aggregation. Compare Figure~\ref{fig:alg-schematic} for a visual representation of the flow between these components. Initially, all participating clients are grouped together in a single cluster. During every aggregation step, a personalised model is computed for each client, using adaptive weights computed based on mutual model similarity between pairs of clients. The mean model of all clients in the cluster serves as an indicator of the success of the intra-cluster collaboration: the mean model converges if either all clients converge, or if the gradients of personalised client models start developing in conflicting directions. In this case, we perform a recursive clustering step, splitting the current cluster in two based on the same mutual model similarity metric that is used for the weighted aggregation. The learning process is then continued in the same manner inside the new clusters.\\
\subsubsection{Similarity metric}
Before discussing the functionality of each component in detail in the following sections, we shall formally introduce the modified similarity metric that underpins both components. The similarity metric in aggregation round $t$ is computed on the basis of model updates \begin{equation}
    \Delta\theta_i = \theta_i-\bar{\theta}^{t-1}_C,
\end{equation}
where $\bar{\theta}^{t-1}_C$ is the cluster-mean model obtained after the previous aggregation step. Using these gradients, we define the similarity metric $sim(\cdot, \cdot)$ of two models $\theta_i$ and $\theta_j$ as
\begin{equation} \label{eq:similarity-metric}
    sim(\Delta\theta_i, \Delta\theta_j) = \frac{1}{L}\sum^L_{\ell}cossim(topR(\Delta\theta_i^\ell), topR(\Delta\theta_j^\ell)),
\end{equation}
where $\Delta\theta_i^{\ell}$ is the $\ell$-th layer of $\Delta\theta_i$ and $L$ is the total number of layers per model. The $topR$ operator is a variant of $topk$, where $k$ is determined by the dimension of the input vector and a ratio $R\in (0,1]$. $TopR$ maps a vector $\vec{v}$ to a vector of the same dimension where the top $k=\lceil dim(\vec{v})\cdot R\rceil$ elements of $\vec{v}$ (in absolute terms) are retained and the remaining elements set to zero. So for $topR(\vec{v}) = \vec{u}$, we have \begin{equation}\label{eq:topr}
    u_i = \begin{cases}
    v_i, & \text{if $|v_i|$ in top $\lceil R\cdot dim(\vec{v})\rceil$ absolute elements of $\vec{v}$}.\\ %
    0, & \text{otherwise}.
  \end{cases}
\end{equation}
The cosine similarity $cossim(\cdot, \cdot)$ is defined in the standard way: \begin{equation}\label{eq:cosine-similarity}
    cossim(\vec{u}, \vec{v}) = \frac{\langle\vec{u},\vec{v}\rangle}{\lVert\vec{u}\rVert\cdot\lVert\vec{v}\rVert}.
\end{equation}
\begin{figure}
    \centering
    \begin{subfigure}[b]{.3\columnwidth}
        \centering
        \includegraphics[width=.6\columnwidth]{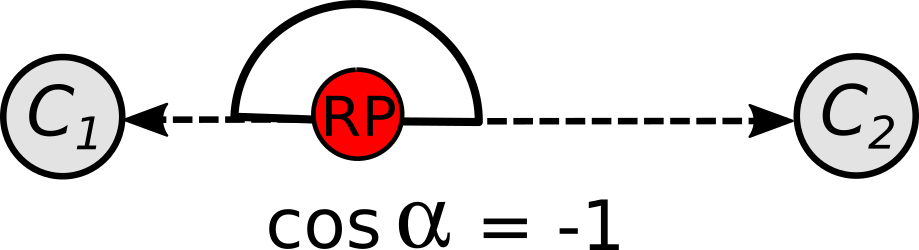}
    \end{subfigure}
    \begin{subfigure}[b]{.3\columnwidth}
        \centering
        \includegraphics[width=.3\columnwidth]{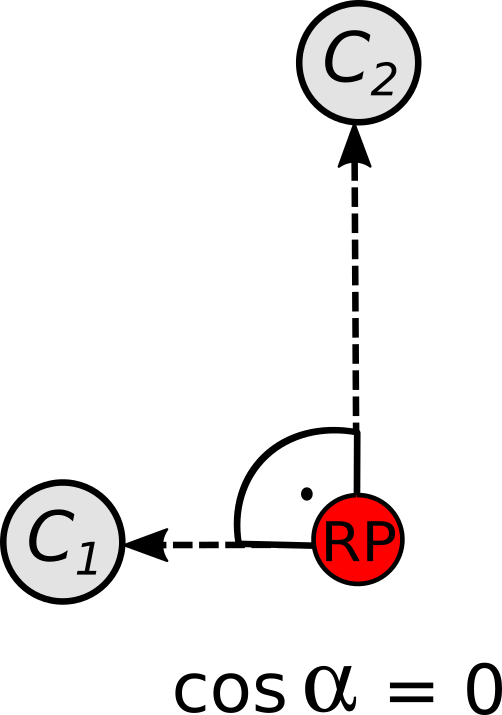}
    \end{subfigure}
    \begin{subfigure}[b]{.3\columnwidth}
        \centering
        \includegraphics[width=.3\columnwidth]{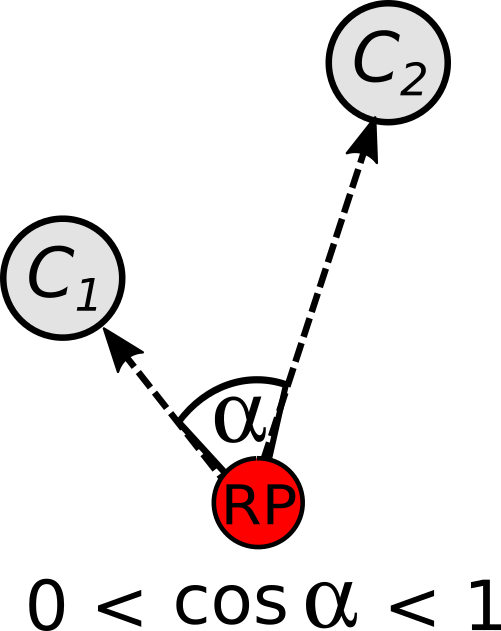}
    \end{subfigure}
    \caption{Geometric interpretation of cosine similarity.}
    \label{fig:cosine-similarity}
    \Description{See text.}
\end{figure}

The rationale for this modification lies in the geometric interpretation of the cosine similarity metric (illustrated in Figure~\ref{fig:cosine-similarity}): recall that the plain cosine similarity describes the cosine of the angle between two given vectors, with a value of $-1$ equivalent to antiparallel vectors, a value of $0$ denoting orthogonal vectors and the maximum value of $1$ denoting parallel vectors. When applied to model gradient updates, this metric can then describe -- quite intuitively -- how similarly two models are developing. This insight is leveraged e.g.~in the CFL algorithm \cite{Sattler2019ClusteredFL}.\\ However, we note that complications can arise from this application of the plain metric to model updates, particularly relating to the potentially high dimensions of models and the choice of reference point. The former point rests on the observation that in high-dimensional spaces, the cosine similarity metric is affected by the ``curse of dimensionality'', rendering comparisons of high-dimensional vectors in dense spaces, such as the weights of a neural network, increasingly difficult. In other works, e.g.~\cite{Sattler2019ClusteredFL}\cite{cai_many_2023}, this is mitigated to an extent by comparing the individual layers of models instead of the complete flattened model. However, with the trend towards larger and larger models, we attempt to find a more general solution by introducing the `topR' filtering of layer-gradients. The intention of this step is to sparsify the space in which vectors are compared, in the hope of obtaining more meaningful results.\\
The latter complication is founded on the need for a reference point in defining the gradient vectors to be compared. As we want to train personalised models, including for clients inside the same cluster, models do not begin each local training round with a common model (as would be the case in e.g. FedAvg, or CFL). Therefore, we need to explicitly define a model to compare to, ideally one that is both close to each client's actual model and whose difference accurately represents the relation of the clients being compared. We choose the cluster-mean model, obtained after the aggregation of the previous round has concluded, as this reference point for our algorithm. %

To recapitulate, we choose to use this modified metric instead of the more common direct applications of cosine similarity for two main reasons: \begin{itemize}
    \item We hope to mitigate the ``curse of dimensionality'' that makes this metric increasingly meaningless for larger vector dimensions.
    \item Selecting the subset of the largest weights for each layer allows us to compare the most impactful, or ``important'' aspects of the models. This could lead to more meaningful decisions about which models to aggregate together.
\end{itemize}
We will show in Section~\ref{sec:evaluation} that, when compared to the use of the pure cosine similarity metric without weight selection, the use of this metric does indeed lead to improved results in our validation experiments.

\subsection{Weighted aggregation}\label{subsec:theory-weighted-aggregation}
\begin{figure}
    \centering
    \begin{subfigure}[b]{0.45\columnwidth}
    \centering
        \includegraphics[width=.5\columnwidth]{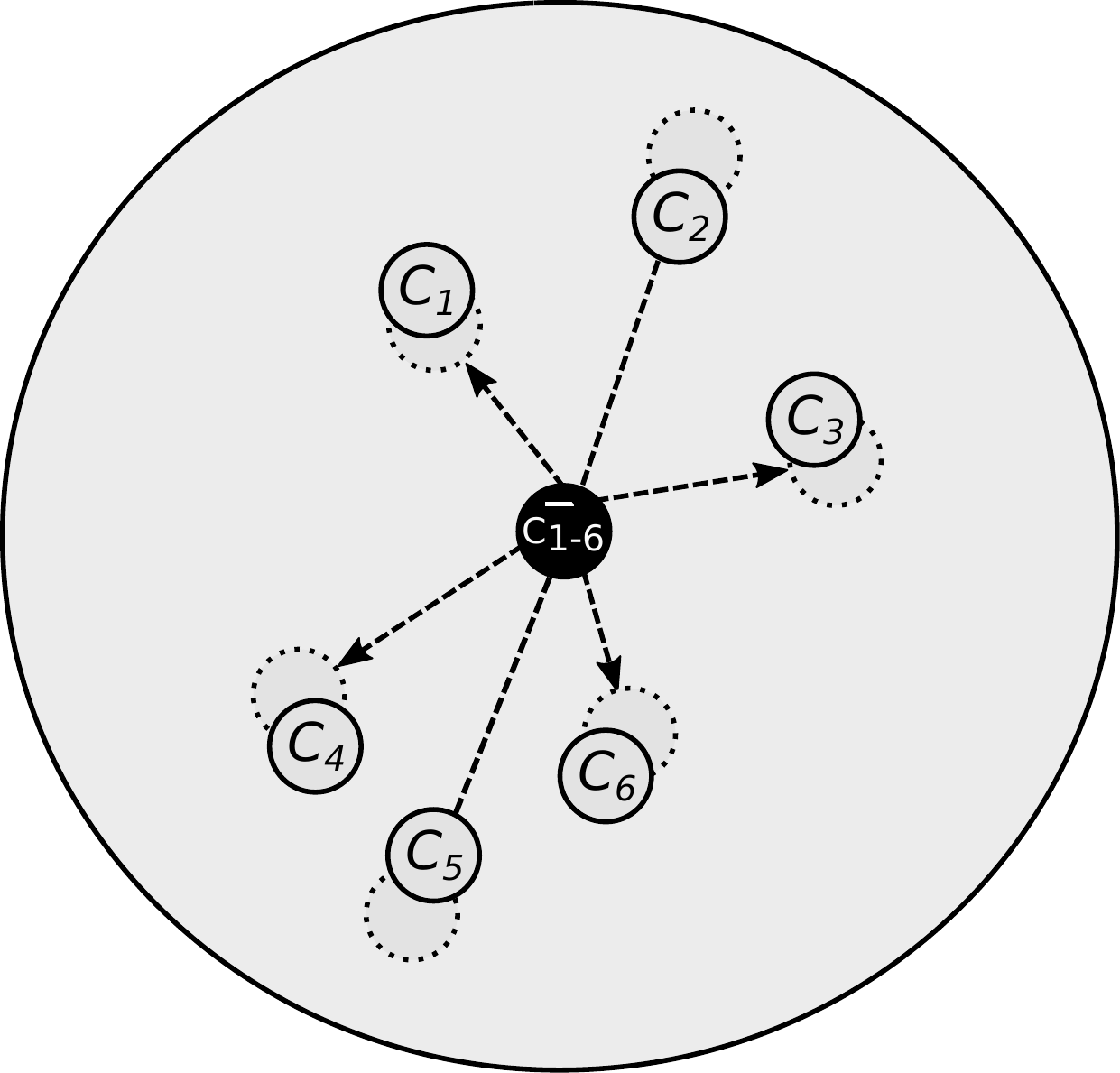}
    \end{subfigure}
    \hfill
    \begin{subfigure}[b]{0.45\columnwidth}
    \centering
        \includegraphics[width=.5\columnwidth]{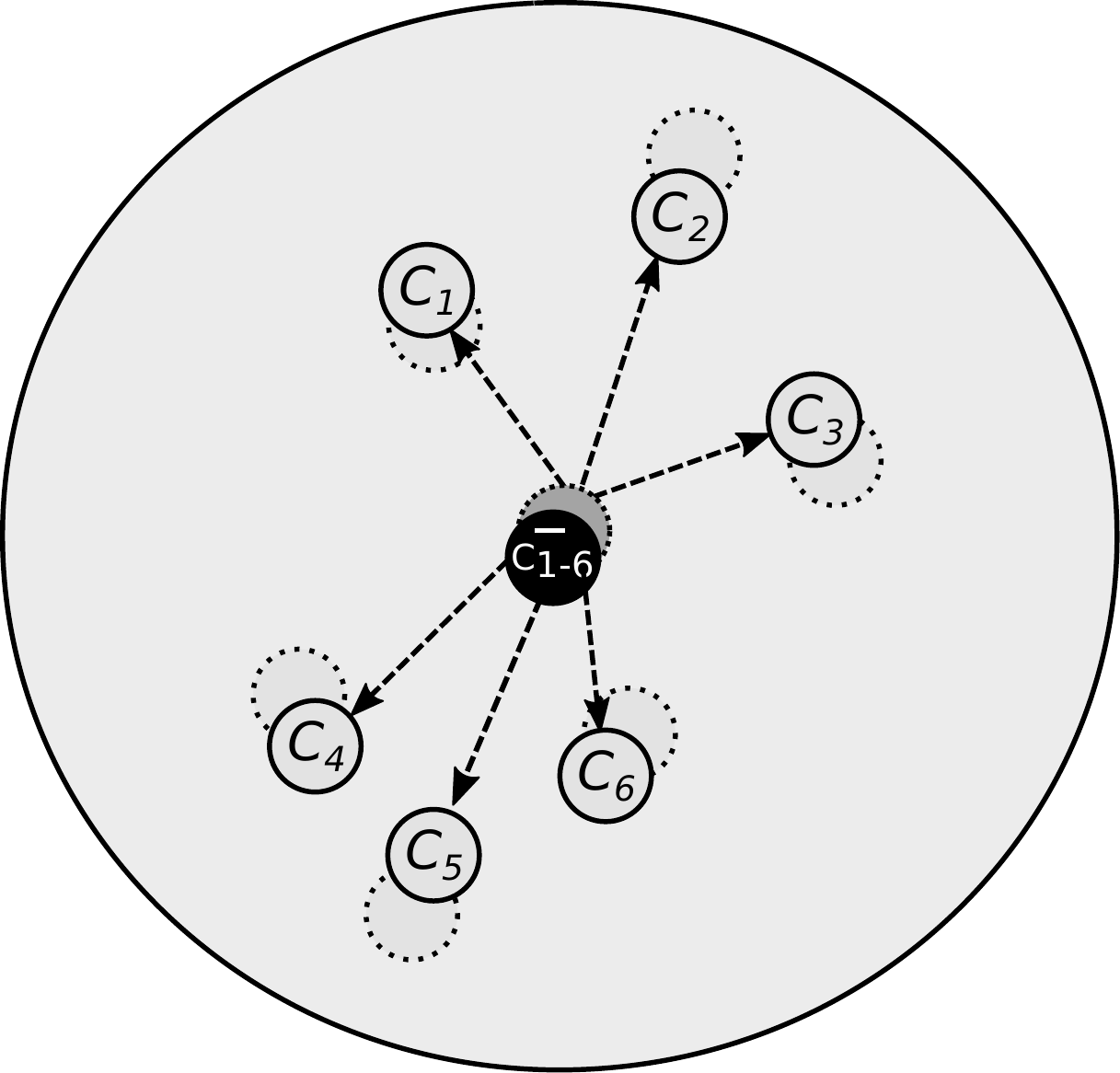}
    \end{subfigure}
    \caption{A weighted aggregation step inside a single cluster. Left: personalised client updates are computed using aggregation weights based on client similarity relative to the cluster-mean. Right: The updated cluster-mean is computed.}
    \label{fig:illustration-weighted-aggregation}
    \Description{First, client models inside the cluster are recomputed, but the cluster-mean reference model remains unchanged. Then the cluster-mean reference model is recomputed based on the updated client models.}
\end{figure}
The weighted aggregation -- described in the pseudocode in Algorithm~\ref{alg:weighted-aggregation} and illustrated in Figure~\ref{fig:illustration-weighted-aggregation} -- is carried out by the server for each separate cluster. For each cluster, the weighted aggregation phase begins with computing the similarity matrix of all clients contained in the cluster. The similarity metric (defined in Equation~\ref{eq:similarity-metric}) returns a value between $-1$, representing the lowest possible mutual similarity, and $+1$, representing the highest possible similarity. These values are then clipped to a minimum lower similarity bound $s_{min}$ -- given to the algorithm as a parameter during initialisation -- and subsequently normalised to the range $[0,1]$ (see line $5$ in Algorithm~\ref{alg:weighted-aggregation}). This step can be used to enforce a minimum similarity required for aggregation, as it essentially excludes all clients whose similarity to another client is lower than the given threshold from the aggregation with that client. Following this precomputing of similarity values, the actual personalised aggregation takes place: to compute the new personalised model for each client, the row corresponding to this client in the similarity matrix is taken as aggregation weight vector, normalised once more so that the sum of weights adds up to one, and finally used to compute the weighted average of all client models -- see lines $9$ and $10$ in Algorithm~\ref{alg:weighted-aggregation}. This aggregation is carried out for each client inside the cluster; then the resulting personalised models are returned to the respective clients.

\begin{algorithm}
    \caption{Weighted aggregation}\label{alg:weighted-aggregation}
    \begin{algorithmic}[1]
    \State $C$ list of $c$ clients in cluster
    \State $\mathcal{W} \gets (0)^{c\times c}$  \Comment{Init aggregation-weight matrix}
    \For{$i\in C$}
        \For{$j\in C$}
        \State $w_{ij}\gets (sim(\Delta\theta_i,\Delta\theta_j)-s_{min})/(1-s_{min})$
        \EndFor
    \EndFor
    \For{$i\in C$}
        \State $\hat{w}_i \gets w_i/\lvert w_i\rvert$
        \State $\theta_i \gets \sum_{c\in C}\hat{w}_{ic}\theta_c $
    \EndFor
    \State \Return $(\theta_c| c\in C)$
    \end{algorithmic}
\end{algorithm}

\subsection{Recursive clustering}\label{subsec:theory-clustering}
\begin{figure}
    \centering
    \includegraphics[width=\columnwidth]{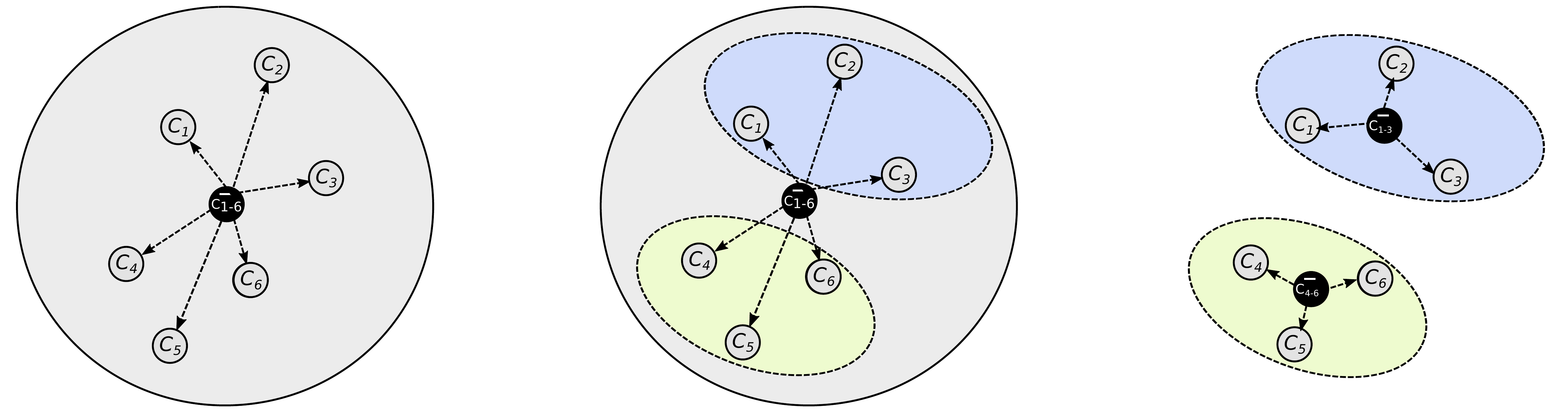}
    \caption{A recursive clustering step. Left: A cluster with cluster-mean at the beginning of an aggregation round. Centre: Clients inside the cluster are split into two clusters based on pairwise similarity relative to cluster-mean. Right: The resulting two clusters with respective cluster-means.}
    \label{fig:illustration-clustering}
    \Description{See caption.}
\end{figure}
The clustering procedure, illustrated in Figure~\ref{fig:illustration-clustering}, is performed whenever a cluster is found to have converged during an aggregation round, i.e.~where the clients inside the cluster no longer benefit from federated collaboration -- the exact convergence criterion is discussed in Section~\ref{subsec:theory-full-algorithm}. The purpose of this procedure is to separate the clients contained in the cluster into two new sub-clusters in such a way that clients whose models are developing similarly are grouped together to continue learning from each other, and clients that are developing in different directions are separated. This bipartitioning is based on the same similarity metric (Equation~\ref{eq:similarity-metric}) as is used for the weighted aggregation. In principle, any suitable clustering algorithm can be utilised to perform the clustering itself; in this work, we choose to use spectral clustering \cite{damle_simple_2018}, as it tends to produce well-balanced clusters, performs well for low numbers of clusters, and an implementation is readily available in common libraries. The clustering procedure is performed no more than once per cluster per aggregation round.

\begin{algorithm}
\caption{Clustering}\label{alg:clustering}
\begin{algorithmic}[1]
\Require $\lVert\Delta\bar{\theta}_C\rVert \leq \varepsilon$ \Comment{Cluster-avg model change $\leq \varepsilon$}
\State $C$ list of $c$ clients in cluster
\State $\mathcal{S} \gets (0)^{c\times c}$  \Comment{Init similarity matrix}
\For{$i\in C$}
\For{$j\in C$}
\State $\Delta \theta_i, \Delta \theta_j \gets \bar{\theta}^{t-1}_C-\theta_i, \bar{\theta}^{t-1}_C-\theta_j$
\State $\mathcal{S}_{ij}\gets sim(\Delta\theta_i,\Delta\theta_j$)
\EndFor
\EndFor
\State $C_1, C_2 \gets SpectralClustering(C, \mathcal{S},2)$ \Comment{Bipartition $\mathcal{C}$}
\State \Return $C_1, C_2$
\end{algorithmic}
\end{algorithm}

\subsection{Full algorithm}\label{subsec:theory-full-algorithm}
The complete algorithm combines the weighted aggregation and clustering components, as detailed in Algorithm~\ref{alg:full-algorithm} and conceptually in Figure~\ref{fig:alg-schematic}. In every round, all local models are trained for a fixed number of steps. Once all models for a given cluster $C$ have been reported to the server, the aggregation phase begins. As a first step, the clustering criterion is checked: the  difference of the cluster-mean model $\Delta\bar{\theta}_C$ of the most recent local updates to the cluster-mean model $\bar{\theta}_C^{t-1}$ following the latest aggregation round is computed (see lines $7-8$ in Algorithm~\ref{alg:full-algorithm}). If the magnitude of this change is less than a given convergence threshold $\epsilon$, we assume that the models of clients inside the cluster are diverging. We therefore trigger the clustering process to bipartition the current cluster $C$ into two new clusters $C_1$ and $C_2$. We then carry out weighted aggregation according to Algorithm~\ref{alg:weighted-aggregation} on the new clusters, before updating the server-side record of current clusters.\\
If the clustering criterion is not met, aggregation continues in the preexisting cluster: weighted aggregation is carried out in this cluster, and client-membership of this cluster is recorded unchanged.\\
In one full server-side aggregation step, this procedure is executed for every cluster, with personal aggregated models returned to the clients of each cluster after aggregation has concluded. The algorithm terminates after $T$ such aggregation rounds. Note that even if clients are still part of a larger cluster after $T-1$ aggregation rounds, the last aggregation step can be skipped after the final local training round, to allow clients a degree of local fine-tuning (see line $23$ in the algorithm). We call this fine-tuning variant of the algorithm FedPref+FT, and the version without a fine-tuning step FedPref-FT.

\begin{algorithm}
    \caption{NewFL-Server}\label{alg:full-algorithm}
    \begin{algorithmic}[1]
        \State $\mathcal{C} \gets \{[1,\dots , n]\}$ \Comment{Initial cluster}
        \State $\theta^0_1,\dots , \theta^0_n \gets $ Initialise client models
        \For{$t \in 1, \dots , T-1$}
        \State $\theta'_1,\dots , \theta'_n \gets $ Train local models %
        \For{$C \in \mathcal{C}$}
            \State $\mathcal{C}_{temp}\gets \{\}$
            \State $\bar{\theta}^{t-1}_C \gets 1/\lvert C\rvert \sum_{c\in C}\theta_{c}^{t-1}$
            \State $\Delta\bar{\theta}_C \gets \bar{\theta}^{t-1}_C - 1/\lvert C\rvert\sum_{c\in C}\theta'_c$
            \If{$\lVert\Delta\bar{\theta}_C\rVert \leq \varepsilon$} \Comment{Cluster converged}
                \State $C_1, C_2 \gets Clustering(\bar{\theta}^{t-1}_C, [\theta'_c | c\in C]) $
                \State $\theta'_{C_1}, \theta'_{C_2} \gets \{\theta^t_c|c\in C_1\}, \{\theta^t_c|c\in C_2\}$
                
                \State $\theta^t_{C_1}\gets WeightedAggregation(\bar{\theta}^{t-1}_{C_1}, \theta'_{C_1})$
                \State $\theta^t_{C_2}\gets WeightedAggregation(\bar{\theta}^{t-1}_{C_2}, \theta'_{C_2})$
                \State $\mathcal{C}_{temp}\gets C_{temp}\cup \{C_1, C_2\}$
            \Else
                \State $\theta'_C\gets \{\theta'_c | c \in C \}$
                \State $\theta^t_C \gets WeightedAggregation(\bar{\theta}^{t-1}_{C}, \theta'_C)$
                \State $\mathcal{C}_{temp}\gets \mathcal{C}_{temp}\cup \{\mathcal{C}\}$
            \EndIf
        \EndFor
        \State $\mathcal{C}\gets \mathcal{C}_{temp}$
        \EndFor
        \State $\theta'_1,\dots , \theta'_n \gets $ Train local models \Comment{Optional fine-tuning, replacing last aggregation round}
    \end{algorithmic}
\end{algorithm}

\section{Client-level evaluation}\label{sec:evaluation}
In this section, we present a thorough experimental evaluation of our algorithm. We begin by introducing the general design of our experiments in Section~\ref{subsec:evaluation-setup}, describing the problems and baselines we have selected for evaluation. In Section~\ref{subsec:evaluation-main-baselines}, we show and discuss the first part of our main validation experiments, evaluating the performance of our algorithm with a focus on average client performance under objective heterogeneity. Following this section, we introduce a multi-objective view of this problem setting in Section~\ref{sec:evaluation-multiobjective}, giving a brief overview of common metrics, and analysing the performance of our algorithm with respect to these metrics.
These experiments are supplemented by studies of specific characteristics of the FedPref algorithm: in Section~\ref{subsec:evaluation-ablation}, we perform an ablation study, comparing the individual performance of the clustering and weighted aggregation components with the combined algorithm; in Section~\ref{subsec:evaluation-topr-minsim}, we analyse the sensitivity of the algorithm to two crucial parameters, and in Section~\ref{subsec:evaluation-clustering} we evaluate the clustering strategy.
\subsection{Implementation and setup}\label{subsec:evaluation-setup}
\begin{figure}[ht]
     \centering
     \begin{subfigure}[b]{0.32\columnwidth}
         \centering
         \includegraphics[width=\columnwidth]{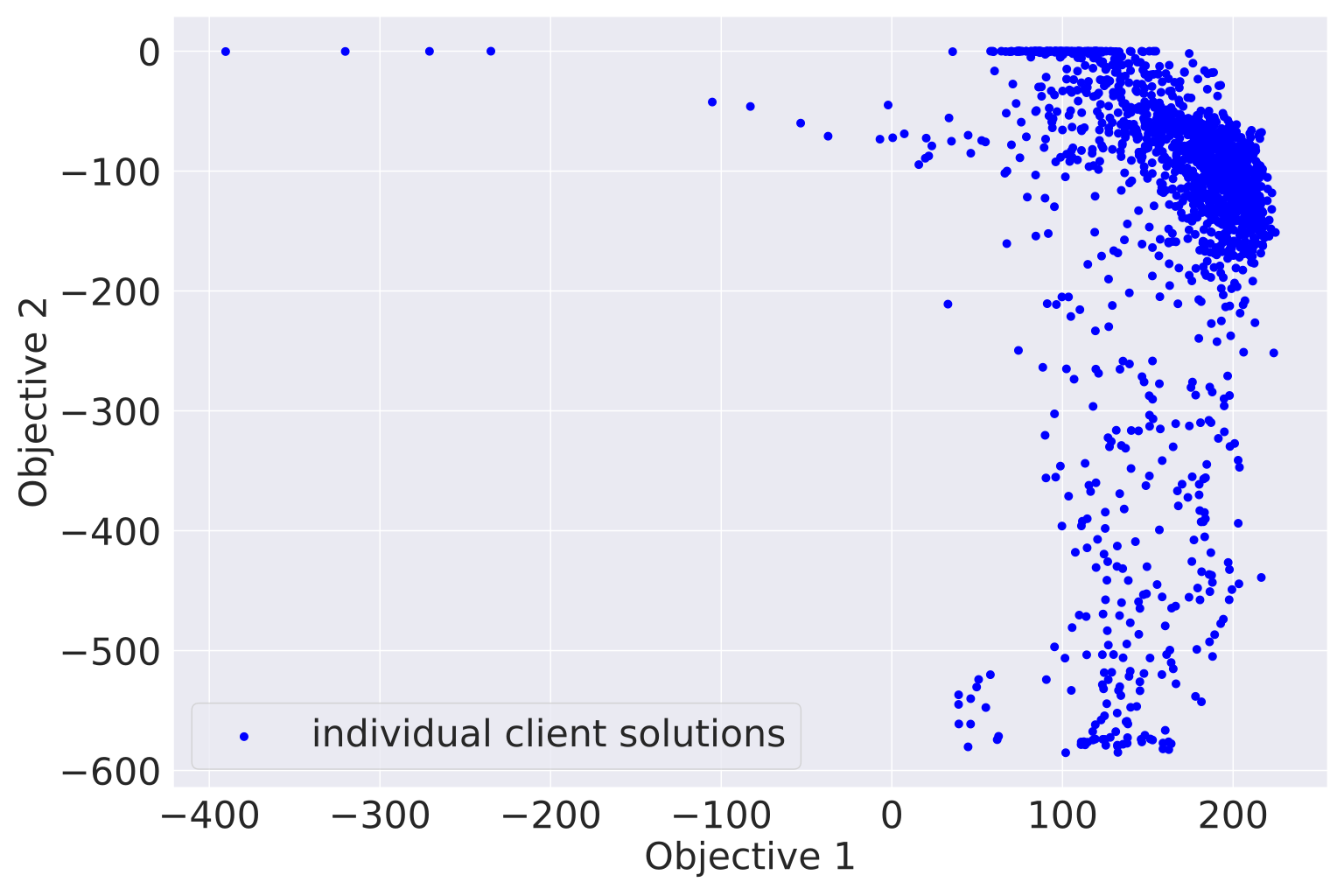}
     \end{subfigure}
     \hfill
     \begin{subfigure}[b]{0.32\columnwidth}
         \centering
         \includegraphics[width=\columnwidth]{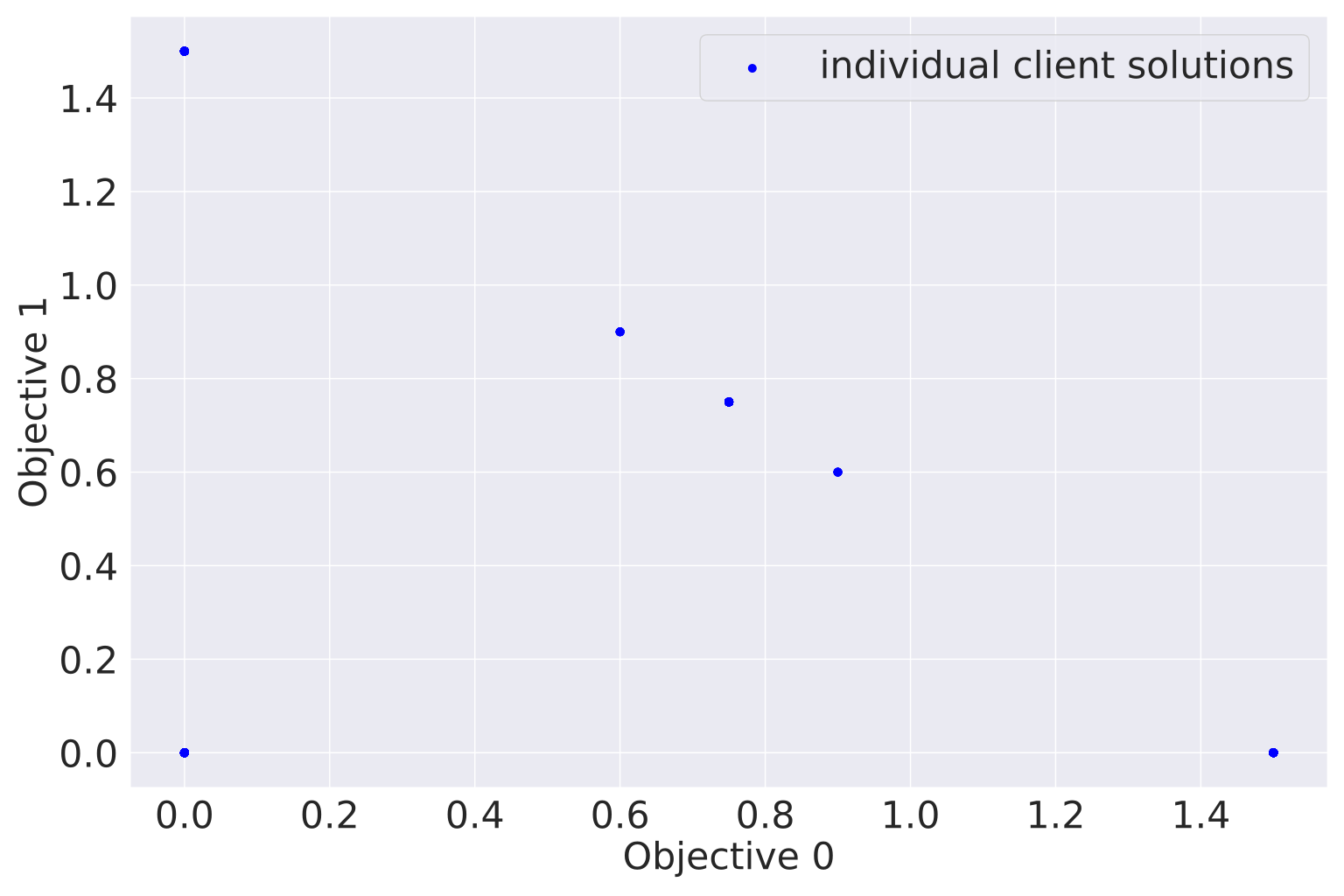}
     \end{subfigure}
     \hfill
     \begin{subfigure}[b]{0.32\columnwidth}
         \centering
         \includegraphics[width=\columnwidth]{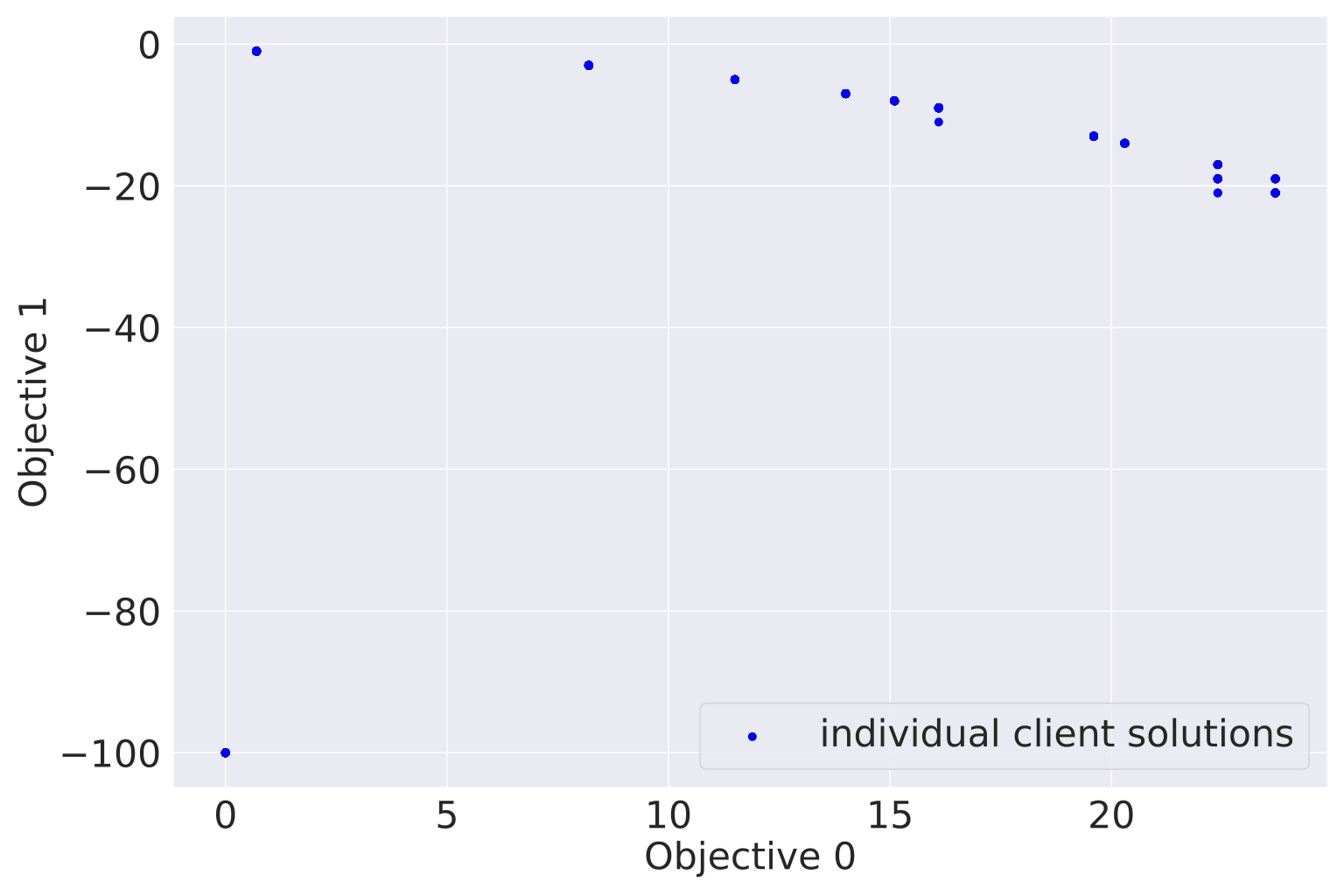}
     \end{subfigure}
          \begin{subfigure}[b]{0.32\columnwidth}
         \centering
         \includegraphics[width=\columnwidth]{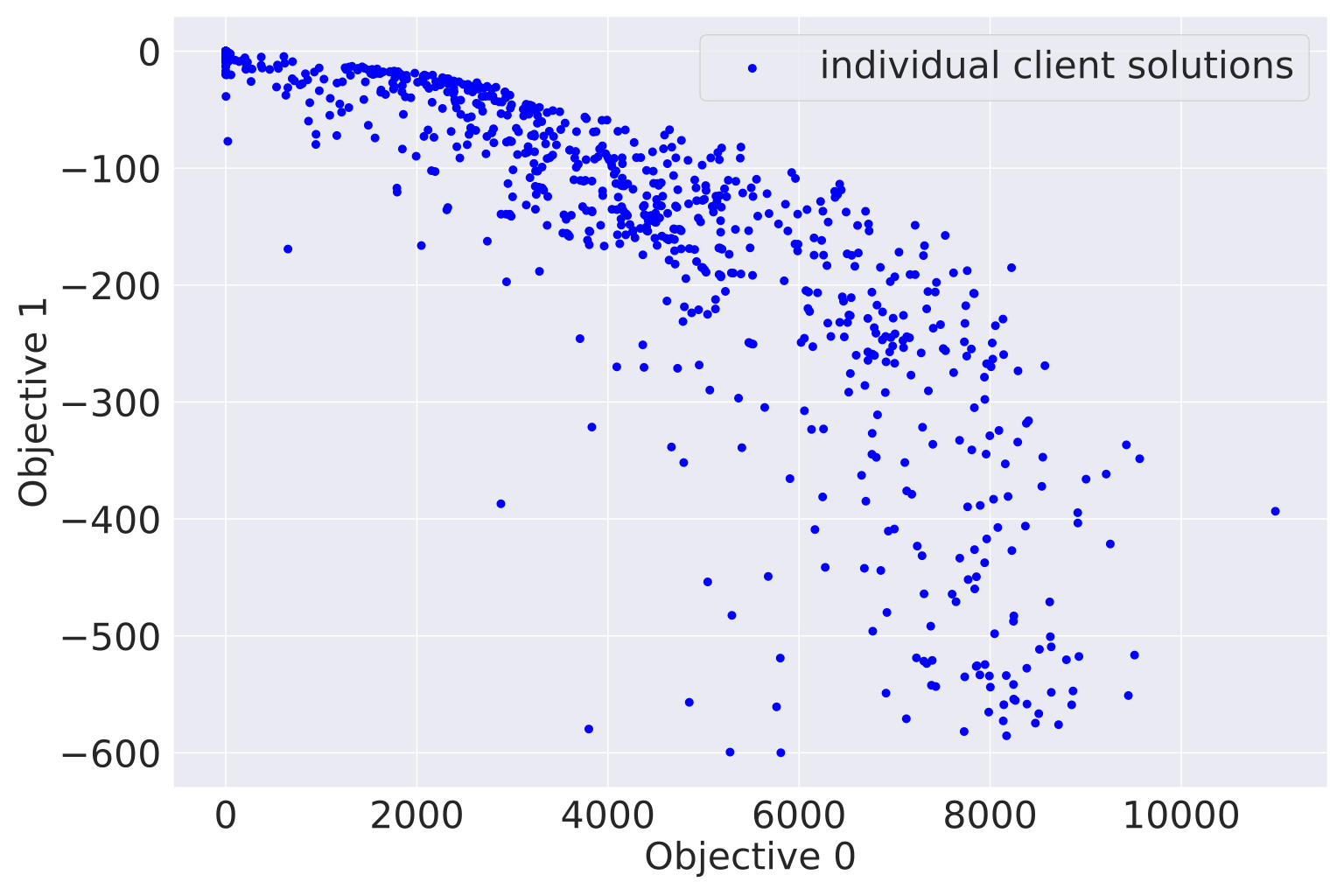}
     \end{subfigure}
          \begin{subfigure}[b]{0.32\columnwidth}
         \centering
         \includegraphics[width=\columnwidth]{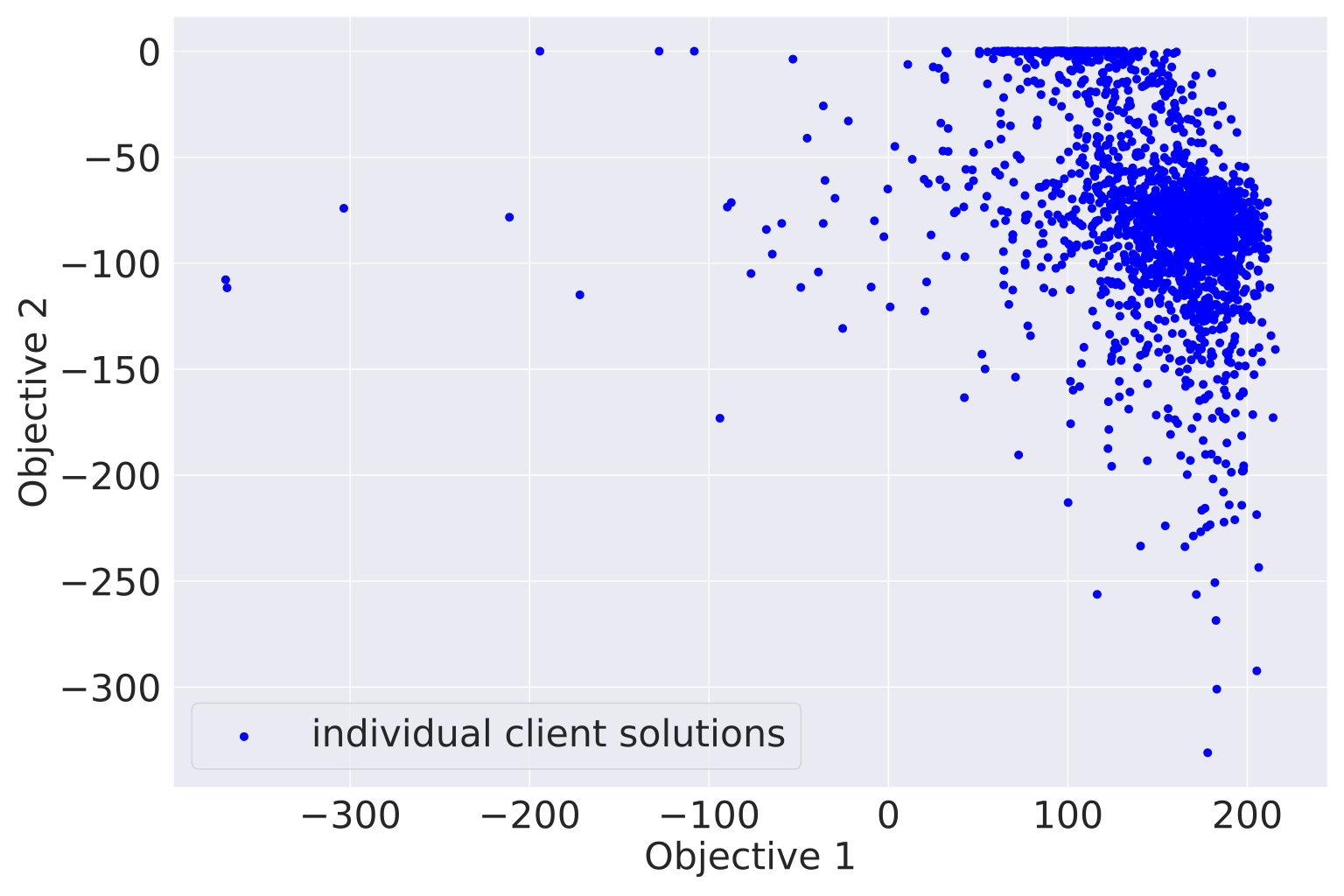}
     \end{subfigure}
        \caption{Sample illustrations of multi-objective solution spaces of different environments. Left to right: MO-Lunar Lander, Deterministic Minecart, Deep-Sea Treasure, MO-Halfcheetah and MO-LLcont.~environments. For MO-LL, DMC, and MO-LLcont., results have dimension $4, 3$ and $4$, respectively, and are here projected into a coordinate plane.}
        \Description{See text.}
        \label{fig:envs-pareto-fronts}
        \vspace{2em}
\end{figure}
\begin{figure}[hb]
     \centering
        \includegraphics[width=\columnwidth]{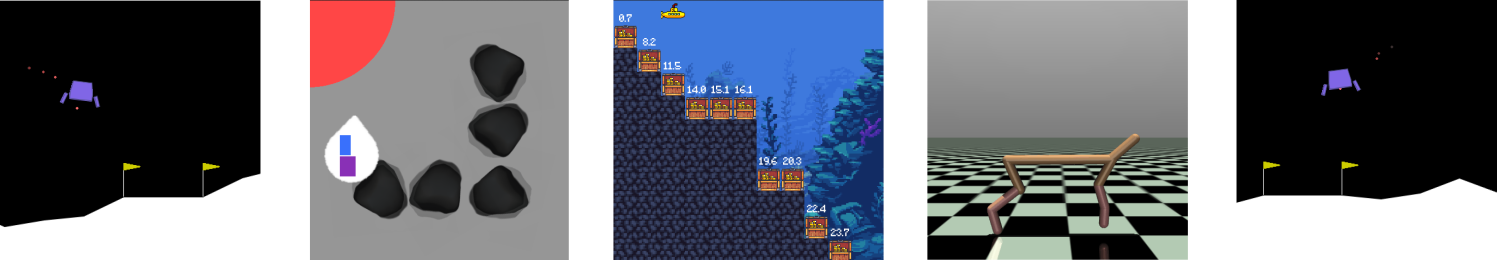}
        \caption{Illustration of multi-objective reinforcement learning environments used for validation experiments. Left to right: MO-Lunar Lander, Deterministic Minecart, Deep-Sea Treasure, MO-Halfcheetah and Continuous MO-Lunar Lander.}
        \label{fig:morl-envs}
        \Description{Left to right: A lunar lander hovering near a marked landing zone; a cart collecting ore from nearby deposits; a submarine diving for treasure; a two-legged, animal-like stick figure standing in a room; and a second flying lunar lander.}
        \vspace{2em}
\end{figure}

We implement our experimental framework using the PyTorch package in the Python programming language. The code of our implementation is publicly available\footnote{https://gitlab.com/maria.hartmann/FedPref}. Faced with a lack of standard multi-objective benchmarking problems for this class of problem, we choose to use a number of multi-objective reinforcement learning (MORL) environments as our validation problems. We reason that these represent an intuitive class of multi-objective problems with varying characteristics and complexity, are extensions of classical RL baselines, and are implemented in a well-documented set of Python libraries  \cite{felten_toolkit_2023}, making them easy to reproduce. We run our experiments on five such MORL environments (illustrated in Figure~\ref{fig:morl-envs}): Deep-Sea Treasure \cite{vamplew_empirical_2011} (DST), Deterministic Minecart \cite{abels_dynamic_2018} (DMC) and the multi-objective extension (MO-LL) of OpenAI's Lunar Lander gym environment, using a classical DQN algorithm \cite{Mnih2015} to solve the scalarised RL problem on each client; and the multi-objective extensions of the halfcheetah (MO-HC) and the continuous variant of the Lunar Lander environment (MO-LLc.), using the DDPG algorithm. These five selected environments represent multi-objective problems with different characteristics: the Deep-Sea Treasure environment is relatively small and has a finite number of optimal solutions. The MO-Lunar Lander environment is more complex and has a large number of optimal or near-optimal solutions closely aligned in the solution space. Conversely, the Deterministic Minecart environment has a sparse reward space, leading to a very low number of optimal solutions, which are mutually distant in the solution space. The MO-Halfcheetah and Continuous Lunar Lander environments, finally, produce continuous rewards and so require a different RL algorithm to be solved, allowing us to examine the effectiveness of the FedPref algorithm across different types of models. These differences, illustrated also in Figure~\ref{fig:envs-pareto-fronts}, present different challenges for federated aggregation. \\
The performance metric we report is the reward obtained by each client, according to its respective preference distribution. On all environments except MO-Halfcheetah, we run federated systems of $20$ clients each; due to the higher complexity of the MO-Halfcheetah environment, we limit the number of clients to $10$ in this case.
\subsection{Comparison to baselines}\label{subsec:evaluation-main-baselines}
\subsubsection{Experiments}
We compare our algorithm both to the classical baselines and to several algorithms developed to deal with other types of heterogeneity. As baselines, we run the same local learning algorithms with no communication between clients (no-communication) and the classical federated averaging (FedAvg) algorithm, aggregating all clients while disregarding heterogeneity. 
To the best of our knowledge, no previous algorithms that target this type of heterogeneity have been proposed in the literature; we therefore validate our approach against three additional algorithms from related fields that appear most relevant to our setting: FedProx \cite{li_federated_2018}, Many-Task Federated Learning \cite{cai_many_2023} (MaTFL) and Clustered Federated Learning \cite{Sattler2019ClusteredFL} (CFL). FedProx is a classical approach to the heterogeneity problem, commonly used as a baseline in data-heterogeneous settings. The underlying strategy appears intuitively to have the potential to transfer to the preference-heterogeneous setting, so we choose to retain this baseline. The MaTFL and CFL algorithms are chosen for the similarity of their approaches with the weighted aggregation and the clustering component of our algorithm, respectively; they also represent the two fields of Multi-Task Federated Learning and data-heterogeneous FL that we identified earlier in this work as most closely related to our problem setting. In addition to the standard versions of these algorithms from the literature, we also consider variants of the non-personalised algorithms that introduce a ``fine-tuning'' phase at the end of model training~\cite{wang_federated_2019}. Fine-tuning is a common strategy to allow a degree of personalisation between clients using an otherwise centralised training strategy. Following \cite{wang_federated_2019}, clients that perform fine-tuning end the training process with a single local training round instead of aggregating a global model. We label these variants as FedAvg+FT, FedProx+FT, CFL+FT, and FedPref+FT.
We tune the hyperparameters for all algorithms via an initial grid search on a set of preferences sampled from a Dirichlet distribution. For each algorithm and environment, we select the best-performing hyperparameter configuration from this search. The details of this parameter search and the local configurations of clients for each RL problem may be obtained from the supplementary material. 
Following the parameter tuning, we run all algorithms repeatedly, with client preference weights generated according to three different distributions: sampled from a Dirichlet distribution, sampled from a Gaussian distribution or weights generated to be equally spaced in the weight simplex. We report the results for all algorithms and distributions in Table~\ref{table:main-results}\footnote{The results are also available on an interactive online platform under the following link: https://wandb.ai/fed-mo/mofl-d/reports/Validating-FedPref-TOMPECS-{}-Vmlldzo5MTQ5MTEx?accessToken=pq620k7nw7fcbaq88s2cf2c89qsugjh0psa76wgzftcwqvcsxsxy8o8psrtsg8vy}. 
\subsubsection{Analysis}
We report the numerical results obtained for all algorithms and distributions in Table~\ref{table:main-results}. In the remainder of this section, we will discuss and contrast these results separately by preference distribution, from the most ``extreme'' preference differences between clients -- the equidistant distribution -- over the Dirichlet distribution to the Gaussian distribution, where client preferences are most similar.\\

\textbf{Equidistantly distributed preference weights.}\\
Under the equidistant distribution of preference weights across clients, we observe that variants of the FedPref algorithm outperform all other algorithms quite significantly on four out of five environments. 
On the MO-Halfcheetah environment, the FedPref+FT algorithm not only yields the highest average client reward of $3168.13$, but it also has a notably lower variance than all other algorithms. Similarly, on the Continuous MO-Lunar Lander environment, FedPref-FT achieves a markedly higher average client reward score than all those compared.
For the MO-LL environment, clients participating in the FedPref+FT algorithm obtain an average scalarised reward of $29.47$, far ahead of the second-highest result of $18.49$ achieved by another algorithm on the same environment. Indeed, the latter result is not accomplished by any federated algorithm, but by the baseline of non-communicating clients, with the remaining federated algorithms achieving much lower scores down to the lowest mean result of $-94.06$, returned by the CFL algorithm. Results for the Deep-Sea Treasure follow a similar pattern, while for the Deterministic Minecart environment no federated algorithm outperforms the result of the non-federated baseline. These results serve to underscore the difficulty of this heterogeneous distribution.\\
The case of equidistant preference weights likely represents the most ``extreme'' scenario among our experiments, where individual client objectives have on average the greatest mutual differences. In general, we would expect this to also map to greater differences in the models that match the preferences of each client, resulting in an advantage for those algorithms training personalised models. This is indeed illustrated in our results, reported in rows 11-20 of Table~\ref{table:main-results}, as all variants of the FedProx and FedAvg algorithms perform notably worse in this scenario than for the other two types of preference distributions. This pattern persists across all experimental environments trained locally using the DQN algorithm (i.e.~MO-LL, DMC and DST). 
More surprisingly, we also observe a poor performance by the CFL algorithm on these environments in this setting - further investigation, reported in the appendix, shows that  CFL \cite{Sattler2019ClusteredFL} tends to yield highly unbalanced clusters in our experiments. As clusters in CFL train a single global model, this can lead to many clients with less personalised models, combined with a small number of clients that are separated early from the collaborative cluster -- for this type of preference distribution, it is likely the large cluster that leads to a crucial lack of diversity. In other environments, the clustering step of CFL is not triggered at all.\\
On the MO-Halfcheetah and Continuous MO-Lunar Lander environments, CFL and the two non-PFL algorithms perform notably better in comparison, though still worse than the FedPref algorithm.
Indeed, in this scenario it becomes particularly important for any personalised algorithm to be able to accurately judge the compatibility of models, and to separate non-compatible models. This appears to be a strength of our algorithm: FedPref not only outperforms all others by a significant margin in four out of five environments. In the fifth environment - the Deterministic Minecart - none of the algorithms tested in our experiments perform better than the non-federated baseline. This might be the result of the high sparsity of the reward space, combined with the greater difference in client objectives, that make it difficult to group clients for aggregation.\\
With respect to the fine-tuning variants, we observe that the addition of a fine-tuning step does not generally lead to improved performances for the compared algorithms. A notable exception is the MO-LL environment, where the algorithms performing non-personalised aggregation deliver notably poor results without fine-tuning, with a drastic relative improvement with the addition of a fine-tuning step. However, the overall results in these cases are still markedly low. It appears that these algorithms fail to converge to a meaningful common solution across clients of such high preference diversity. In this context, disengaging from the federation naturally leads to improved local results.\\

\textbf{Uniformly distributed preference weights.}\\
In terms of expected client similarity, the Dirichlet preference distribution represents a ``middle ground'' between the other two types of distribution explored in this work. Preference weights are sampled uniformly at random from the weight simplex. %
Our results, reported in rows 1-10 of Table~\ref{table:main-results}, show variants of the FedPref algorithm again outperforming all others on four out of five experimental environments. Compared to the results obtained under the equidistant preference distribution, some of the gains of the FedPref algorithm over those compared, though still existent, are less drastic, particularly on the relatively dense solution space of the MO-HC and MO-LL environments: on MO-LL, e.g.~the FedProx algorithm yields a mean scalarised client reward of $31.2$, relatively close to the top result of $37.27$ achieved by the FedPref-FT algorithm. For the MO-LLc.~environment, which appears to have an even higher localised density than MO-LL, the FedAvg algorithm even outperforms the FedPref algorithm under this distribution.
However, the difference in favour of FedPref remains larger for the Deep-Sea Treasure environment, likely due to its discrete solution set: Here, the FedPref+FT variant of our algorithm obtains a mean scalarised client reward of $4.41$, still followed by the no-communication baseline with an average reward of $-0.43$. The ranking of algorithmic results is similar on the Deterministic Minecart environment, though less decisive. It appears that the lower density of (optimal) solutions available in the latter two environments, combined with the intermediate objective heterogeneity of this setting, continues to present a difficult challenge to the federated algorithms from the literature.\\ 
The addition of a fine-tuning step again yields lower mean client scores for all algorithms performing non-personalised aggregation. In most cases, the results for the fine-tuned variant of an algorithm are markedly worse than for the same algorithm without fine-tuning. We theorise that over the course of the federated training process, clients in the federated system have learned to converge to a mutually beneficial, globally optimal ``compromise'' solution. This compromise, however, is quite fragile, as all individual clients operate under different preferences, i.e.~different loss functions, and the distance between client-optimal solutions and the global model appears too great to overcome during fine-tuning. The difference in performance between fine-tuning and non-fine-tuning variants is less great for the FedPref algorithm, most likely because diverse clients are separated more successfully at an earlier stage of the training process.\\

\textbf{Gaussian-distributed preference weights.}\\
Finally, in the setting where weight preferences are drawn from a Gaussian distribution, three different algorithms achieve the top scores for different environments (see results in row 21-30 of Table~\ref{table:main-results}): %
Results on the Deep-Sea Treasure environment remain dominated by both variants of the FedPref algorithm, with no other federated algorithm outperforming the non-federated baseline. Similarly, FedPref-FT outperforms all others on the MO-LL and MO-Halfcheetah environments. For the compared algorithms, the addition of a fine-tuning step to the various algorithms has a similarly negative effect as in the experiments under a Dirichlet preference distribution.\\
Under the Gaussian distribution, clients are more likely to have more similar preferences, potentially supporting more similar models. In this case, plain (equally-weighted) aggregation appears to do well, with the CFL algorithm delivering the second-best performance on the MO-LL environment. The two non-PFL algorithms also perform notably better under this preference distribution than in the other two settings - in fact, in this case the plain FedAvg algorithm outperforms all others in the DMC environment, and the FedProx algorithm achieves the top result in the Continuous MO-Lunar Lander environment. The former result may in part be owing to the sparse solution space of the problem, with the clients in federation jointly converging on a single local optimum; but it is nonetheless part of a wider trend. 
In contrast to the very different performance of the compared algorithms on some environments, FedPref appears to adapt quite well to this setting, delivering the best performance in two environments and the second-best in two others.\\

In general, these results indicate that the FedPref algorithm is capable of adapting to a range of different preference distributions and problem types, outperforming all compared algorithms in the majority of experiments. In almost all cases where our algorithm does not deliver the best performance, it is outperformed by only one other, and by different algorithms for different problems. Furthermore, these results are preserved across different local training algorithms and different model architectures.
This shows the high flexibility and robustness of our algorithm, making it a good overall choice in the general case, where the distribution of preference weights of the characteristics of the learning problem may be unknown.

\begin{table*}
\caption{Experimental results comparing our proposed FedPref algorithm to MaTFL, CFL, FedProx, FedAvg and individual learning without cooperation.}\label{table:main-results}
\centering
\begin{tabular}{lllllll} 
\toprule 
 & & MO-LL($\uparrow$)& DMC($\uparrow$)& DST($\uparrow$)& MO-HC($\uparrow$)& MO-LLc.($\uparrow$)\\ 
\toprule 
\multirow{10}{*}{Dirichlet} & No comm.& $14.32~\sigma 13.3$& $-2.52~\sigma 0.9$& $-0.43~\sigma 1.8$& $2440.30~\sigma 511.3$& $11.95~\sigma 12.6$\\
 & FedAvg& $30.52~\sigma 14.7$& $-3.20~\sigma 3.4$& $-16.50~\sigma 24.7$& $2234.06~\sigma 1202.9$& \textbf{34.19}$~\sigma 11.4$\\
 &  FedAvg+FT& $4.31~\sigma 11.0$& $-5.41~\sigma 0.9$& $-36.40~\sigma 10.0$& $2901.84~\sigma 798.1$& $16.29~\sigma 6.3$\\
 & FedProx& $31.20~\sigma 16.9$& $-3.34~\sigma 3.6$& $-11.05~\sigma 22.1$& $2172.35~\sigma 1231.2$& $32.45~\sigma 11.2$\\
 &  FedProx+FT& $3.21~\sigma 11.3$& $-5.44~\sigma 0.6$& $-34.52~\sigma 7.2$& $3017.22~\sigma 795.3$& $13.82~\sigma 12.5$\\
 & CFL& $31.18~\sigma 17.8$& $-2.76~\sigma 0.9$& $-12.90~\sigma 21.8$& $2835.50~\sigma 817.7$& $26.90~\sigma 12.5$\\
 &  CFL+FT& $10.09~\sigma 13.2$& $-3.82~\sigma 2.1$& $-35.14~\sigma 10.2$& $2864.62~\sigma 871.0$& $19.81~\sigma 7.6$\\
 & MaTFL& $7.82~\sigma 9.9$& $-4.39~\sigma 2.2$& $-6.32~\sigma 3.6$& $1596.59~\sigma 558.9$& $10.98~\sigma 7.3$\\
 & FedPref+FT (ours)& $32.22~\sigma 11.3$& $-2.42~\sigma 1.6$& \textbf{4.41}$~\sigma 1.7$& $2980.26~\sigma 784.4$& $32.17~\sigma 7.0$\\
 &  FedPref-FT (ours)& \textbf{37.27}$~\sigma 11.9$& \textbf{-1.90}$~\sigma 1.2$& $1.21~\sigma 3.4$& \textbf{3104.59}$~\sigma 742.3$& $32.91~\sigma 9.8$\\
\toprule 
\multirow{10}{*}{Equidistant} & No comm.& $18.49~\sigma 10.5$& \textbf{-1.70}$~\sigma 1.5$& $0.68~\sigma 1.9$& $2265.67~\sigma 440.5$& $22.49~\sigma 6.9$\\
 & FedAvg& $-55.64~\sigma 46.9$& $-6.77~\sigma 0.3$& $-17.87~\sigma 26.2$& $1952.42~\sigma 321.4$& $28.04~\sigma 6.8$\\
 &  FedAvg+FT& $-16.95~\sigma 5.6$& $-6.24~\sigma 0.4$& $-36.01~\sigma 3.1$& $3023.42~\sigma 204.2$& $6.42~\sigma 8.1$\\
 & FedProx& $-73.98~\sigma 45.3$& $-6.75~\sigma 0.2$& $-23.23~\sigma 26.8$& $1948.46~\sigma 532.2$& $27.75~\sigma 11.5$\\
 &  FedProx+FT& $-9.69~\sigma 4.9$& $-6.08~\sigma 0.4$& $-36.17~\sigma 2.6$& $3019.46~\sigma 206.9$& $1.01~\sigma 8.2$\\
 & CFL& $-94.06~\sigma 30.8$& $-2.74~\sigma 0.5$& $-34.62~\sigma 23.5$& $2831.28~\sigma 304.4$& $25.97~\sigma 4.6$\\
 &  CFL+FT& $4.70~\sigma 5.6$& $-2.35~\sigma 0.7$& $-37.17~\sigma 2.9$& $3097.66~\sigma 282.7$& $11.71~\sigma 10.4$\\
 & MaTFL& $11.99~\sigma 6.5$& $-4.03~\sigma 1.2$& $-6.16~\sigma 2.8$& $1643.90~\sigma 239.3$& $19.20~\sigma 5.6$\\
 & FedPref+FT (ours)& \textbf{29.47}$~\sigma 3.5$& $-2.21~\sigma 1.7$& \textbf{2.78}$~\sigma 2.5$& \textbf{3168.13}$~\sigma 145.7$& $36.52~\sigma 4.8$\\
 &  FedPref-FT (ours)& $29.26~\sigma 4.2$& $-2.35~\sigma 1.8$& $2.26~\sigma 2.2$& $3044.61~\sigma 239.9$& \textbf{45.28}$~\sigma 11.8$\\
\toprule 
\multirow{10}{*}{Gaussian} & No comm.& $15.33~\sigma 13.4$& $-3.61~\sigma 2.2$& $1.13~\sigma 0.9$& $2575.39~\sigma 790.7$& $17.08~\sigma 5.9$\\
 & FedAvg& $31.53~\sigma 13.1$& \textbf{-2.47}$~\sigma 3.3$& $-13.32~\sigma 25.6$& $2028.17~\sigma 1282.5$& $33.15~\sigma 11.3$\\
 &  FedAvg+FT& $14.99~\sigma 9.7$& $-3.87~\sigma 2.1$& $-37.42~\sigma 5.5$& $2758.24~\sigma 1081.8$& $19.37~\sigma 8.5$\\
 & FedProx& $32.75~\sigma 13.0$& $-2.94~\sigma 3.1$& $-2.07~\sigma 16.0$& $2042.50~\sigma 1183.2$& \textbf{33.31}$~\sigma 7.3$\\
 & FedProx+FT& $14.61~\sigma 10.1$& $-4.41~\sigma 1.7$& $-37.15~\sigma 9.1$& $2732.12~\sigma 940.0$& $19.72~\sigma 8.0$\\
 & CFL& $37.73~\sigma 11.5$& $-4.00~\sigma 1.8$& $-24.82~\sigma 27.5$& $2635.25~\sigma 812.3$& $25.56~\sigma 8.0$\\
 &  CFL+FT& $18.30~\sigma 9.2$& $-3.71~\sigma 1.6$& $-33.75~\sigma 8.3$& $2852.83~\sigma 776.1$& $16.41~\sigma 8.5$\\
 & MaTFL& $6.48~\sigma 9.4$& $-5.04~\sigma 1.1$& $-6.24~\sigma 3.4$& $1355.33~\sigma 489.7$& $13.36~\sigma 7.1$\\
 & FedPref+FT (ours)& $36.43~\sigma 7.3$& $-2.53~\sigma 1.8$& $2.87~\sigma 2.6$& $2786.28~\sigma 888.8$& $33.06~\sigma 6.7$\\
 &  FedPref-FT (ours)& \textbf{38.90}$~\sigma 8.8$& $-2.86~\sigma 1.5$& \textbf{2.88}$~\sigma 3.4$& \textbf{2923.27}$~\sigma 897.0$& $33.18~\sigma 7.7$\\
\bottomrule
\end{tabular}
\end{table*}

\subsection{Ablation study}\label{subsec:evaluation-ablation}
We perform an ablation study of our algorithm, comparing the performance of the full algorithm with that of its individual components, i.e.~performing only the weighted aggregation strategy or only the clustering strategy, respectively. For all three configurations, we also consider variants that perform a single round of fine-tuning at the end of the training process. The hyperparameters of all versions remain fixed to the values obtained for our algorithm in previous experiments.
The results, reported in Table~\ref{table:ablation-study}, show different outcomes for the  different types of problems we study. The effect of fine-tuning, too, appears to vary for different environments.\\
For the Deterministic Minecart environment with its very sparse reward space, we observe that the Clustering+FT component performs better individually than combined -- the Clustering+FT component achieves the highest average scalarised client reward of $-1.53$, whereas the combined components yield a mean reward of $-1.90$ without fine-tuning. In this case, it is likely that the individual clients' preferences ultimately lead to very different optimal models, with less benefit obtained from extensive cooperation between different models. Separating clients early enough during the training process would then be crucial, before a ``compromise'' model emerges that differs greatly from individually optimal models. Otherwise, such a consensus model might be so different from the optimal fit for an individual clients' preferences that the latter becomes too hard to recover in training even if clients are separated at a later stage.\\
As a side remark, we note that this hypothesis is also supported when comparing the results for the FedProx and FedAvg algorithm, discussed in Section~\ref{subsec:evaluation-main-baselines}, Table~\ref{table:main-results}, for this environment. The approach behind these two algorithms forces a high level of collaboration between the clients, and does not lead to high overall results for this environment when preferences are sufficiently different. Even fine-tuning does not generally deliver improvements, likely because the at the end of training, the global consensus model is too far distant from locally optimal models to reach.\\
Returning to the ablation study, we suggest that the clustering component without weighted aggregation likely succeeds more quickly in separating very different models early during the training process, with the cluster-mean model converging more definitively. This separation appears not fully effective, as seen by the poor performance of the clustering component without fine-tuning; yet it succeeds in enabling the training of models that are sufficiently diverse that a single fine-tuning round can ameliorate these problems. The weighted-aggregation component in isolation and the full FedPref algorithm appear slightly less successful at separating diverse clients early during training, leading to consensus models that do not improve with fine-tuning. In spite of this, variants of all three versions succeed in outperforming the compared approaches under the Dirichlet distribution -- see Table~\ref{table:main-results}.\\
Interestingly, the success of the individual clustering component over the FedPref algorithm is reversed on DST, the other environment with a sparser solution space. Here, the combination of clustering and weighted aggregation appears to encourage effective separation. 
In contrast, we observe significantly improved results for the combination of both components over each component individually for the MO-LL and MO-HC environments, both when comparing fine-tuning variants and between non-fine-tuning variants. 
For the MO-Halfcheetah environment, we observe rather similar average client performances for variants of all three configurations, with the FedPref-FT algorithm outperforming the other configurations. Once again, fine-tuning slightly decreases performance for the clustering component and the combined components, indicating that perhaps in these cases some clients might have benefited from earlier separation. However, this effect is not strong here.
The results for the Continuous MO-Lunar Lander show a quite similar pattern to the MO-LL environment, with the main difference being the relatively weaker performance of FedPref-FT. In this case, the Clustering-FL component outperforms both FedPref variants.
We draw three general impressions from the ablation study: \begin{itemize}
    \item A variant of the FedPref algorithm performs better than its individual components on three out of five environments, and results near to the highest score in the other two cases.
    \item The success of each component appears related to the characteristics of the problem being solved.
    \item The effect of a fine-tuning step may serve as a useful indicator of the effectiveness of diverse clients' cooperation within the FedPref algorithm. 
\end{itemize}
\begin{table}[ht]
\caption{Experimental results comparing the individual components of our algorithm.}\label{table:ablation-study}
\centering
\begin{tabular}{llllll} 
& MO-LL & DMC & DST & MO-HC & MO-LLc.\\
\toprule
Clustering only + FT & $27.27~\sigma 13.9$ & \textbf{-1.53}$~\sigma 0.8$ & $0.95 ~\sigma 3.8$ & $2940.45 ~\sigma 982.4$ & $28.24~\sigma 9.7$ \\
Clustering only - FT & $34.21~\sigma 16.7$ & $-4.27~\sigma 7.1$ & $0.55~\sigma 3.1$ & $3049.63~\sigma 683.1$ & \textbf{34.58}$~\sigma 15.4$ \\
Weighted agg.~only + FT & $17.70~\sigma 9.8$ & $-2.20 ~\sigma 0.9$ & $-31.24 ~\sigma 8.5$ & $3044.93~\sigma 821.6$ & $26.12~\sigma 9.3$ \\
Weighted agg.~only - FT & $32.54~\sigma 21.1$ & $-2.08~\sigma 1.1$ & $-21.83~\sigma 13.6$ & $2425.13~\sigma 1490.2$ & $34.36~\sigma 15.1$ \\
FedPref+FT (combined) & $32.22~\sigma 12.0$ & $-2.42 ~\sigma 1.7$ & \textbf{4.41}$~\sigma 1.8$ & $2980.26~\sigma 826.8$ & $32.17~\sigma 7.4$ \\
FedPref-FT (combined) & \textbf{37.27}$~\sigma 12.5$ & $-1.90~\sigma 1.2$ & $1.21~\sigma 3.6$ & \textbf{3104.59}$~\sigma 782.5$ & $32.91~\sigma 10.4$ \\
\bottomrule
\end{tabular}
\end{table}

\subsection{Impact of topR parameter and similarity bound}\label{subsec:evaluation-topr-minsim}

\begin{figure}[ht]
     \centering
     \begin{subfigure}[b]{0.19\columnwidth}
         \centering
         \includegraphics[width=\columnwidth]{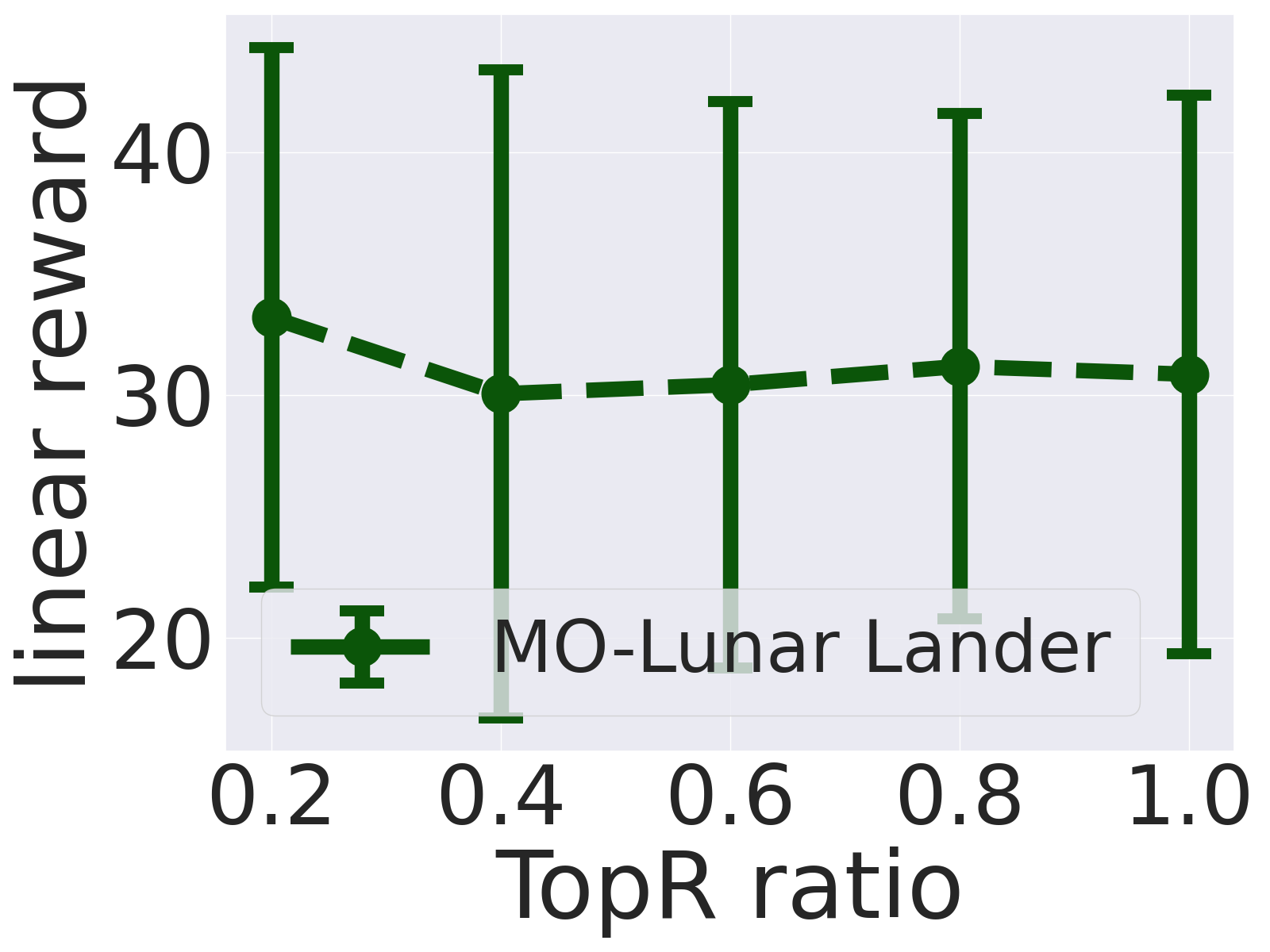}
     \end{subfigure}
     \hfill
     \begin{subfigure}[b]{0.19\columnwidth}
         \centering
         \includegraphics[width=\columnwidth]{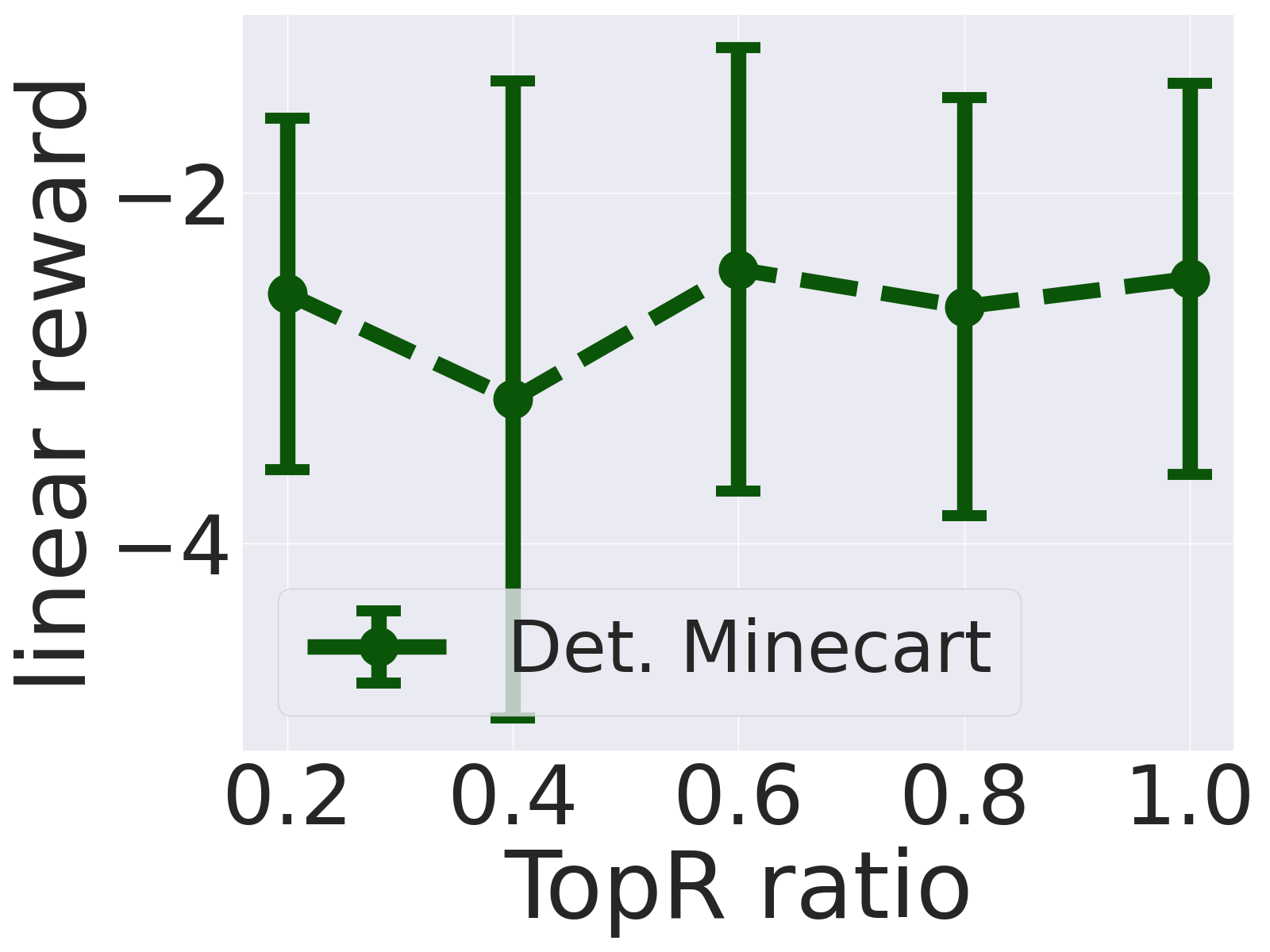}
     \end{subfigure}
     \hfill
     \begin{subfigure}[b]{0.19\columnwidth}
         \centering
         \includegraphics[width=\columnwidth]{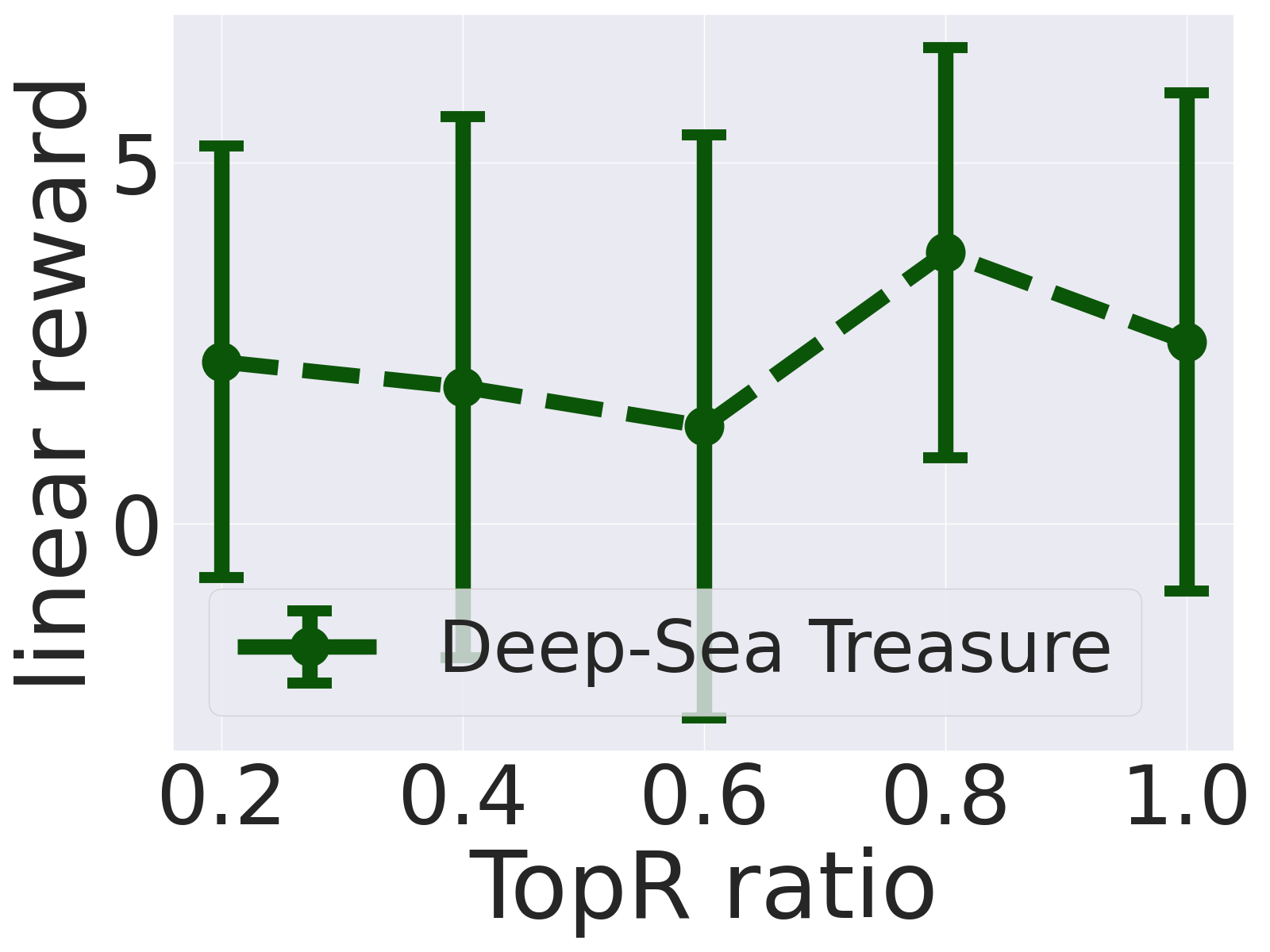}
     \end{subfigure}
     \begin{subfigure}[b]{0.19\columnwidth}
         \centering
         \includegraphics[width=\columnwidth]{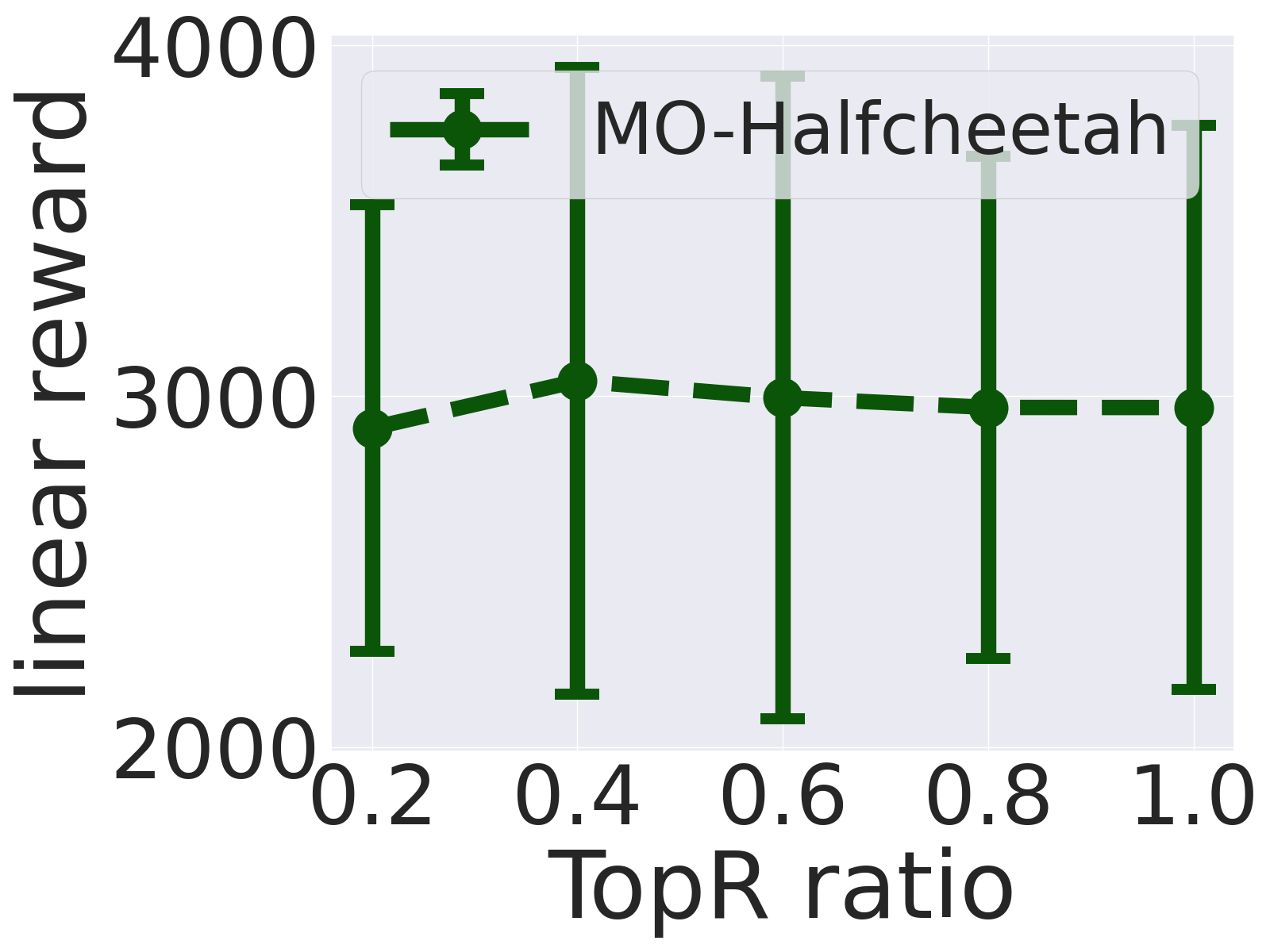}
     \end{subfigure}
     \begin{subfigure}[b]{0.19\columnwidth}
         \centering
         \includegraphics[width=\columnwidth]{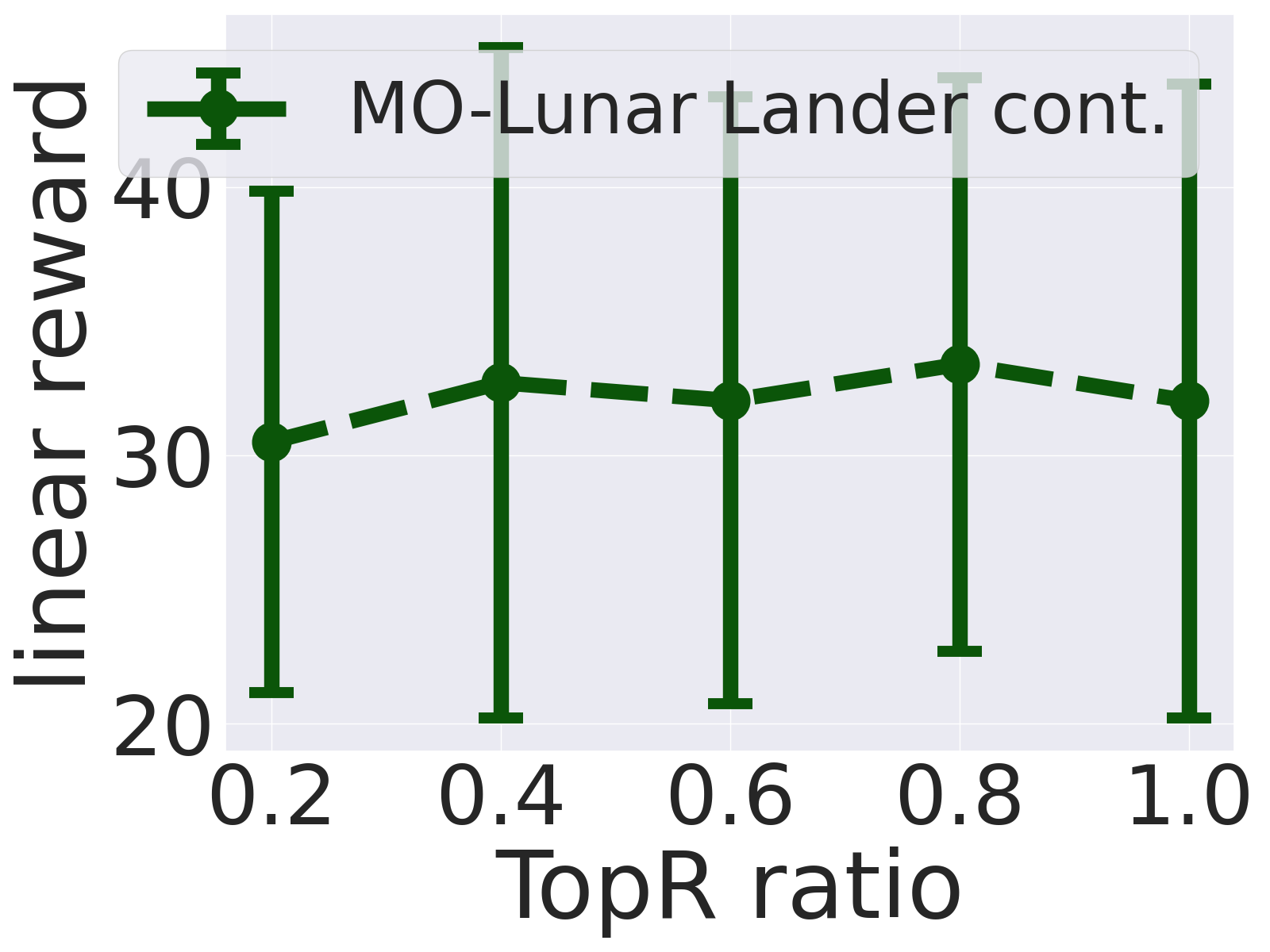}
     \end{subfigure}
        \caption{Impact of the choice of $topR$ parameter on average reward obtained by clients. Left to right: results for MO-Lunar Lander, Deterministic Minecart, Deep-Sea Treasure, MO-Halfcheetah and Continuous MO-Lunar Lander environments.}
        \Description{See text.}
        \label{fig:topr-sensitivity}
        \vspace{2em}
\end{figure}
\begin{figure}[hb]
     \centering
     \begin{subfigure}[b]{0.3\columnwidth}
         \centering
         \includegraphics[width=\columnwidth]{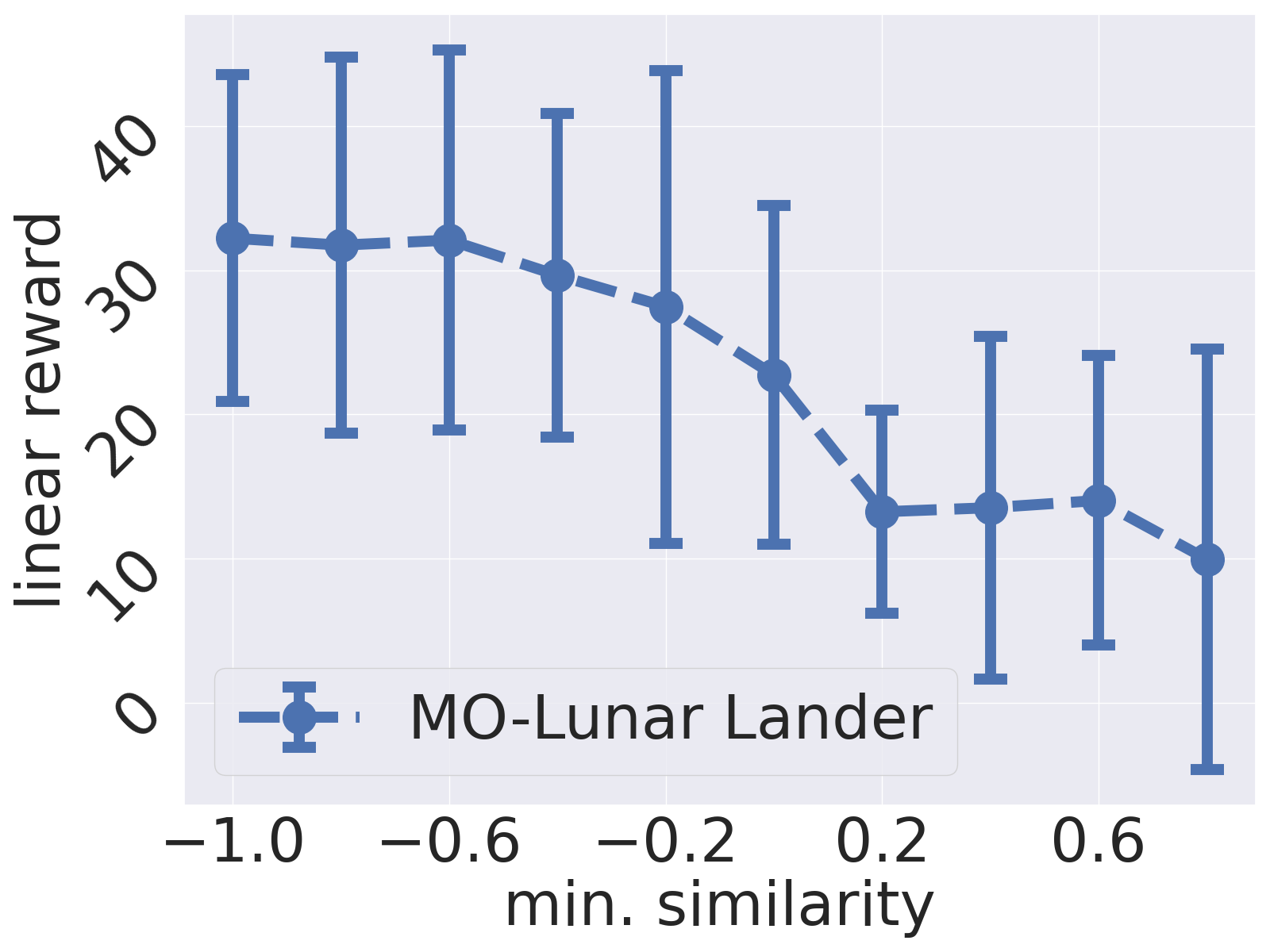}
     \end{subfigure}
     \hfill
     \begin{subfigure}[b]{0.3\columnwidth}
         \centering
         \includegraphics[width=\columnwidth]{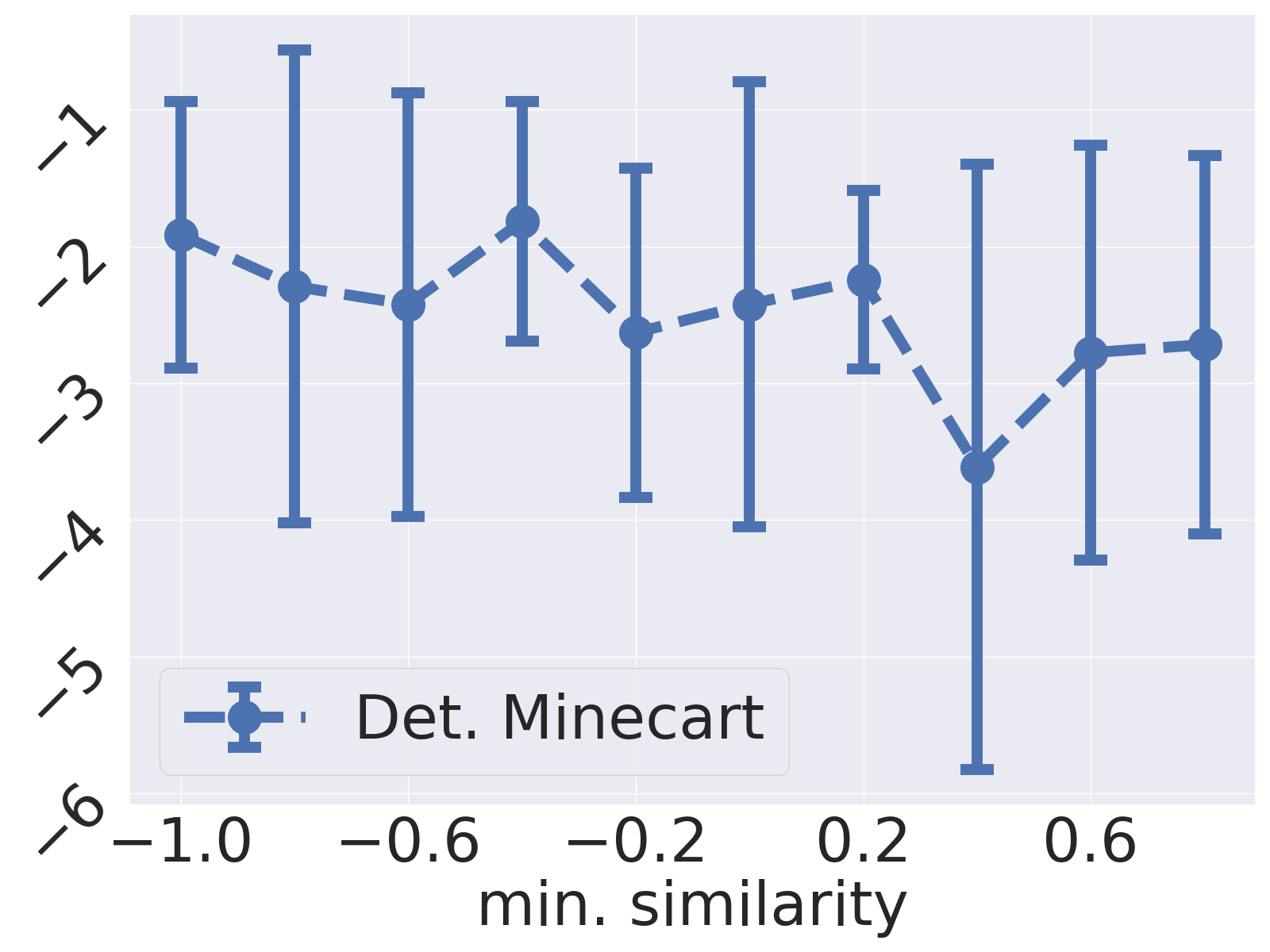}
     \end{subfigure}
     \hfill
     \begin{subfigure}[b]{0.3\columnwidth}
         \centering
         \includegraphics[width=\columnwidth]{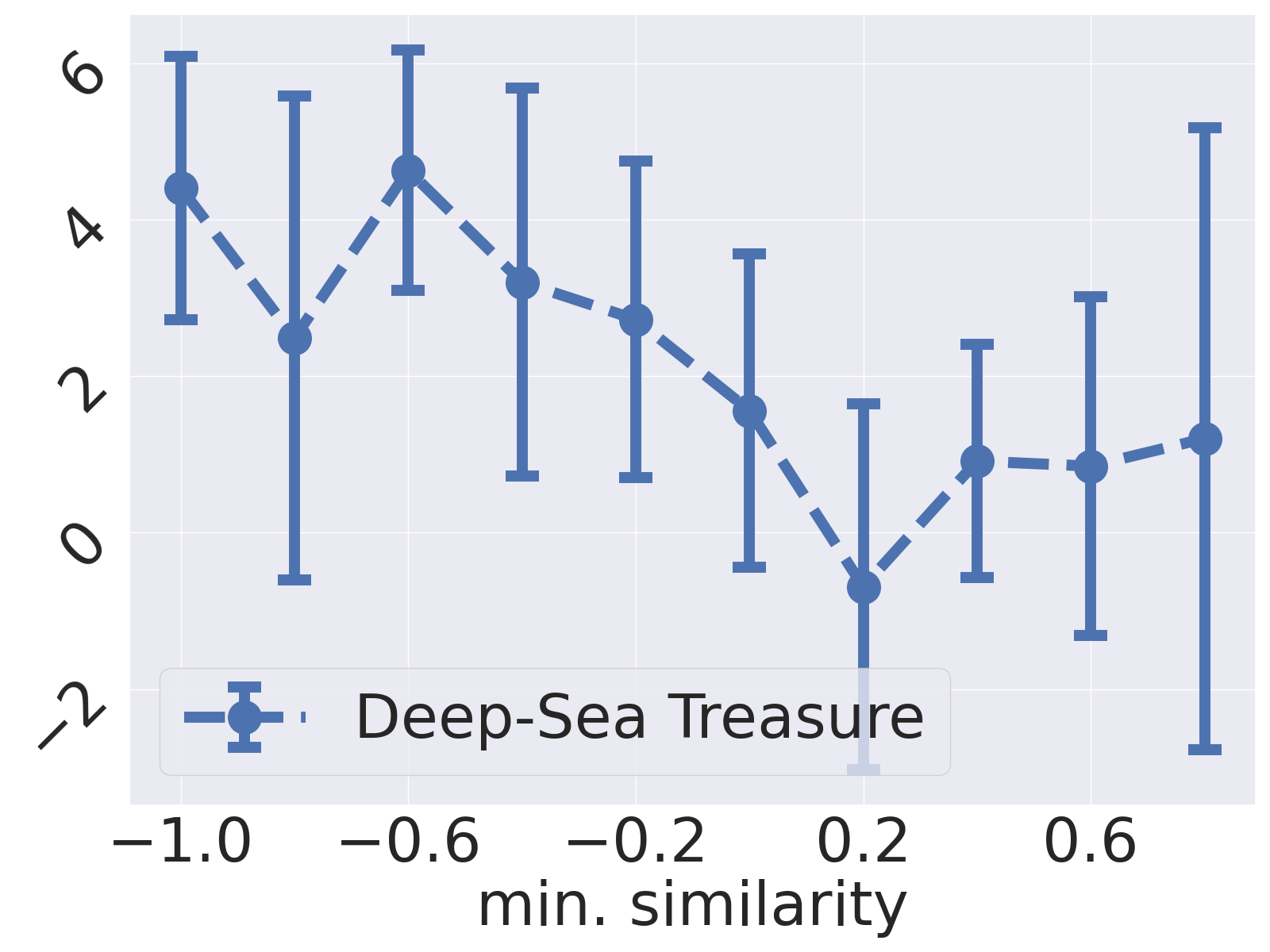}
     \end{subfigure}
     \begin{subfigure}[b]{0.3\columnwidth}
         \centering
         \includegraphics[width=\columnwidth]{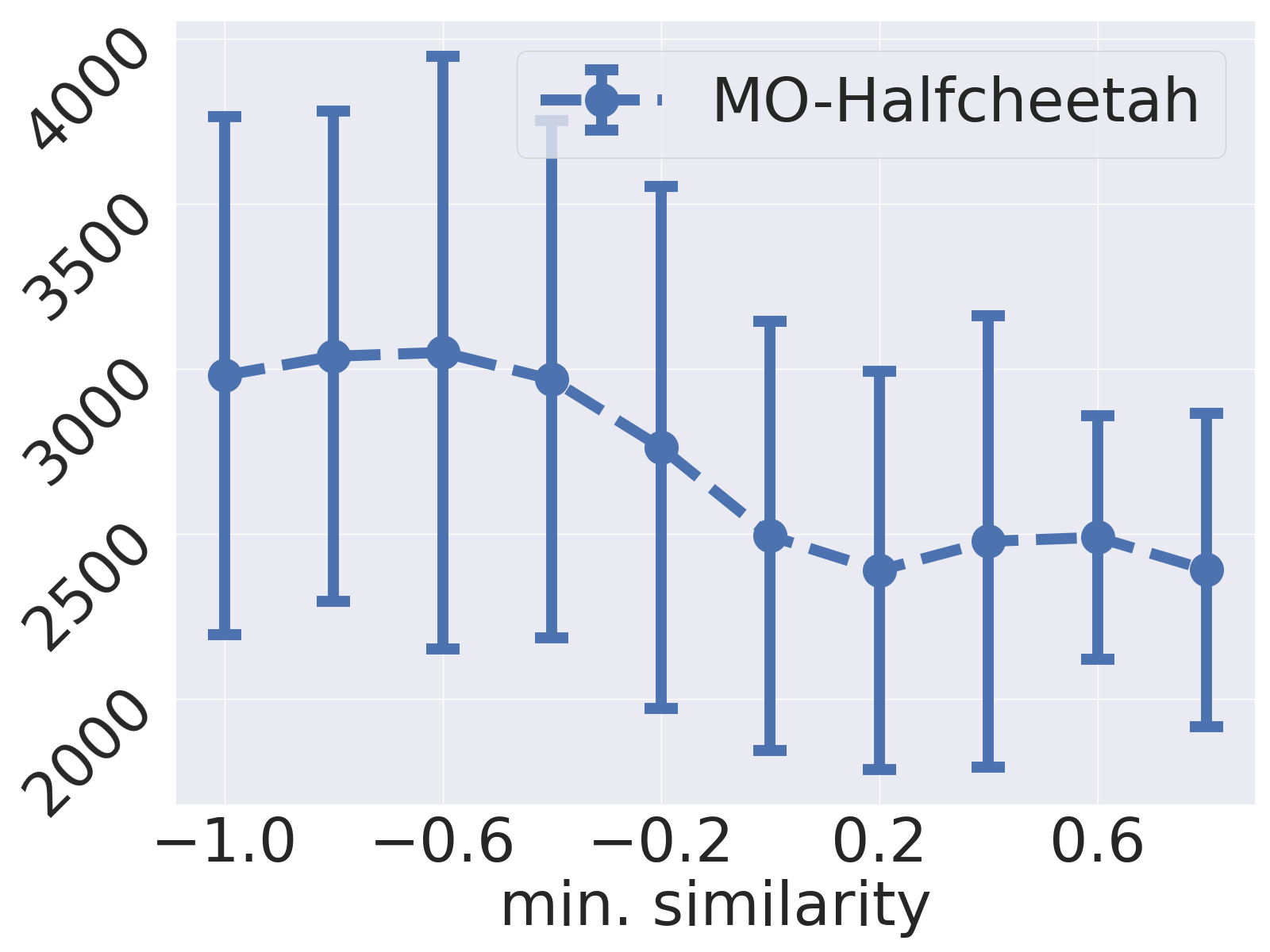}
     \end{subfigure}
     \begin{subfigure}[b]{0.3\columnwidth}
         \centering
         \includegraphics[width=\columnwidth]{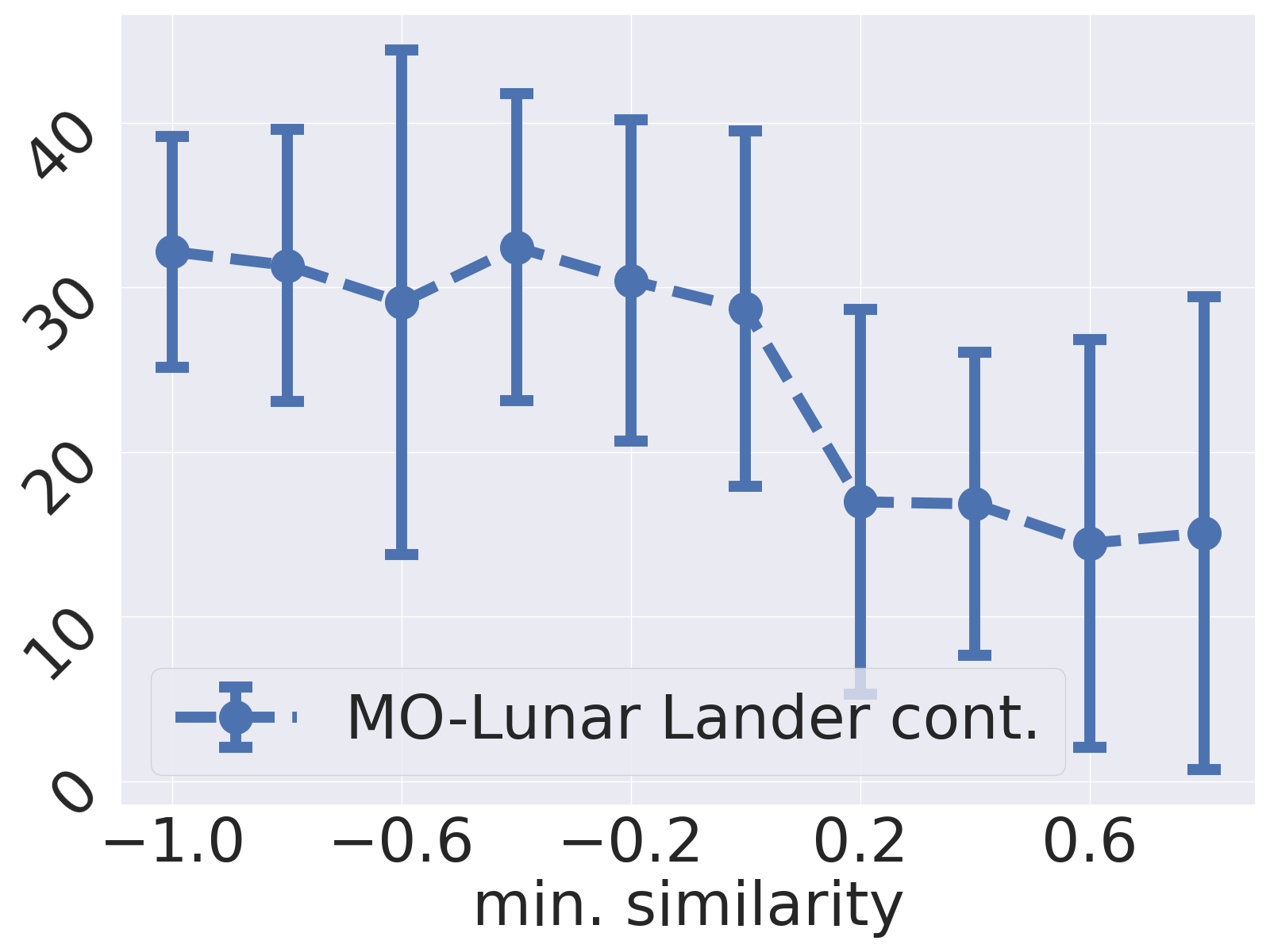}
     \end{subfigure}
        \caption{Impact of the choice of minimum-similarity threshold on average reward obtained by clients. Left to right: results for MO-Lunar Lander, Deterministic Minecart, Deep-Sea Treasure, MO-Halfcheetah and Continuous MO-Lunar Lander environments.}
        \Description{See text.}
        \label{fig:minsim-sensitivity}
        \vspace{2em}
\end{figure}
We study the performance impact of the choice of two hyperparameters that are integral to our algorithm: the parameter $R$ for the \textit{topR} operator, and the lower similarity bound used in computing aggregation weights.
The parameter $R$ describes the proportion of each model layer to be used by our metric in calculating similarity (see Equations \ref{eq:topr} and \ref{eq:similarity-metric}, respectively, for the definitions of the $topR$ operator and our similarity metric). For an intuition on the meaning of $R$, consider that $R$ describes the proportion of model parameters to be compared when rating the similarity of two clients. The higher $R$ is, the more parameters are taken into account; for $R=1$, all model parameters are compared, recovering the ``normal'' cosine similarity metric.
The minimum similarity bound is used during the weighted aggregation step to include only models exceeding a given similarity value in the aggregation.\\

The results nevertheless support the use of this modified similarity metric: the use of a well-tuned $topR$ parameter is shown to improve performance compared to the standard metric that is recovered with $R=0$ in four out of five studied environments, in some cases quite significantly. For the MO-LL environment, the highest mean scalarised client reward of $33.18$ is obtained for $R=0.2$, representing an improvement of approximately $7.6\%$ over the result of $30.84$ for $R=0$; the most successful configuration on the MO-LLc.~environment leads to a circa $4.2\%$ higher mean reward. For the DST environment, the improvement is even greater: from $2.52$ for $R=0$ to $3.76$ for $R=0.8$, an increase of roughly $49\%$.

The relative improvement for the MO-HC environment is somewhat lower, but it does exist: a $topR$ parameter of $0.4$ shows roughly a $2.6\%$ improvement over the plain metric, from $2967.83$ to $3043.55$. 
From these observations, we conclude that a modification of the plain cosine similarity metric for quantifying model similarity does have promise; however, the high variance we observe during the sensitivity across all environments indicates that the stability of such a metric leaves room for improvement.\\
The results for the minimum-similarity threshold (see Figure~\ref{fig:minsim-sensitivity}) show commonalities across all five environments, suggesting that thresholds lower than $0$ are remarkably beneficial to the learning outcome of our algorithm: the relative improvement in mean scalarised client reward between a similarity threshold of $0$ and the optimal discovered value ranges from $13\%$ for the MO-LLc.~environment with threshold $-0.4$ to a full $296.8\%$ improvement for the DST environment with threshold $-0.6$. Indeed, it appears that this pattern is in general quite stable, so fixing the minimum-similarity threshold to $-1$ even without tuning this parameter is likely to lead to good results.\\
Though counter-intuitive at first glance, given the geometric interpretation of cosine similarity, this outcome is quite reasonable in the context of our algorithm. Firstly, we note that the purpose of our algorithm's clustering strategy is to group those nodes into clusters that can benefit from collaboration. Hence, improved results for a lower minimum-similarity threshold indicate that this grouping is successful, as even relatively dissimilar clients inside the same cluster improve with collaboration. Secondly, the fact that clients train personalised and therefore different models means that some dissimilarity is induced by definition of the metric, through the choice of the cluster-mean model as a reference point in computing the cosine similarity. 
\subsection{Validation of clustering strategy}\label{subsec:evaluation-clustering}

\begin{figure}
    \centering
    \includegraphics[width=.4\columnwidth]{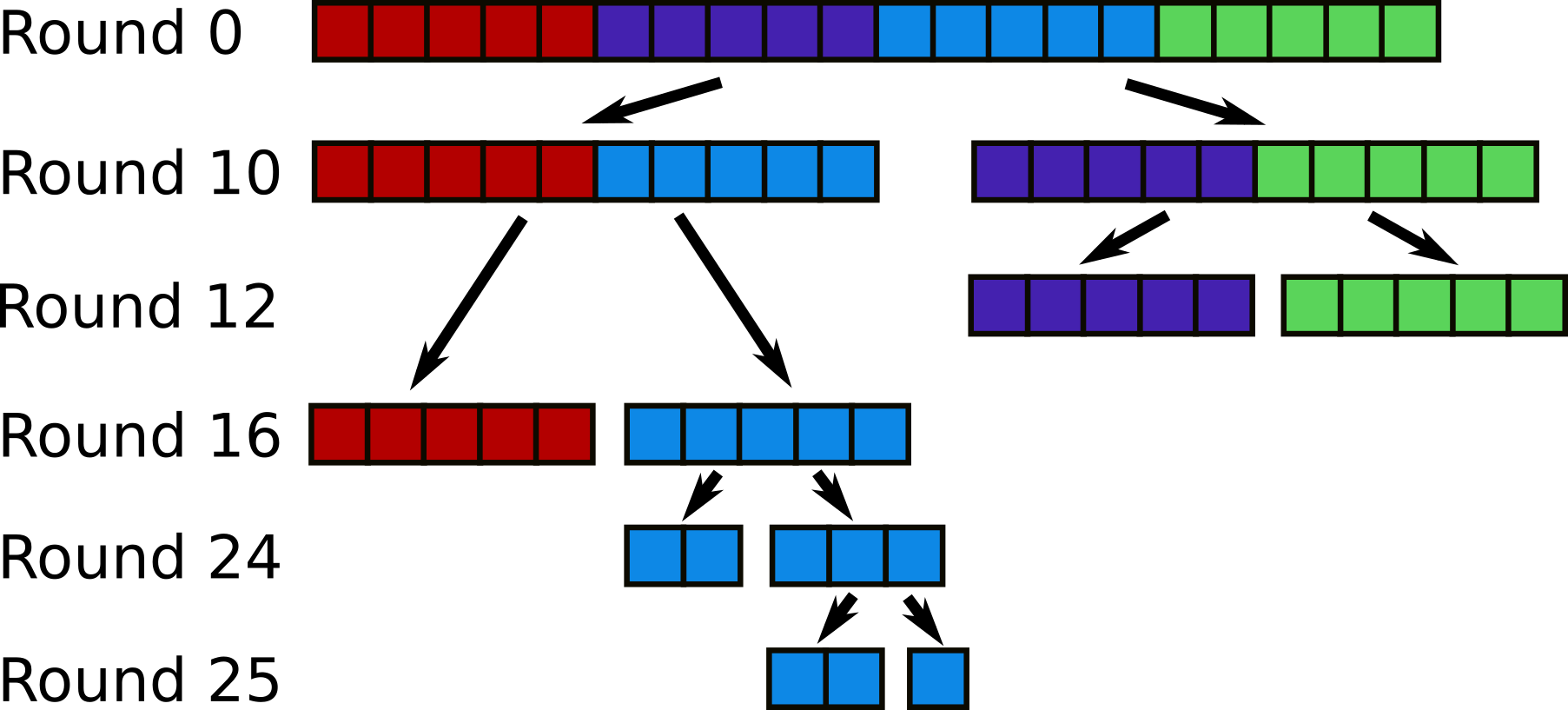}
    \caption{Cluster states at different training stages during a single experimental run on the MO-LL environment. Clients with the same preferences are represented as boxes of the same colour.
}
    \Description{Clients are separated progressively in repeated clustering steps, with clients of the same colour grouped together.}
    \label{fig:cval-states-illustration}
    \vspace{2em}
\end{figure}

We validate the clustering strategy on all three environments by running FedPref on several artificially constructed configurations where multiple clients share the same preferences. We construct two types of configurations: one where preferences are distributed among equal numbers of clients each ($4$ distinct preference weights, with each preference weight held by $5$ clients), and one where the number of clients varies for each preference ($4$ distinct preference weights, held by $2, 3, 6$ and $9$ clients, respectively). We observe how well the clustering algorithm groups similar clients, and we study how the similarity between clients develops during training. Due to scope constraints, we present only one such configuration here; visualisations of other configurations are included in the supplementary material.\\
Figure~\ref{fig:cval-heatmaps-ll-eq} shows client similarity values at selected steps of the training process on the MO-Lunar Lander environment under the balanced preference distribution. (Note that, for ease of visualisation, clients with the same preference weight are grouped together by index.) The evolution of client clusters over the duration of the training process is visualised in Figure~\ref{fig:cval-states-illustration}. In this illustration, clients with the same preferences are represented as boxes of the same colour; boxes that form a connected bar represent a cluster. Note that the relative positioning of boxes does not necessarily correspond to client index.
We see in the visualisation that client models initially develop individually, but varying preference similarity is already reflected in the model similarity computed by our metric. At the earliest visualised stage (after five aggregation steps; left-most image in Figure~\ref{fig:cval-heatmaps-ll-eq}), all clients are still grouped together in a single cluster. Nevertheless, the weighted aggregation strategy gives individual models the freedom to develop separately, yet also appears to be successful in encouraging the aggregation of clients with the same objectives.\\
In the second image, at an intermediate stage of the training process, a split into multiple clusters has occurred. The grouping of clients with the same preferences is preserved across experimental runs, but multiple groups of such clients continue to collaborate at this training stage, with different groups clustered together in different experimental runs. In the instance visualised here, clients $1-5$ and $11-15$ are all contained in the same cluster.\\ In the final image, close to the end of the training phase, we observe that the similarity of the models obtained by clients with the same preferences is very high, while the similarity to other models appears lower than before. This indicates that these clients have been separated into individual clusters, and that the personalised models within these clusters are converging; Figure~\ref{fig:cval-states-illustration} confirms this impression. In this visualisation of clustering states, we see that all sets of clients with the same preferences have been separated correctly by the end of aggregation round $16$. 
\begin{figure}[ht]
     \centering
         \includegraphics[height=3cm]{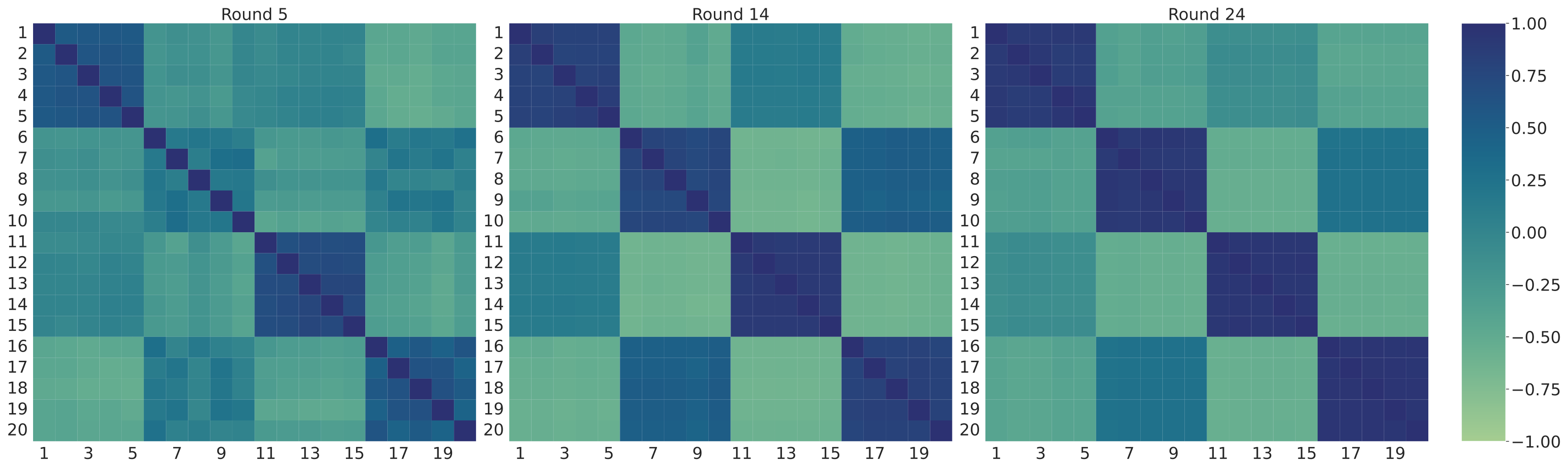}
        \caption{Mutual client similarity at different stages during a single experimental run on the MO-LL environment. Left to right: client similarities after aggregation round $5, 14$ and $26$ of $28$, respectively.} %
        \label{fig:cval-heatmaps-ll-eq}
        \Description{A heatmap of client similarities.Four balanced clusters are apparent, becoming increasingly delineated as training progresses.}
        \vspace{2em}
\end{figure}

\section{A different point of view: multi-objective evaluation}\label{sec:evaluation-multiobjective}
Up to this point, we have considered the preference heterogeneity problem only from the traditional client-level FL viewpoint, with the sole aim of optimising the performance of each client according to its preferences. However, we note that a system-level view is also particularly relevant in this preference-heterogeneous setting, and should be considered in judging the performance of any algorithm designed to solve it. %
Recall that in this setting, clients are solving a problem with multiple objectives, with each client's preferences describing the relative importance of each objective. The general aim of modelling a problem with multiple objectives is to capture the inherent complexity of the real world, allowing for the consideration of different, potentially conflicting influences. In solving the problem for different preference weights, the assumption is that these preferences will be met in a meaningful way, i.e. that different preference distributions will in fact lead to substantially different trade-off solutions. 
Any algorithmic approach that fails to do so arguably largely invalidates the premise for using multiple objectives in the first place. Our algorithm endeavours to meet this underlying expectation, allowing client models to diverge even while aggregating related models. \\
To evaluate this aspect of algorithmic performance, we propose to use standard multi-objective metrics from the field of multi-objective optimisation (MOO) to evaluate the set of solutions generated by all federated clients under the different preference distributions. In the remainder of this section, we will first introduce these metrics, followed by the evaluation and discussion of our experiments using these metrics.

\subsection{Background: multi-objective metrics}
\begin{figure}[ht]
  \centering
  \includegraphics[width=.8\linewidth]{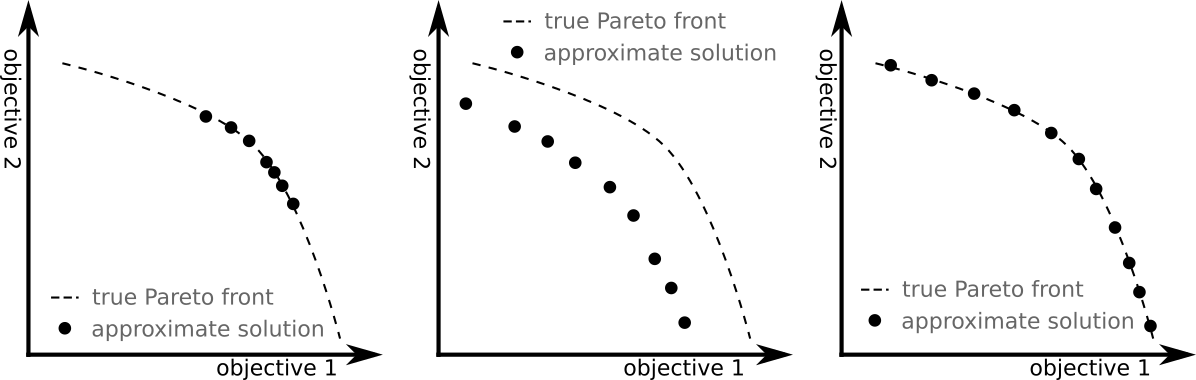}
  \caption{Illustration of diversity and convergence in a multi-objective context. Left: a solution set with good convergence and poor diversity; centre: poor convergence and good diversity; right: good convergence and good diversity. Both objectives are being maximised in these examples.}
  \label{fig:diversity-convergence}
  \Description{Left: many solutions grouped close together on the true Pareto front. Centre: well-distributed solutions, but located far away from the true Pareto front. Right: well-distributed solutions lying on the true Pareto front.}
\end{figure}

A standard method in the field of multi-objective optimisation is to study the subset of optimal trade-off solutions, or \textit{Pareto front}, found by the algorithm \cite{ngatchou_pareto_2005}. Intuitively, given a set of multi-objective solutions $\mathcal{S}$, a point $s$ lies on the Pareto front iff the value of one objective in $s$ cannot be improved without reducing that of another.\\
Many metrics designed to measure characteristics of a multi-objective solution set have been proposed in the literature, with most focused on quantifying the \textit{diversity} and \textit{convergence} of solutions \cite{riquelme_performance_2015}. The diversity of a solution set describes the distribution of solutions in space - it is often considered more desirable to find solutions that are different from one another, in order to present a greater range of options to an end user selecting among the different possible trade-offs. The notion of convergence refers to the closeness of the obtained solutions to the underlying 'true' Pareto front - the closer the better. A set of multi-objective solutions is generally considered to be of high quality iff it has both a high diversity and high convergence -- only one of these characteristics is not sufficient, as illustrated in  Figure~\ref{fig:diversity-convergence}. A set of solutions with high diversity and low convergence may offer a large selection of trade-off solutions, but all solutions are far removed from optimality. Conversely, a set of solutions with high convergence and low diversity may contain solutions that are close to optimal, but fail to cover the range of possible trade-offs. Only a set of solutions with both high convergence and high diversity yields a full range of near-optimal trade-off solutions.\\
In practice, the notions of diversity and convergence are difficult to quantify for the general case, in part because the 'true' Pareto front is often unknown; various surrogate metrics have been proposed in the MOO literature. In this work, we focus on four common state-of-the-art metrics: the hypervolume \cite{zitzler_multiobjective_1999}, sparsity \cite{xu_prediction_2020}, inverted generational distance (IGD) \cite{coello_solving_2005}, and cardinality of the Pareto front. \\
\textbf{Hypervolume $(\uparrow)$.} The hypervolume metric is computed as the combined volume of the set of hypercubes spanned by the solutions on the Pareto front and a pre-defined minimal reference point. This metric captures both diversity and convergence: more diverse solutions on the Pareto front generate hypercubes with less mutual overlap, increasing the overall hypervolume, while more optimal individual solutions are further removed from the reference point, leading to a greater volume of their respective hypercubes. However, this metric suffers from some weaknesses that make it unfit to be used in isolation, e.g. small numbers of solutions that are near-optimal for a particular trade-off have the potential to dominate a more diverse set of different solutions. Therefore, a thorough analysis requires the use of other metrics in combination with the hypervolume metric.\\
\textbf{Sparsity $(\downarrow )$.} The sparsity metric measures the mutual distance between solutions on the Pareto front; as such, it describes the diversity of a set of solutions. This metric, too, is limited when used in isolation: aside from not capturing convergence, it is also influenced by the number of solutions involved. A set of only two relatively close solutions will return a lower sparsity score than the same set with another, more distant solution added. This characteristic suggests the use of the cardinality metric to support a sparsity analysis.\\
\textbf{Inverted Generational Distance (IGD) $(\downarrow )$.} This metric quantifies the convergence in terms of the distance of the Pareto front of the solution set to the 'true' Pareto front. As the true Pareto front of a problem is rarely known in practice, it is usually approximated as the Pareto front obtained when combining all solutions generated during an experimental campaign.\\
\textbf{Cardinality $(\uparrow )$.} Cardinality is the number of solutions in a given set that lie on the Pareto front. This metric is used in support of other performance indicators such as the sparsity metric.

\subsection{Experimental evaluation using multi-objective metrics}
We re-evaluate the same experiments presented and discussed in Section~\ref{sec:evaluation} under multi-objective aspects. To the best of our knowledge, the performance of federated algorithms under multi-objective aspects has not been discussed before; hence there are no obvious additional baselines to consider for comparison. We note that in this setting, a main challenge for federated system can be expected to lie in the need for solution diversity, as client models trained in federation need to achieve some level of convergence to effectively exchange information. Indeed, the classical FL baseline of a system of clients without communication continues to provide a relevant challenge here, as non-cooperating clients might be expected to achieve a high diversity of solutions by default.\\
The four multi-objective metrics are reported in Table~\ref{table:hypervolume-results} (Hypervolume), Table~\ref{table:igd-results} (IGD), Table~\ref{table:sparsity-results} (sparsity) and Table~\ref{table:cardinality-results} (cardinality).\\

\textbf{Equidistantly distributed preference weights.} 
We first observe that a variant of the FedPref algorithm generates or matches the highest cardinality, i.e. the number of solution on the Pareto front, in three out of five experimental environments. For example, in the Deep-Sea Treasure environment, the mean cardinality achieved by FedPref is $8.0$, meaning that of the $20$ clients per federated system, on average $8$ clients find a distinct optimal trade-off solution. This stands in marked contrast to the non-PFL algorithms and CFL, which find only $1$ and $1.2$ such solutions, respectively, or $\leq 4.0$ and $2.2$ for the corresponding fine-tuned variants. These algorithms achieve higher cardinalities in environments with more dense solution spaces, most notably MO-LL and MO-LLc. However, these higher cardinalities are likely caused mainly by statistical differences in model evaluations. The fine-tuning variants of these algorithms also yield higher cardinality scores in many cases, as might be expected; yet these remain generally lower than the highest scores.\\
In the Deterministic Minecart environment, the mean cardinality of $3.2$ obtained by the FedPref+FT algorithm is only beaten by the value of $3.3$ of the CFL+FT algorithm and the non-federated baseline, and only by a small margin. These results match our observations of the success of each algorithm when considering average scalarised client performance in Section~\ref{sec:evaluation}. \\
The corresponding hypervolume results support the overall impression given by the examination of cardinality values: higher cardinality values correspond to the higher hypervolume values achieved for each experimental environment, though the overall ranking does not translate exactly. For example, in the MO-Halfcheetah environment, where the FedPref-FT and FedProx+FT algorithms obtain cardinality values of $6.9$ and $7.0$, respectively, the FedPref-FT algorithm nevertheless yields a higher average hypervolume, indicating that a solution set with higher diversity or convergence was found. A comparison of sparsity and IGD results suggests the former explanation.\\
Another notable exception to this observed correlation between high cardinality and high hypervolume occurs when comparing algorithms with their fine-tuning variants. Although the fine-tuning step generally appears to yield more distinct optimal solutions, this does not always translate to equally great improvements in hypervolume. This is likely because new solutions obtained by fine-tuning do not diverge too far from others.\\

\textbf{Uniformly distributed preference weights.}
On average, more than half of all federated clients solving the MO-LL, MO-HC and MO-LLc.~ environments with a variant of the FedPref algorithm find a distinct optimal trade-off solution in all experimental configurations. For the DST environment, which has only $10$ discrete solutions in total, the FedPref+FT algorithm consistently leads to the identification of more than $70\%$ of possible solutions. As for the previous distribution, a variant of FedPref achieves the highest cardinality for the MO-LL, DST, and MO-LLc.~environments, including the highest overall cardinality of $13.7$ for the MO-Lunar Lander. On the other two environments, the number of trade-off solutions found for FedPref is again only slightly lower than the highest achieved by any algorithm. The effects of adding a fine-tuning step to the various algorithms appear similar to the those discussed for the equidistant preference distribution.
Similarly to previous observations, the number of optimal trade-off solutions found appears to translate well to higher observed hypervolume values, with a variant of FedPref accomplishing the highest rank in the same three environments as for the cardinality.
This indicates that the set of solutions found by the federated system executing FedPref does have a high level of diversity, and that the comparably high sparsity values we observe for these same configurations are likely due to greater spread of optimal solutions. \\
As a final observation, we note that in many cases, the FedPref algorithm yields more distinct optimal trade-off solutions than the non-federated baseline, suggesting that PFL can assist in effectively exploring the solution space and finding an even more diverse solution set than non-collaborative clients.\\

\textbf{Gaussian-distributed preference weights.}
For this distribution, finding a diverse set of solutions appears to present a particular challenge, probably because preferences are more likely to be more similar here, allowing clients to jointly exploit local optima more successfully. 
With respect to the cardinality metric, we observe a similar pattern as before, with a variant of the FedPref algorithm again obtaining or matching the highest cardinality values in three out of five environments, and nearing the highest value in all others. However, the hypervolume results are less decisive here than for the other two distributions: FedPref does reach the highest hypervolume value on only three out of five environments. However, on the MO-HC environment, the highest hypervolume score is reached by the non-federated baseline, with FedPref following in second place, and this ranking is only just reversed on the DST environment. It is likely that in this preference distribution setting, the federated clients succeed more readily in collaborating, leading to higher individual results, as seen in the analysis in Section\ref{subsec:evaluation-main-baselines}. The down-side of this enhanced collaboration could be a loss of diversity, as indicated by the slightly higher hypervolume values accomplished by the non-federated baseline, where clients do not collaborate at all, in this case. Nonetheless, consideration of the corresponding sparsity and IGD values shows that variants of FedPref yield lower results than the non-federated baseline for both metrics. This indicates that the set of results found by FedPref is overall more evenly distributed.

In conclusion, in analysing multi-objective metrics across all five environments, we observe that the FedPref algorithm leads most consistently of all federated algorithms to a diverse set of good trade-off solutions. A general challenge from a multi-objective viewpoint is the lack of solution diversity brought on by the aggregation of client models. This is most evident in the results of the FedAvg and FedProx algorithms, which find generally low numbers of distinct optimal trade-off solutions, even with the addition of a fine-tuning phase. Despite the relatively high mean scalarised reward we have observed in the client-level evaluation of the previous section, these results are arguably not very satisfactory from a multi-objective point of view of the system. The compared personalised FL algorithms, CFL and MaTFL, generally perform somewhat better, with variants of CFL in particular achieving a relatively high solution diversity at times. This behaviour may stem from the unbalanced clustering strategy of CFL, which we have previously remarked upon in Section~\ref{subsec:evaluation-main-baselines}. However, unlike for FedPref, the performance of CFL is not consistent across preference distributions and environments. As with the client-level evaluation, FedPref proves to be the most adaptable federated algorithm of all those evaluated, both with respect to different types of preference heterogeneity and multi-objective problems with different characteristics.
\begin{table*}
    \caption{Hypervolume$(\uparrow)$ metric for multi-objective solutions obtained by our proposed FedPref algorithm, compared to MaTFL, CFL, FedProx, FedAvg and individual learning without cooperation. Where indicated in the header, both the main value and the standard deviation value have been divided by the given power.}\label{table:hypervolume-results}
\begin{tabular}{lllllll} 
 & & MO-LL~$(\cdot 10^{7})$& DMC& DST~$(\cdot 10^{1})$& MO-HC~$(\cdot 10^{4})$& MO-LLc.~$(\cdot 10^{7})$\\ 
\toprule 
\multirow{10}{*}{Dirichlet} & No comm.& $175.54~\sigma 7.1$& $198.01~\sigma 28.2$& $204.47~\sigma 20.1$& $429.78~\sigma 45.7$& $77.20~\sigma 10.9$\\
 & FedAvg& $176.53~\sigma 7.6$& $90.12~\sigma 60.3$& $88.89~\sigma 72.6$& $242.72~\sigma 67.5$& $91.20~\sigma 5.1$\\
 & FedAvg+FT& $169.13~\sigma 20.8$& $142.99~\sigma 13.9$& $134.48~\sigma 53.6$& $410.93~\sigma 23.2$& $94.55~\sigma 9.7$\\
 & FedProx& $172.67~\sigma 13.5$& $90.13~\sigma 60.3$& $103.10~\sigma 67.5$& $235.33~\sigma 65.8$& $90.11~\sigma 6.4$\\
 & FedProx+FT& $169.05~\sigma 24.5$& $151.93~\sigma 18.1$& $147.67~\sigma 53.5$& $419.97~\sigma 19.0$& $92.10~\sigma 7.8$\\
 & CFL& $178.41~\sigma 14.8$& \textbf{209.29}$~\sigma 15.7$& $131.44~\sigma 74.3$& $406.33~\sigma 37.1$& $94.76~\sigma 9.7$\\
 & CFL+FT& $173.04~\sigma 14.4$& $206.50~\sigma 24.3$& $172.20~\sigma 26.7$& $395.06~\sigma 39.6$& $92.26~\sigma 7.8$\\
 & MaTFL& $163.81~\sigma 14.6$& $86.38~\sigma 36.7$& $214.25~\sigma 18.3$& $299.03~\sigma 49.5$& $80.16~\sigma 7.8$\\
 & FedPref+FT (ours)& $193.94~\sigma 7.5$& $202.79~\sigma 29.7$& \textbf{220.36}$~\sigma 10.3$& $421.43~\sigma 39.5$& $100.04~\sigma 5.5$\\
 & FedPref-FT (ours)& \textbf{201.97}$~\sigma 7.0$& $203.74~\sigma 13.2$& $218.04~\sigma 18.3$& \textbf{435.87}$~\sigma 36.5$& \textbf{101.57}$~\sigma 8.8$\\
\toprule 
\multirow{10}{*}{Equidistant} & No comm.& $156.00~\sigma 25.3$& \textbf{204.43}$~\sigma 13.2$& $216.56~\sigma 13.8$& $424.79~\sigma 53.5$& $53.73~\sigma 13.0$\\
 & FedAvg& $69.10~\sigma 48.6$& $29.80~\sigma 0.2$& $89.88~\sigma 73.3$& $235.81~\sigma 31.0$& $78.96~\sigma 9.6$\\
 & FedAvg+FT& $85.78~\sigma 21.2$& $138.32~\sigma 40.9$& $157.36~\sigma 3.6$& $426.24~\sigma 25.8$& $54.83~\sigma 13.5$\\
 & FedProx& $46.44~\sigma 38.5$& $29.85~\sigma 0.1$& $74.04~\sigma 74.0$& $247.38~\sigma 32.7$& $77.25~\sigma 10.9$\\
 & FedProx+FT& $96.54~\sigma 13.4$& $137.51~\sigma 15.4$& $161.75~\sigma 23.9$& $426.88~\sigma 31.8$& $50.86~\sigma 14.5$\\
 & CFL& $29.76~\sigma 28.3$& $183.64~\sigma 33.4$& $45.29~\sigma 69.0$& $414.97~\sigma 31.5$& $80.58~\sigma 17.8$\\
 & CFL+FT& $130.92~\sigma 11.9$& $202.94~\sigma 14.0$& $171.12~\sigma 24.4$& $427.94~\sigma 39.4$& $65.00~\sigma 16.8$\\
 & MaTFL& $150.15~\sigma 25.9$& $79.08~\sigma 36.0$& $200.99~\sigma 25.8$& $290.91~\sigma 41.9$& $58.93~\sigma 12.8$\\
 & FedPref+FT (ours)& $166.37~\sigma 19.1$& $197.45~\sigma 27.6$& $222.37~\sigma 8.5$& \textbf{442.63}$~\sigma 18.4$& $74.28~\sigma 5.8$\\
 & FedPref-FT (ours)& \textbf{182.90}$~\sigma 11.4$& $199.28~\sigma 8.4$& \textbf{225.23}$~\sigma 4.9$& $441.17~\sigma 27.3$& \textbf{101.04}$~\sigma 6.7$\\
\toprule 
\multirow{10}{*}{Gaussian} & No comm.& $176.61~\sigma 11.7$& $201.23~\sigma 11.8$& $221.37~\sigma 8.1$& \textbf{453.61}$~\sigma 43.9$& $83.10~\sigma 7.6$\\
 & FedAvg& $178.50~\sigma 8.2$& $92.04~\sigma 58.6$& $99.25~\sigma 65.4$& $220.48~\sigma 95.6$& $90.14~\sigma 6.3$\\
 & FedAvg+FT& $177.79~\sigma 13.2$& $126.71~\sigma 43.0$& $149.97~\sigma 58.5$& $402.24~\sigma 39.8$& $98.86~\sigma 4.4$\\
 & FedProx& $181.00~\sigma 10.7$& $69.89~\sigma 53.2$& $133.89~\sigma 44.7$& $222.09~\sigma 83.4$& $89.06~\sigma 6.6$\\
 & FedProx+FT& $174.14~\sigma 11.5$& $133.09~\sigma 31.0$& $143.09~\sigma 15.2$& $398.41~\sigma 41.5$& $96.28~\sigma 4.8$\\
 & CFL& $187.52~\sigma 6.2$& $204.03~\sigma 23.4$& $97.56~\sigma 82.0$& $413.57~\sigma 28.5$& $91.74~\sigma 5.2$\\
 & CFL+FT& $181.28~\sigma 7.8$& \textbf{215.13}$~\sigma 13.3$& $159.97~\sigma 19.9$& $437.62~\sigma 27.5$& $92.08~\sigma 10.9$\\
 & MaTFL& $167.94~\sigma 16.8$& $101.00~\sigma 46.9$& $201.87~\sigma 27.8$& $280.87~\sigma 45.7$& $76.94~\sigma 12.2$\\
 & FedPref+FT (ours)& $197.28~\sigma 8.4$& $195.67~\sigma 38.2$& $215.89~\sigma 10.1$& $420.45~\sigma 47.7$& \textbf{101.23}$~\sigma 7.9$\\
 & FedPref-FT (ours)& \textbf{202.06}$~\sigma 4.9$& $203.88~\sigma 13.4$& \textbf{221.44}$~\sigma 10.7$& $450.20~\sigma 25.0$& $100.21~\sigma 7.0$\\
\bottomrule
\end{tabular}
\end{table*}

\begin{table*}
    \caption{Sparsity$(\downarrow)$ metric for multi-objective solutions obtained by our proposed FedPref algorithm, compared to MaTFL, CFL, FedProx, FedAvg and individual learning without cooperation. Lower sparsity means that the mutual distance between solutions obtained by the algorithm is lower. The metric has zero-value by definition if there is only one solution on the Pareto front.}\label{table:sparsity-results}
\begin{tabular}{lllllll} 
\toprule 
 & & MO-LL~$(\cdot 10^{2})$& DMC& DST& MO-HC~$(\cdot 10^{4})$& MO-LLc.~$(\cdot 10^{1})$\\ 
\toprule 
\multirow{10}{*}{Dirichlet} & No comm.& $68.97~\sigma 94.1$& $2.00~\sigma 0.8$& $27.60~\sigma 8.6$& $488.85~\sigma 318.4$& $246.92~\sigma 95.3$\\
 & FedAvg& $3.29~\sigma 2.5$& \textbf{0.00}$~\sigma 0.0$& \textbf{0.00}$~\sigma 0.0$& $7.75~\sigma 18.7$& \textbf{27.71}$~\sigma 20.0$\\
 & FedAvg+FT& $64.07~\sigma 60.6$& $0.01~\sigma 0.0$& $9.42~\sigma 6.5$& $296.29~\sigma 104.7$& $477.85~\sigma 302.2$\\
 & FedProx& $2.43~\sigma 1.6$& $0.00~\sigma 0.0$& $0.00~\sigma 0.0$& \textbf{2.62}$~\sigma 5.1$& $31.37~\sigma 27.2$\\
 & FedProx+FT& $59.04~\sigma 47.5$& $0.23~\sigma 0.4$& $16.70~\sigma 16.0$& $278.82~\sigma 134.1$& $338.70~\sigma 269.5$\\
 & CFL& \textbf{1.47}$~\sigma 1.0$& $1.03~\sigma 0.5$& $34.29~\sigma 72.5$& $341.84~\sigma 208.5$& $221.40~\sigma 93.5$\\
 & CFL+FT& $85.21~\sigma 135.7$& $0.94~\sigma 0.2$& $42.52~\sigma 71.0$& $333.20~\sigma 145.3$& $334.66~\sigma 298.9$\\
 & MaTFL& $25.65~\sigma 13.2$& $1.46~\sigma 1.5$& $57.06~\sigma 29.0$& $380.86~\sigma 398.7$& $247.25~\sigma 194.1$\\
 & FedPref+FT (ours)& $19.35~\sigma 8.6$& $1.83~\sigma 1.0$& $28.00~\sigma 7.6$& $408.45~\sigma 167.6$& $194.61~\sigma 88.7$\\
 & FedPref-FT (ours)& $19.38~\sigma 17.0$& $1.88~\sigma 0.8$& $33.07~\sigma 7.9$& $517.10~\sigma 377.2$& $92.45~\sigma 68.8$\\
\toprule 
\multirow{10}{*}{Equidistant} & No comm.& $200.89~\sigma 307.6$& $2.63~\sigma 2.1$& $25.12~\sigma 5.2$& $301.92~\sigma 126.3$& $513.09~\sigma 189.3$\\
 & FedAvg& $8.74~\sigma 3.5$& \textbf{0.00}$~\sigma 0.0$& \textbf{0.00}$~\sigma 0.0$& \textbf{3.43}$~\sigma 7.9$& $31.79~\sigma 9.5$\\
 & FedAvg+FT& $384.42~\sigma 162.3$& $0.47~\sigma 1.3$& $10.25~\sigma 3.0$& $250.54~\sigma 79.1$& $991.11~\sigma 659.6$\\
 & FedProx& $5.31~\sigma 1.9$& $0.00~\sigma 0.0$& $0.00~\sigma 0.0$& $33.23~\sigma 60.2$& \textbf{24.54}$~\sigma 25.5$\\
 & FedProx+FT& $278.97~\sigma 159.3$& $0.02~\sigma 0.1$& $17.59~\sigma 17.6$& $288.65~\sigma 144.8$& $814.88~\sigma 382.5$\\
 & CFL& \textbf{4.29}$~\sigma 2.6$& $1.66~\sigma 0.8$& $0.44~\sigma 0.9$& $338.76~\sigma 139.4$& $512.27~\sigma 391.4$\\
 & CFL+FT& $626.89~\sigma 420.1$& $1.77~\sigma 0.7$& $20.79~\sigma 37.7$& $288.57~\sigma 99.2$& $606.37~\sigma 383.7$\\
 & MaTFL& $121.98~\sigma 111.8$& $1.12~\sigma 1.4$& $43.09~\sigma 14.6$& $402.82~\sigma 327.8$& $630.16~\sigma 304.6$\\
 & FedPref+FT (ours)& $53.02~\sigma 29.8$& $2.04~\sigma 0.7$& $24.42~\sigma 5.1$& $336.65~\sigma 99.4$& $728.32~\sigma 775.4$\\
 & FedPref-FT (ours)& $25.82~\sigma 11.3$& $2.35~\sigma 0.5$& $23.85~\sigma 4.8$& $337.47~\sigma 127.0$& $263.87~\sigma 179.2$\\
\toprule 
\multirow{10}{*}{Gaussian} & No comm.& $66.56~\sigma 105.2$& $2.03~\sigma 0.7$& $33.23~\sigma 10.8$& $490.86~\sigma 415.2$& $207.62~\sigma 154.5$\\
 & FedAvg& $1.87~\sigma 0.7$& \textbf{0.00}$~\sigma 0.0$& \textbf{0.00}$~\sigma 0.0$& $2.46~\sigma 5.4$& $21.98~\sigma 9.1$\\
 & FedAvg+FT& $86.31~\sigma 87.5$& $0.12~\sigma 0.3$& $32.64~\sigma 53.0$& $225.45~\sigma 107.7$& $257.47~\sigma 164.4$\\
 & FedProx& $2.70~\sigma 2.2$& $0.00~\sigma 0.0$& $0.00~\sigma 0.0$& \textbf{0.74}$~\sigma 1.1$& \textbf{20.03}$~\sigma 12.0$\\
 & FedProx+FT& $82.10~\sigma 84.1$& $0.03~\sigma 0.1$& $12.41~\sigma 3.3$& $238.79~\sigma 124.2$& $266.19~\sigma 179.9$\\
 & CFL& \textbf{0.95}$~\sigma 0.9$& $1.28~\sigma 1.1$& $29.23~\sigma 87.0$& $443.66~\sigma 323.8$& $164.09~\sigma 65.1$\\
 & CFL+FT& $24.34~\sigma 11.2$& $0.86~\sigma 0.1$& $25.57~\sigma 72.5$& $247.99~\sigma 152.1$& $256.48~\sigma 201.6$\\
 & MaTFL& $132.91~\sigma 126.5$& $1.68~\sigma 1.4$& $40.85~\sigma 17.0$& $327.88~\sigma 182.6$& $212.49~\sigma 91.3$\\
 & FedPref+FT (ours)& $21.57~\sigma 11.8$& $1.76~\sigma 0.9$& $25.18~\sigma 3.7$& $284.49~\sigma 141.6$& $259.21~\sigma 167.8$\\
 & FedPref-FT (ours)& $23.70~\sigma 20.4$& $1.88~\sigma 0.8$& $27.61~\sigma 15.6$& $260.01~\sigma 108.0$& $123.25~\sigma 60.7$\\
\bottomrule
\end{tabular}
\end{table*}

\begin{table*}
    \caption{Inverted Generational Distance (IGD, $\downarrow$) metric for multi-objective solutions obtained by our proposed FedPref algorithm, compared to MaTFL, CFL, FedProx, FedAvg and individual learning without cooperation. Lower IGD means that the obtained solution set is closer to the ``true'' set of trade-off solutions.}\label{table:igd-results}
\begin{tabular}{lllllll} 
\toprule 
 & & MO-LL& DMC& DST& MO-HC& MO-LLc.\\ 
\toprule 
\multirow{10}{*}{Dirichlet} & No comm.& $45.42~\sigma 4.4$& $0.41~\sigma 0.2$& $1.42~\sigma 1.0$& $851.67~\sigma 307.4$& $55.43~\sigma 8.1$\\
 & FedAvg& $85.13~\sigma 13.0$& $10.01~\sigma 9.4$& $41.19~\sigma 41.5$& $2737.53~\sigma 684.5$& $68.28~\sigma 6.3$\\
 & FedAvg+FT& $57.25~\sigma 9.9$& $0.81~\sigma 0.2$& $14.52~\sigma 25.9$& $733.68~\sigma 210.4$& $60.43~\sigma 14.2$\\
 & FedProx& $93.19~\sigma 15.5$& $9.99~\sigma 9.3$& $32.71~\sigma 38.8$& $2773.36~\sigma 738.3$& $68.54~\sigma 6.2$\\
 & FedProx+FT& $60.71~\sigma 11.8$& $0.73~\sigma 0.2$& $13.47~\sigma 26.2$& \textbf{731.80}$~\sigma 259.4$& $57.40~\sigma 8.5$\\
 & CFL& $106.68~\sigma 15.7$& $0.22~\sigma 0.1$& $23.57~\sigma 34.2$& $804.38~\sigma 470.8$& \textbf{47.25}$~\sigma 5.3$\\
 & CFL+FT& $51.05~\sigma 6.2$& \textbf{0.19}$~\sigma 0.1$& $6.30~\sigma 0.9$& $733.44~\sigma 213.2$& $51.04~\sigma 3.6$\\
 & MaTFL& $44.64~\sigma 4.9$& $1.01~\sigma 0.2$& $1.93~\sigma 0.8$& $973.75~\sigma 207.4$& $51.04~\sigma 5.0$\\
 & FedPref+FT (ours)& \textbf{39.22}$~\sigma 3.0$& $0.38~\sigma 0.2$& \textbf{0.77}$~\sigma 0.3$& $738.36~\sigma 183.2$& $47.86~\sigma 6.7$\\
 & FedPref-FT (ours)& $47.24~\sigma 7.5$& $0.37~\sigma 0.2$& $1.09~\sigma 0.9$& $770.74~\sigma 286.1$& $51.38~\sigma 8.1$\\
\toprule 
\multirow{10}{*}{Equidistant} & No comm.& $57.24~\sigma 9.4$& $0.40~\sigma 0.2$& $0.81~\sigma 0.4$& $539.53~\sigma 141.6$& $76.47~\sigma 14.9$\\
 & FedAvg& $236.25~\sigma 115.2$& $19.38~\sigma 0.2$& $41.13~\sigma 41.5$& $2463.58~\sigma 263.7$& $76.65~\sigma 6.5$\\
 & FedAvg+FT& $138.72~\sigma 29.5$& $2.63~\sigma 5.5$& $4.83~\sigma 0.3$& \textbf{469.81}$~\sigma 77.6$& $73.82~\sigma 15.0$\\
 & FedProx& $299.47~\sigma 121.7$& $19.33~\sigma 0.1$& $49.63~\sigma 42.3$& $2311.35~\sigma 574.5$& $83.29~\sigma 7.9$\\
 & FedProx+FT& $121.49~\sigma 16.4$& $0.91~\sigma 0.3$& $4.91~\sigma 1.2$& $542.68~\sigma 188.4$& $81.91~\sigma 13.5$\\
 & CFL& $383.86~\sigma 99.7$& $0.40~\sigma 0.2$& $66.43~\sigma 39.0$& $504.73~\sigma 131.9$& $57.72~\sigma 10.2$\\
 & CFL+FT& $87.04~\sigma 8.7$& \textbf{0.35}$~\sigma 0.1$& $6.19~\sigma 0.7$& $480.66~\sigma 51.6$& $70.09~\sigma 12.5$\\
 & MaTFL& $58.39~\sigma 4.3$& $1.04~\sigma 0.2$& $2.16~\sigma 1.1$& $931.54~\sigma 291.1$& $65.97~\sigma 11.4$\\
 & FedPref+FT (ours)& $50.17~\sigma 3.2$& $0.42~\sigma 0.2$& $0.59~\sigma 0.3$& $536.04~\sigma 96.4$& $58.35~\sigma 4.3$\\
 & FedPref-FT (ours)& \textbf{48.80}$~\sigma 6.8$& $0.46~\sigma 0.1$& \textbf{0.48}$~\sigma 0.2$& $540.33~\sigma 193.3$& \textbf{45.68}$~\sigma 3.7$\\
\toprule 
\multirow{10}{*}{Gaussian} & No comm.& $42.80~\sigma 3.4$& $0.39~\sigma 0.2$& $0.97~\sigma 0.5$& $663.02~\sigma 294.8$& $52.37~\sigma 4.2$\\
 & FedAvg& $88.64~\sigma 9.6$& $8.23~\sigma 9.2$& $32.90~\sigma 38.7$& $2853.54~\sigma 604.7$& $72.67~\sigma 4.9$\\
 & FedAvg+FT& $56.38~\sigma 11.2$& $0.89~\sigma 0.3$& $13.71~\sigma 26.1$& $817.40~\sigma 354.0$& $53.55~\sigma 5.5$\\
 & FedProx& $86.44~\sigma 17.1$& $10.17~\sigma 9.3$& $15.74~\sigma 25.4$& $2721.86~\sigma 595.6$& $69.02~\sigma 7.5$\\
 & FedProx+FT& $55.36~\sigma 5.9$& $0.86~\sigma 0.3$& $6.16~\sigma 1.4$& $816.10~\sigma 236.0$& $53.48~\sigma 7.8$\\
 & CFL& $105.42~\sigma 12.4$& $0.27~\sigma 0.2$& $40.89~\sigma 41.7$& $694.61~\sigma 258.5$& $48.59~\sigma 6.8$\\
 & CFL+FT& $47.16~\sigma 4.6$& \textbf{0.17}$~\sigma 0.1$& $6.59~\sigma 0.6$& $925.00~\sigma 816.9$& $50.40~\sigma 5.3$\\
 & MaTFL& $48.34~\sigma 6.7$& $0.95~\sigma 0.2$& $2.07~\sigma 1.4$& $919.70~\sigma 126.0$& $55.94~\sigma 7.2$\\
 & FedPref+FT (ours)& \textbf{39.60}$~\sigma 3.4$& $0.37~\sigma 0.3$& \textbf{0.71}$~\sigma 0.3$& \textbf{637.25}$~\sigma 241.1$& \textbf{46.50}$~\sigma 7.4$\\
 & FedPref-FT (ours)& $42.34~\sigma 4.0$& $0.36~\sigma 0.2$& $0.76~\sigma 0.5$& $690.35~\sigma 361.6$& $50.02~\sigma 9.0$\\
\bottomrule
\end{tabular}
\end{table*}

\begin{table*}
    \caption{Cardinality$(\uparrow)$ metric for multi-objective solutions obtained by our proposed FedPref algorithm, compared to MaTFL, CFL, FedProx, FedAvg and individual learning without cooperation. Higher cardinality means that a higher number of distinct trade-off solutions was found.}\label{table:cardinality-results}
\begin{tabular}{lllllll} 
\toprule 
 & & MO-LL& DMC& DST& MO-HC& MO-LLc.\\ 
\toprule 
\multirow{10}{*}{Dirichlet} & No comm.& $12.20~\sigma 1.2$& $3.30~\sigma 0.8$& $6.70~\sigma 1.1$& $5.40~\sigma 0.8$& $11.60~\sigma 3.0$\\
 & FedAvg& $8.30~\sigma 2.8$& $1.00~\sigma 0.0$& $1.00~\sigma 0.0$& $2.90~\sigma 1.1$& $9.60~\sigma 2.2$\\
 & FedAvg+FT& $10.20~\sigma 1.3$& $1.10~\sigma 0.3$& $3.00~\sigma 1.6$& $6.50~\sigma 1.3$& $9.40~\sigma 2.2$\\
 & FedProx& $9.30~\sigma 2.4$& $1.00~\sigma 0.0$& $1.00~\sigma 0.0$& $2.90~\sigma 0.5$& $10.20~\sigma 2.6$\\
 & FedProx+FT& $10.70~\sigma 2.0$& $1.40~\sigma 0.5$& $3.40~\sigma 0.9$& \textbf{6.80}$~\sigma 1.3$& $9.20~\sigma 2.1$\\
 & CFL& $7.90~\sigma 1.8$& \textbf{3.60}$~\sigma 0.7$& $2.00~\sigma 1.3$& $6.00~\sigma 0.9$& $12.00~\sigma 2.9$\\
 & CFL+FT& $10.10~\sigma 1.9$& $3.60~\sigma 0.7$& $1.90~\sigma 0.7$& $6.00~\sigma 1.1$& $11.20~\sigma 2.2$\\
 & MaTFL& $11.20~\sigma 1.6$& $1.50~\sigma 0.5$& $5.70~\sigma 0.9$& $5.00~\sigma 1.2$& $12.10~\sigma 1.1$\\
 & FedPref+FT (ours)& \textbf{13.70}$~\sigma 1.8$& $3.40~\sigma 0.7$& \textbf{7.40}$~\sigma 1.0$& $5.80~\sigma 1.2$& $12.50~\sigma 2.1$\\
 & FedPref-FT (ours)& $13.30~\sigma 3.1$& $3.30~\sigma 0.5$& $6.80~\sigma 1.1$& $5.90~\sigma 1.3$& \textbf{12.70}$~\sigma 2.7$\\
\toprule 
\multirow{10}{*}{Equidistant} & No comm.& $9.30~\sigma 1.3$& \textbf{3.30}$~\sigma 0.5$& $7.40~\sigma 1.0$& $6.40~\sigma 1.1$& $10.80~\sigma 1.5$\\
 & FedAvg& $9.30~\sigma 3.4$& $1.00~\sigma 0.0$& $1.00~\sigma 0.0$& $2.90~\sigma 1.1$& $8.70~\sigma 2.4$\\
 & FedAvg+FT& $7.20~\sigma 1.5$& $1.20~\sigma 0.4$& $4.00~\sigma 0.4$& $6.80~\sigma 1.2$& $9.90~\sigma 1.4$\\
 & FedProx& $9.90~\sigma 3.2$& $1.00~\sigma 0.0$& $1.00~\sigma 0.0$& $2.70~\sigma 0.9$& $9.40~\sigma 2.9$\\
 & FedProx+FT& $8.40~\sigma 1.9$& $1.10~\sigma 0.3$& $3.80~\sigma 0.9$& \textbf{7.00}$~\sigma 1.2$& $10.60~\sigma 2.1$\\
 & CFL& $8.20~\sigma 1.6$& $3.00~\sigma 0.6$& $1.20~\sigma 0.4$& $6.50~\sigma 0.8$& $12.00~\sigma 2.5$\\
 & CFL+FT& $9.50~\sigma 2.2$& $3.30~\sigma 0.5$& $2.20~\sigma 0.4$& $6.80~\sigma 0.7$& $10.00~\sigma 2.1$\\
 & MaTFL& $9.00~\sigma 1.7$& $1.40~\sigma 0.5$& $5.40~\sigma 0.8$& $5.40~\sigma 1.4$& $9.70~\sigma 1.6$\\
 & FedPref+FT (ours)& \textbf{10.70}$~\sigma 1.3$& $3.20~\sigma 0.6$& $7.80~\sigma 1.0$& $6.50~\sigma 0.7$& \textbf{12.30}$~\sigma 1.7$\\
 & FedPref-FT (ours)& $10.70~\sigma 1.5$& $3.10~\sigma 0.3$& \textbf{8.00}$~\sigma 0.8$& $6.90~\sigma 1.0$& $10.00~\sigma 2.5$\\
\toprule 
\multirow{10}{*}{Gaussian} & No comm.& $12.20~\sigma 1.5$& $3.20~\sigma 0.4$& $7.10~\sigma 0.7$& $6.10~\sigma 1.2$& \textbf{12.00}$~\sigma 3.3$\\
 & FedAvg& $7.60~\sigma 1.5$& $1.00~\sigma 0.0$& $1.00~\sigma 0.0$& $3.00~\sigma 0.4$& $9.40~\sigma 2.8$\\
 & FedAvg+FT& $10.10~\sigma 2.4$& $1.20~\sigma 0.4$& $3.40~\sigma 1.1$& $6.70~\sigma 0.6$& $8.90~\sigma 3.7$\\
 & FedProx& $9.10~\sigma 3.4$& $1.00~\sigma 0.0$& $1.00~\sigma 0.0$& $3.90~\sigma 1.4$& $11.90~\sigma 2.6$\\
 & FedProx+FT& $10.80~\sigma 1.8$& $1.10~\sigma 0.3$& $3.00~\sigma 1.0$& $6.80~\sigma 1.3$& $10.00~\sigma 3.0$\\
 & CFL& $6.90~\sigma 3.0$& $3.40~\sigma 0.8$& $1.20~\sigma 0.4$& $5.50~\sigma 1.6$& $11.80~\sigma 2.6$\\
 & CFL+FT& $12.00~\sigma 2.5$& \textbf{3.90}$~\sigma 0.3$& $1.80~\sigma 0.6$& $6.70~\sigma 1.3$& $11.20~\sigma 2.3$\\
 & MaTFL& $11.70~\sigma 1.7$& $1.70~\sigma 0.6$& $5.80~\sigma 1.2$& $5.10~\sigma 1.1$& $11.90~\sigma 2.1$\\
 & FedPref+FT (ours)& \textbf{12.60}$~\sigma 2.0$& $3.30~\sigma 0.8$& $7.10~\sigma 0.5$& $6.30~\sigma 1.1$& $11.80~\sigma 2.4$\\
 & FedPref-FT (ours)& $12.50~\sigma 2.2$& $3.30~\sigma 0.5$& \textbf{7.80}$~\sigma 1.3$& \textbf{7.00}$~\sigma 1.1$& $10.70~\sigma 1.5$\\
\bottomrule
\end{tabular}
\end{table*}

\subsection{Ablation study - multi-objective performance}
In this section, we briefly revisit our previously discussed ablation study under multi-objective aspects. The corresponding multi-objective metric results are shown in Table~\ref{table:ablation-study-mo}. For four out of five experimental environments, the results quite clearly indicate that a variant of the FedPref algorithm yields a better performance under multi-objective aspects than either of its components in isolation. For the MO-LLc.~environment, the optimal values for three out of four metrics are obtained by the FedPref algorithm. For each of the MO-LL, DMC, and MO-LLc. environments, a variant of the combined algorithm achieves or matches both the highest hypervolume and cardinality, and in both cases also the lowest IGD value. In combination, these results indicate that the solution sets obtained by the respective federated systems do accomplish the highest diversity and convergence, and that the elevated sparsity in these cases is simply a consequence of the overall wider distribution of solutions. Results for the DST algorithm give a similar impression.\\

The results for the MO-HC environment follow a slightly different pattern: FedPref-FT yields the highest hypervolume value in this case, but the best scores for the remaining metrics are obtained by variants of the Weighted aggregation component. It appears that this component in isolation achieves a higher convergence than the FedPref algorithm, at the cost of a reduction in the diversity of solutions.\\
Finally, we note that the addition of a fine-tuning step at the end of the local training phase generally does not greatly impact the multi-objective results in most cases. While variants with fine-tuning tend to produce higher cardinalities (i.e.~more optimal trade-off solutions), this does not generally translate to improved results in the other metrics. A brief fine-tuning phase is likely not sufficient to significantly increase the diversity of the solution set; new trade-off solutions discovered during fine-tuning would tend to be very similar to each other.

\begin{table}[ht]
   \caption{Experimental results comparing multi-objective metrics obtained by the individual components of our algorithm.}\label{table:ablation-study-mo}
\centering
\begin{tabular}{llllll} 
\toprule 
  & & Hypervolume ($\uparrow$) & Cardinality ($\uparrow$)& Sparsity ($\downarrow$) & IGD ($\downarrow$) \\ 
\toprule 
\multirow{5}{*}{Clustering only+FT} & MO-LL& $187.16~\sigma 11.1(\cdot 10^{7})$& \textbf{13.70}$~\sigma 2.1$& $17.12~\sigma 6.5(\cdot 10^{2})$& $42.12~\sigma 4.3$\\
 & DMC& $174.08~\sigma 16.7$& $2.60~\sigma 0.5$& $1.17~\sigma 0.2$& $0.37~\sigma 0.1$\\
 & DST& $211.92~\sigma 24.9(\cdot 10^{1})$& $7.20~\sigma 1.2$& $28.99~\sigma 11.5$& $1.47~\sigma 1.2$\\
 & MO-HC& $417.69~\sigma 51.5(\cdot 10^{4})$& $5.90~\sigma 1.1$& $434.17~\sigma 175.2(\cdot 10^{4})$& $819.40~\sigma 270.3$\\
 & MO-LLc.& $97.95~\sigma 6.4(\cdot 10^{7})$& $12.20~\sigma 3.1$& $242.55~\sigma 153.5(\cdot 10^{1})$& $50.27~\sigma 5.8$\\
\toprule 
\multirow{5}{*}{Clustering only-FT} & MO-LL& $194.61~\sigma 7.6(\cdot 10^{7})$& $11.70~\sigma 2.9$& $19.35~\sigma 14.5(\cdot 10^{2})$& $49.28~\sigma 11.5$\\
 & DMC& $166.12~\sigma 30.5$& $2.70~\sigma 0.8$& $1.46~\sigma 0.8$& $0.43~\sigma 0.2$\\
 & DST& \textbf{222.85}$~\sigma 4.9(\cdot 10^{1})$& $6.80~\sigma 1.3$& $43.30~\sigma 33.4$& $0.99~\sigma 0.8$\\
 & MO-HC& $411.80~\sigma 41.3(\cdot 10^{4})$& $6.00~\sigma 1.4$& $500.84~\sigma 373.0(\cdot 10^{4})$& $810.57~\sigma 276.1$\\
 & MO-LLc.& $97.22~\sigma 10.6(\cdot 10^{7})$& $11.30~\sigma 2.8$& $158.75~\sigma 185.1(\cdot 10^{1})$& $54.90~\sigma 6.7$\\
\toprule 
\multirow{5}{*}{Weighted agg.~only+FT} & MO-LL& $175.09~\sigma 14.6(\cdot 10^{7})$& $12.40~\sigma 1.9$& $77.43~\sigma 95.1(\cdot 10^{2})$& $50.04~\sigma 6.2$\\
 & DMC& $145.41~\sigma 28.5$& $2.30~\sigma 0.5$& $8.63~\sigma 18.2$& $0.74~\sigma 0.2$\\
 & DST& $180.06~\sigma 27.7(\cdot 10^{1})$& $4.10~\sigma 1.3$& $21.83~\sigma 23.5$& $4.02~\sigma 1.5$\\
 & MO-HC& $426.41~\sigma 20.7(\cdot 10^{4})$& $6.60~\sigma 1.2$& $269.32~\sigma 107.9(\cdot 10^{4})$& \textbf{679.38}$~\sigma 205.9$\\
 & MO-LLc.& $98.90~\sigma 4.7(\cdot 10^{7})$& $10.40~\sigma 2.0$& $285.12~\sigma 145.8(\cdot 10^{1})$& $50.11~\sigma 4.2$\\
\toprule 
\multirow{5}{*}{Weighted agg.~only-FT} & MO-LL& $186.64~\sigma 13.9(\cdot 10^{7})$& $8.80~\sigma 2.7$& \textbf{5.86}$~\sigma 7.2(\cdot 10^{2})$& $75.88~\sigma 16.7$\\
 & DMC& $59.68~\sigma 29.9$& $1.10~\sigma 0.3$& \textbf{0.18}$~\sigma 0.5$& $1.11~\sigma 0.2$\\
 & DST& $157.11~\sigma 22.6(\cdot 10^{1})$& $2.20~\sigma 1.2$& \textbf{12.85}$~\sigma 14.2$& $5.98~\sigma 1.8$\\
 & MO-HC& $300.40~\sigma 110.3(\cdot 10^{4})$& \textbf{7.20}$~\sigma 1.2$& \textbf{202.30}$~\sigma 311.4(\cdot 10^{4})$& $1461.76~\sigma 522.8$\\
 & MO-LLc.& $93.47~\sigma 7.7(\cdot 10^{7})$& $9.40~\sigma 3.5$& \textbf{51.86}$~\sigma 45.9(\cdot 10^{1})$& $64.97~\sigma 4.9$\\
\toprule 
\multirow{5}{*}{FedPref+FT (combined)} & MO-LL& $193.94~\sigma 7.5(\cdot 10^{7})$& $13.70~\sigma 1.8$& $19.35~\sigma 8.6(\cdot 10^{2})$& \textbf{39.22}$~\sigma 3.0$\\
 & DMC& $202.79~\sigma 29.7$& \textbf{3.40}$~\sigma 0.7$& $1.83~\sigma 1.0$& $0.38~\sigma 0.2$\\
 & DST& $220.36~\sigma 10.3(\cdot 10^{1})$& \textbf{7.40}$~\sigma 1.0$& $28.00~\sigma 7.6$& \textbf{0.77}$~\sigma 0.3$\\
 & MO-HC& $421.43~\sigma 39.5(\cdot 10^{4})$& $5.80~\sigma 1.2$& $408.45~\sigma 167.6(\cdot 10^{4})$& $738.36~\sigma 183.2$\\
 & MO-LLc.& $100.04~\sigma 5.5(\cdot 10^{7})$& $12.50~\sigma 2.1$& $194.61~\sigma 88.7(\cdot 10^{1})$& \textbf{47.86}$~\sigma 6.7$\\
\toprule 
\multirow{5}{*}{FedPref-FT (combined)} & MO-LL& \textbf{201.97}$~\sigma 7.0(\cdot 10^{7})$& $13.30~\sigma 3.1$& $19.38~\sigma 17.0(\cdot 10^{2})$& $47.24~\sigma 7.5$\\
 & DMC& \textbf{203.74}$~\sigma 13.2$& $3.30~\sigma 0.5$& $1.88~\sigma 0.8$& \textbf{0.37}$~\sigma 0.2$\\
 & DST& $218.04~\sigma 18.3(\cdot 10^{1})$& $6.80~\sigma 1.1$& $33.07~\sigma 7.9$& $1.09~\sigma 0.9$\\
 & MO-HC& \textbf{435.87}$~\sigma 36.5(\cdot 10^{4})$& $5.90~\sigma 1.3$& $517.10~\sigma 377.2(\cdot 10^{4})$& $770.74~\sigma 286.1$\\
 & MO-LLc.& \textbf{101.57}$~\sigma 8.8(\cdot 10^{7})$& \textbf{12.70}$~\sigma 2.7$& $92.45~\sigma 68.8(\cdot 10^{1})$& $51.38~\sigma 8.1$\\
\bottomrule
    \end{tabular}
\end{table}

\section{Conclusion and outlook}
In this work, we have discussed multi-objective preference heterogeneity, a novel type of heterogeneity problem that arises naturally in many real-world settings, including those where Federated Learning may be used. In this setting, all clients deal with multiple objectives, but each objective has a different individual importance to each client. We have proposed a new algorithm to perform Federated Learning in this setting, based on a combination of recursive clustering and weighted aggregation, both using a modified model-similarity metric. This algorithm preserves the privacy of clients: it is capable of functioning using only the respective client model updates; no further information about client objectives is required. %
We have validated the performance of our algorithm on multiple and varied problems and preference distributions, comparing it to classical benchmarks as well as other heterogeneity-mitigating algorithms. 
We have analysed the results from two different points of view:
First, we have considered the traditional client-centric view, demonstrating that our algorithm outperforms the alternatives in many cases in terms of mean client performance, and represents a reliable choice in all others. Further experiments were carried out to study the characteristics of the algorithm.\\

In addition, we have discussed a multi-objective view of the federated system, analysing the performance of the FedPref algorithm under multi-objective aspects. Our results show that, while the algorithm does not currently explicitly enforce multi-objective characteristics, it nevertheless performs well on several common multi-objective metrics. These results, too, persist across different types of problems and heterogeneity distributions, in contrast to the compared algorithms.

\section{Limitations and future work}
As this work presents an initial solution tailored to the objective-heterogeneous setting, several challenges inherent to the federated setting remain to be addressed in future work. This includes, in particular, scenarios dealing with combined occurrences of different types of heterogeneity, such as data or hardware heterogeneity combined with the preference heterogeneity discussed here. In principle, we expect that the model similarity-based design of our algorithm could adapt without change to a setting that includes data heterogeneity; heterogeneity induced by differences in client capabilities might require the integration of additional strategies dedicated to this purpose. Such strategies already exist in the literature; their integration into the cluster-aggregation step of FedPref appears quite feasible.\\
In addition, we have introduced the multi-objective view of the federated system in this paper. Our experiments demonstrate that it appears to be possible to design a personalised federated algorithm that achieves both high individual client performance and a diverse set of client solutions for different preferences. Nevertheless, further study of the implications of various MOO metrics in this setting would be useful. Furthermore, it would be interesting to apply this multi-objective analysis to other federated algorithms from the state of the art.

\begin{acks}
This work is partially funded by the joint research programme UL/SnT--ILNAS on Technical Standardisation for Trustworthy ICT, Aerospace, and Construction. The experiments presented in this paper were carried out using the HPC facilities of the University of Luxembourg~\cite{VCPKVO_HPCCT22} {\small -- see \url{https://hpc.uni.lu}}. The authors would like to thank Florian Felten for proof-reading an earlier version of this work.
\end{acks}

\bibliographystyle{ACM-Reference-Format}
\bibliography{bibliography}
\newpage
\appendix
\section{Appendix}
\subsection*{Details of experiment configurations}
\subsubsection{Hyperparameter tuning}
We perform an initial hyperparameter search for all algorithms and environments, with all evaluated parameter values listed in Table~\ref{app:table:hpo-params}. Each configuration was run five times and five different randomly-generated preference distributions. For all environments except the MO-Halfcheetah environment, all experiments were performed with $20$ simulated clients per system. Due to the comparably high computational cost of training on the MO-Halfcheetah environment, the number of clients was reduced to $10$ clients per system in this case. The five preference distributions remained fixed across all hyperparameter configurations to promote the comparability of results. The metric used to assess performance was the mean linearised reward obtained by the clients using their personalised preference weights. The parameter values selected as a result of the hyperparameter tuning are given in Table~\ref{app:table:params-selected-dqn} and ~\ref{app:table:params-selected-ddpg}.

\begin{table*}[hb]
\caption{Complete list of parameter configurations tested during hyperparameter tuning of DQN algorithms.}\label{app:table:params-selected-dqn}
\begin{threeparttable}
\centering
\begin{tabular}{llllll} 
& & MO-LL & DMC & DST & Comment \\
\toprule
No comm. & - & - & - & - & No federated parameters. \\
\hline
FedAvg & Num. local iterations & $(2, 5, 10\cdot 10^3$ & $(2, 5, 10)\cdot 10^3$ & $(5, 10, 15)\cdot 10^2$ & \\
\hline
\multirow{2}{*}{FedProx} & Num. local iterations & $(2, 5, 10\cdot 10^3$ & $(2, 5, 10)\cdot 10^3$ & $(5, 10, 15)\cdot 10^2$ & \\
                        & Proximal term $\mu$ & $0.01, 0.1, 1$ & $0.01, 0.1, 1$ & $0.01, 0.1, 1$ & Based on \cite{li_federated_2018}\\
                        \hline
\multirow{3}{*}{CFL} & Num. local iterations & $(2, 5, 10\cdot 10^3$ & $(2, 5, 10)\cdot 10^3$ & $(5, 10, 15)\cdot 10^2$ & \\
                    & Clustering threshold & $2.5, 5, 7.5$ & $2, 3, 5$ & $2.5, 5, 7.5$ & See \tnote{a}\\
                    & Patience & $1, 2$ & $1, 2$ & $1, 2$ & See \tnote{b} \\
                    \hline
\multirow{2}{*}{MaTFL} & Num. local iterations & $(2, 5, 10\cdot 10^3$ & $(2, 5, 10)\cdot 10^3$ & $(5, 10, 15)\cdot 10^2$ & \\
                    & Num. voting clients $k$ & $5, 8, 10$ & $5, 8, 10$ & $5, 8, 10$ & Adequate according to \cite{cai_many_2023}\\
                    \hline
\multirow{4}{*}{Ours} & Num. local iterations & $(2, 5, 10\cdot 10^3$ & $(2, 5, 10)\cdot 10^3$ & $(5, 10, 15)\cdot 10^2$ & \\
                    & Clustering threshold & $2.5, 5, 7.5$ & $2, 3, 5$ & $2.5, 5, 7.5$ & Same as for CFL \\
                    & Patience & $1, 2$ & $1, 2$ & $1, 2$ & Same as for CFL \\
                    & Min. similarity & $-1, 0$ & $-1, 0$ & $-1, 0$ & Used during aggregation\tnote{c} \\
                    
\bottomrule
\end{tabular}
\begin{tablenotes}
   \item[a] Based on max.~observed gradient magnitude ,as suggested in \cite{Sattler2019ClusteredFL}.
   \item[b] Rounds below threshold before clustering triggered. Introduced by us to handle slow initial gradient ramp-up.
   \item[c] See Section~\ref{subsec:theory-weighted-aggregation} in the main paper for explanation.
\end{tablenotes}
\end{threeparttable}
\end{table*}

\begin{table*}
\caption{Complete list of parameter configurations tested during hyperparameter tuning of DDPG algorithms.}\label{app:table:params-selected-ddpg}
\begin{threeparttable}
\centering
\begin{tabular}{llllll} 
& & MO-LLcont. & MO-HC & Comment \\
\toprule
No comm. & - & - & - & No federated parameters. \\
\hline
FedAvg & Num. local iterations & $5000, 10000, 15000$ & $25000, 37500, 50000$  & \\
\hline
\multirow{2}{*}{FedProx} & Num. local iterations & $5000, 10000, 15000$ & $25000, 37500, 50000$ & \\
                        & Proximal term $\mu$ & $0.01, 0.1, 1$ & $0.01, 0.1, 1$ & Based on recommendations in \cite{li_federated_2018}\\
                        \hline
\multirow{5}{*}{CFL} & Num. local iterations & $5000, 10000, 15000$ & $25000, 37500, 50000$ & \\
                    & Clustering threshold & $10, 15, 20$ & $20, 30, 40$ & Based on max.~observed gradient\\
                    & & & & magnitude\tnote{a} \\
                    & Patience & $1, 2$ & $1, 2$ & Rounds below threshold before\\
                    & & & & clustering triggered\tnote{b} \\
                    \hline
\multirow{2}{*}{MaTFL} & Num. local iterations & $5000, 10000, 15000$ & $25000, 37500, 50000$ & \\
                    & Num. voting clients $k$ & $5, 8, 10$ & $3, 5, 8$ &Adequate range according to \cite{cai_many_2023}\\
                    \hline
\multirow{4}{*}{Ours} & Num. local iterations & $5000, 10000, 15000$ & $25000, 37500, 50000$ & \\
                    & Clustering threshold & $10, 15, 20$ & $20, 30, 40$  & Same as for CFL \\
                    & Patience & $1, 2$ & $1, 2$ & Same as for CFL \\
                    & Min. similarity & $-1, 0$ & $-1, 0$ & Used in computing aggregation weights\tnote{c} \\
                    
\bottomrule
\end{tabular}
\begin{tablenotes}
   \item[a] As suggested in \cite{Sattler2019ClusteredFL}.
   \item[b] Introduced by us to handle slow initial gradient ramp-up.
   \item[c] See Section~\ref{subsec:theory-weighted-aggregation} in the main paper for explanation.
\end{tablenotes}
\end{threeparttable}
\end{table*}

\begin{table*}
\caption{Parameter configurations selected for each algorithm following hyperparameter tuning.}\label{app:table:hpo-params}
\centering
\begin{tabular}{lllllll} 
\toprule
& & MO-LL & DMC & DST & MO-HC & MO-LLc.\\
\toprule
No comm. & - & - & - & - & - & - \\
\hline
FedAvg & Number local iterations & $5000$ & $5000$ & $500$ & $25000$ & $5000$\\
\hline
\multirow{2}{*}{FedProx} & Number local iterations & $5000$ & $5000$ & $500$ & $25000$ & $5000$\\
                        & Proximal term $\mu$ & $1$ & $1$ & $1$ & $0.01$ & $0.01$\\
                        \hline
\multirow{3}{*}{CFL} & Number local iterations & $10000$ & $2000$ & $1000$ & $25000$ & $5000$ \\
                    & Clustering threshold & $5$ & $2$ & $5$ & $30$ & $15$\\
                    & Patience & $2$ & $2$ & $2$ & $1$ & $2$\\
                    \hline
\multirow{2}{*}{MaTFL} & Number local iterations & $2000$ & $5000$ & $1000$ & $25000$ & $5000$\\
                    & Number voting clients $k$ & $10$ & $10$ & $10$ & $8$ & $10$\\
                    \hline
\multirow{4}{*}{FedPref (ours)} & Number local iterations & $5000$ & $5000$ & $500$ & $25000$ & $10000$\\
                    & Clustering threshold & $5$ & $3$ & $5$ & $30$ & $15$\\
                    & Patience & $1$ & $2$ & $2$ & $2$ & $2$\\
                    & Minimum similarity & $-1$ & $0$ & $-1$ & $-1$ & $-1$ \\
                    
\bottomrule
\end{tabular}
\end{table*}

\subsubsection{MORL environment parameters}
Where available, the hyperparameters for the three MORL environments used in our experiments were obtained from published benchmark configurations. Where no such configurations were available, they were obtained by manual tuning. All modified parameters are reported below, in Tables~\ref{appx:tab:rl-params-dqn}, and \ref{appx:tab:rl-params-ddpg}. Parameters that are not listed can be assumed to be set to the default setting, as implemented in the DQN and DDPG algorithms, respectively, of the stable-baselines3 package.

\begin{table} [ht]
    \centering
    \begin{tabular}{l lll}
        \toprule
        Parameter name & MO-Lunar Lander & Det. Minecart & Deep-Sea Treasure \\
        \hline
        env & mo-lunar-lander-v2 & minecart-deterministic-v0 & deep-sea-treasure-v0\\
        policy & MlpPolicy & MlpPolicy & MlpPolicy\\
        learning\_rate & $0.00063$ & $0.0002$ & $0.004$\\
        batch\_size & $64$ & $64$ & $128$\\
        buffer\_size & $50000$ & $50000$ & $10000$\\
        learning\_starts & $0$ & $50000$ & $1000$\\
        gamma & $0.99$ & $0.99$ & $0.98$\\
        target\_update\_interval & $250$ & $750$ & $600$\\
        train\_freq & $4$ & $32$ & $16$\\
        gradient\_steps & $-1$ & $32$ & $8$\\
        exploration\_fraction & $0.12$ & $0.8$ & $0.2$\\
        exploration\_final\_eps & $0.1$ & $0.05$ & $0.07$\\
        net\_arch & [$256, 256$] & [$256$, $256$] & [$256$, $256$]\\
        \hline
    \end{tabular}
    \caption{Set of parameters used for the local training of the MO-Lunar Lander, Deterministic Minecart and Deep-Sea Treasure environments, using the DQN algorithm.}
    \label{appx:tab:rl-params-dqn}
\end{table}

\begin{table}[ht]
    \centering
    \begin{tabular}{l ll}
        \toprule
        Parameter name & MO-Halfcheetah & MO-Lunar Lander cont. \\
        \hline
        env & mo-halfcheetah-v4 & mo-lunar-lander-continuous-v2 \\
        noise\_std & $0.1$ & $0.1$\\
        policy & MlpPolicy & MlpPolicy\\
        learning\_rate & $0.001$ & $0.001$\\
        buffer\_size & $200000$ & $200000$ \\
        learning\_starts & - & $10000$\\
        gamma & $0.98$ & $0.98$ \\
        train\_freq & $1$ & $1$ \\
        gradient\_steps & $1$ & -\\
        net\_arch & [$400, 300$] & [$400, 300$]\\
        \hline
    \end{tabular}
    \caption{Set of parameters used for the local training of the MO-Halfcheetah and MO-Lunar Lander continuous environment.}
    \label{appx:tab:rl-params-ddpg}
\end{table}

\subsubsection{Computing resources}
The number of experiments presented in this paper amounts to $4125$ individual experimental runs. This corresponds to a total runtime of approximately $8560$ hours on a single node of the computing cluster available to us.

\subsection{Supplementary experimental results}\label{appx:subsec:experiment-results}
In this section, we include supplementary numerical results and plots that exceeded the scope of the main paper. 
\subsubsection{Impact of topR parameter and similarity bound}
This section lists numerical experimental results for the parameter sensitivity analyses carried out in the main paper; these same results are presented there in visual form, with some numbers quoted. We refer the reader to the relevant section in the main paper for the analysis and discussion of these results. Table~\ref{appx:tab:topr-sensitivity-results} contains the results for the sensitivity analysis of the $topR$ parameter; Table~\ref{appx:tab:minsim-sensitivity-results} shows the results for the analysis of the minimum similarity threshold in aggregation. All experiments were carried out with $10$ different random seeds, on preference weights drawn from a Dirichlet distribution. Experiments on all environments were run on systems of $20$ federated clients, with the exception of the MO-Halfcheetah environment, which was restricted to systems of $10$ clients due to its high computational cost.

\begin{table}[t]
    \centering
    \begin{tabular}{llllll}
    \toprule
    Threshold & MO-LL & DMC & DST & MO-HC & MO-LLc.\\
    \toprule
        $-1.0$ & \textbf{32.22} $\sigma 11.3$ & $-1.91 \sigma 1.0$ & $4.41 \sigma 1.7$ & $2980.26 \sigma 784.4$ & $32.17 \sigma 7.0$\\
        $-0.8$ & $31.74 \sigma 11.3$ & $-2.29 \sigma 1.0$ & $2.49 \sigma 1.7$ & $3038.06 \sigma 784.4$ & $31.34 \sigma 7.0$\\
        $-0.6$ & $32.09 \sigma 13.0$ & $-2.42 \sigma 1.7$ & \textbf{4.63} $\sigma 3.1$ & \textbf{3049.70} $\sigma 741.8$ & $29.08 \sigma 8.3$ \\
        $-0.4$ & $29.65 \sigma 13.2$ & \textbf{-1.82} $\sigma 1.5$ & $3.20 \sigma 1.5$ & $2967.62 \sigma 896.8$ & \textbf{32.44} $\sigma 15.3$\\
        $-0.2$ & $27.47 \sigma 11.2$ & $-2.63 \sigma 0.9$ & $2.73 \sigma 2.5$ & $2761.86 \sigma 783.1$ & $30.41 \sigma 9.3$\\
        $0.0$ & $22.73 \sigma 16.4$ & $-2.42 \sigma 1.2$ & $1.56 \sigma 2.0$ & $2494.44 \sigma 791.0$ & $28.70 \sigma 9.8$\\
        $0.2$ & $13.24 \sigma 11.7$ & $-2.24 \sigma 1.6$ & $-0.69 \sigma 2.0$ & $2389.85 \sigma 649.7$ & $16.98 \sigma 10.8$\\
        $0.4$ & $13.53 \sigma 7.1$ & $-3.61 \sigma 0.7$ & $0.92 \sigma 2.3$ & $2477.75 \sigma 603.5$ & $16.84 \sigma 11.7$\\
        $0.6$ & $14.03 \sigma 11.9$ & $-2.78 \sigma 2.2$ & $0.85 \sigma 1.5$ & $2489.83 \sigma 682.9$ & $14.46 \sigma 9.2$ \\
        $0.8$ & $9.95 \sigma 10.0$ & $-2.72 \sigma 1.5$ & $1.20 \sigma 2.2$ & $2391.18 \sigma 369.0$ & $15.08 \sigma 12.4$ \\
        
    \end{tabular}
    \caption{Numerical results for minimum-similarity sensitivity analysis visualised in the main part of the paper. All configurations except those on the MO-HC environment were run with $10$ different random seeds across $20$ clients per run. Due to the higher computational cost of solving the MO-HC environment, experiments on this environment were restricted to $10$ clients per run, also for $10$ runs per configuration.}
    \label{appx:tab:minsim-sensitivity-results}
\end{table}

\begin{table}[b]
    \centering
    \begin{tabular}{clllll}
    \toprule
    $topR$ & MO-LL & DMC & DST & MO-HC & MO-LLc.\\
    \toprule
       $0.2$ & \textbf{33.18}$~\sigma 11.1$ & $-2.58~\sigma 1.0$ & $2.24~\sigma 3.0$ & $2909.41~\sigma 636.1$ & $30.51~\sigma 9.3$\\
       $0.4$ & $30.04~\sigma 11.1$ & $-3.18~\sigma 1.0$ & $1.89~\sigma 3.0$ & \textbf{3043.55}$~\sigma 636.1$ & $32.71~\sigma 9.3$\\
       $0.6$ & $30.42~\sigma 13.3$ & \textbf{-2.44}$~\sigma 1.8$ & $1.35~\sigma 3.8$ & $2995.96~\sigma 893.1$ & $32.04~\sigma 12.5$\\
       $0.8$ & $31.17~\sigma 11.7$ & $-2.65~\sigma 1.3$ & \textbf{3.76}$~\sigma 4.0$ & $2968.06~\sigma 915.7$ & \textbf{33.39}$~\sigma 11.3$ \\
       $1.0$ & $30.84~\sigma 10.4$ & $-2.49~\sigma 1.2$ & $2.52~\sigma 2.8$ & $2967.83~\sigma 714.8$ & $32.03~\sigma 10.7$\\
    \end{tabular}
    \caption{Numerical results for $topR$ sensitivity analysis visualised in the main part of the paper. All configurations except those on the MO-HC environment were run with $10$ different random seeds across $20$ clients per run. Due to the higher computational cost of solving the MO-HC environment, experiments on this environment were restricted to $10$ clients per run, also for $10$ runs per configuration.}
    \label{appx:tab:topr-sensitivity-results}
\end{table}

\subsubsection{FedPref clustering validation}
\begin{figure}
    \centering
    \includegraphics[width=.75\columnwidth]{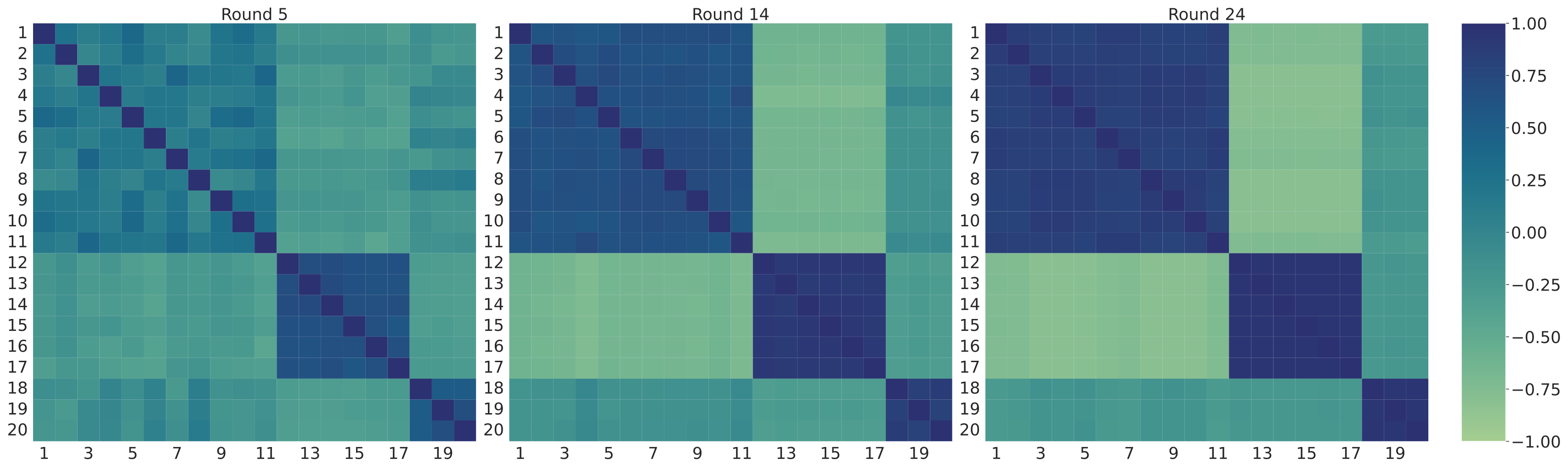}
    \caption{Mutual client similarity at different stages during a single experimental run on the MO-LL environment, with unbalanced preference distribution. Left to right: client similarities after aggregation round $5, 14$ and $24$ of $28$, respectively.}
    \label{appx:fig:cval-heatmap-ll-uneq}
    \Description{Three heatmaps. In each, three clusters of various sizes are apparent, becoming increasingly clearly delineated as training progresses.}
\end{figure}

In this section, we show and briefly discuss additional results of the clustering validation experiments.\\
\textbf{MO-Lunar Lander.}
Figure~\ref{appx:fig:cval-heatmap-ll-uneq} shows the similarity of clients at three training stages during training in the MO-LL environment on an unbalanced preference distribution. Three groups with distinct similarity are clearly recognisable from the earliest stages of the training process; these correspond to the sets of clients that have been assigned the same preferences, with two such sets evidently grouped together. Later stages show the gradual separation of the different sets, likely through the clustering process. However, the two client sets that showed a high similarity from the beginning (both contained in the largest, top-left block in the figure) appear to remain in the same cluster until the end of the training process, never being separated. This could indicate either that the two different preference weights assigned to the two sets are naturally compatible during the training process, or that the FedPref algorithm might sometimes struggle to fully separate incompatible sets of clients before they converge to a local optimum. The latter could also be a consequence of the imbalanced distribution of potentially incompatible clients in this case; perhaps a small number of incompatible clients is 'dominated' by the remaining large number of compatible clients in the same cluster.

\begin{figure}
    \centering
    \includegraphics[width=.75\columnwidth]{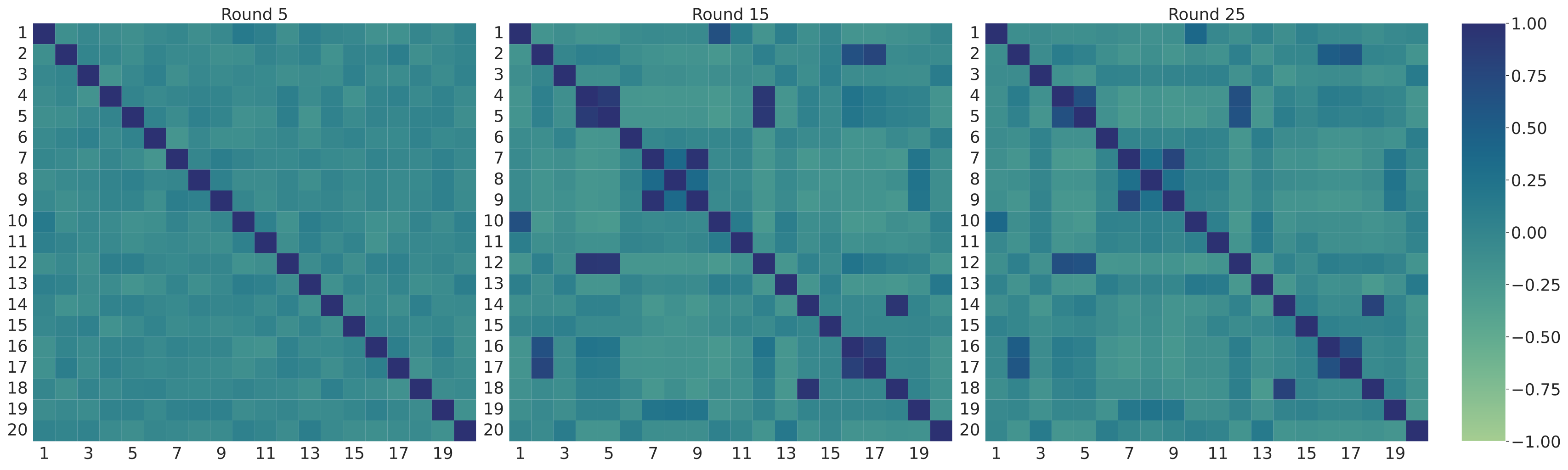}
    \caption{Mutual client similarity at different stages during a single experimental run on the DMC environment, with balanced preference distribution. Left to right: client similarities after aggregation round $5, 15$ and $25$ of $38$, respectively.}
    \label{appx:fig:cval-heatmap-dmc-eq}
    \Description{Three heatmaps. At the beginning of training, no pattern is apparent. Later, some pair-wise similarities appear, but these do not generally match the underlying preference similarities. One cluster becomes faintly apparent.}
\end{figure}

\begin{figure}
    \centering
    \includegraphics[width=.75\columnwidth]{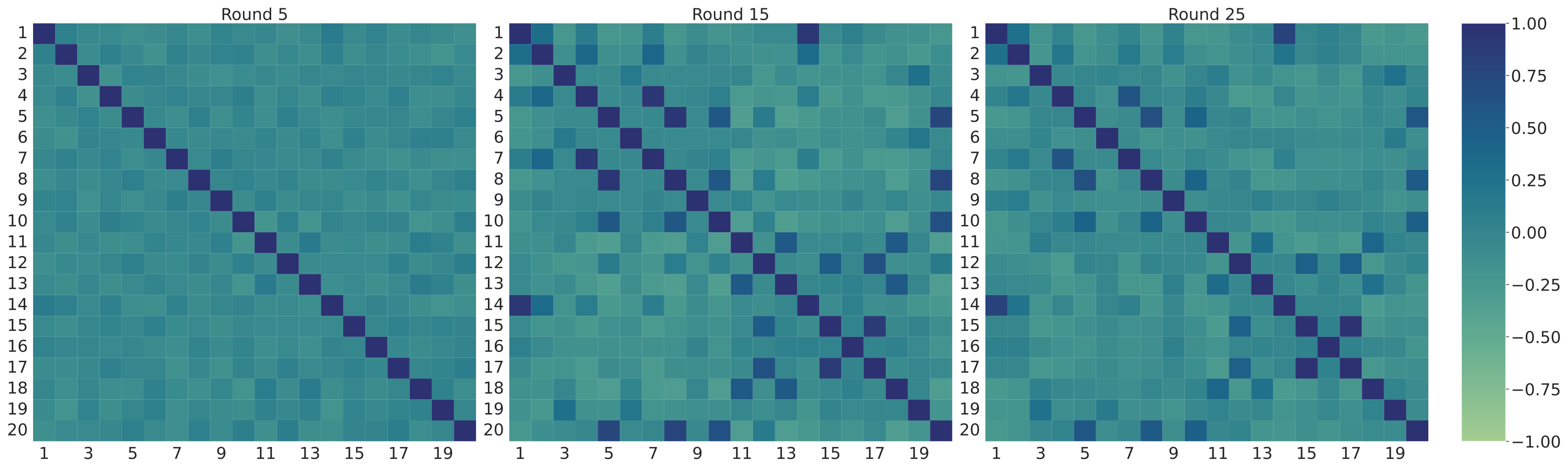}
    \caption{Mutual client similarity at different stages during a single experimental run on the DMC environment, with unbalanced preference distribution. Left to right: client similarities after aggregation round $5, 15$ and $25$ of $38$, respectively.}
    \label{appx:fig:cval-heatmap-dmc-uneq}
    \Description{Three heatmaps. At the beginning of training, no pattern is apparent. Later, some pair-wise similarities appear, but these do not generally match the underlying preference similarities.}
\end{figure}

\textbf{Det.~Minecart.} Figure~\ref{appx:fig:cval-heatmap-dmc-eq} and Figure~\ref{appx:fig:cval-heatmap-dmc-uneq} show client similarities during training on the Det.~Minecart environment with the balanced and unbalanced distribution of preferences, respectively. These results also illustrate the challenges of this environment that were discussed in the main part of the paper: the sparse reward space appears to make it difficult to reliably discover client similarities during the clustering process. We observe in both figures that clients never reach high levels of similarity as seen in the results of the MO-LL environment; it is likely that this also impedes the clustering process, leading to a suboptimal grouping into clusters. However, some successful collaboration appears to take place, as evidences by the darker-coloured patches in the middle and right images in both figures. This matches our experimental conclusions in the main paper, that the FedPref algorithm does accomplish some useful collaboration leading to improvement of client results, but highly sparse solution spaces remain a challenge.\\


\begin{figure}[bh]
    \centering
    \includegraphics[width=.75\columnwidth]{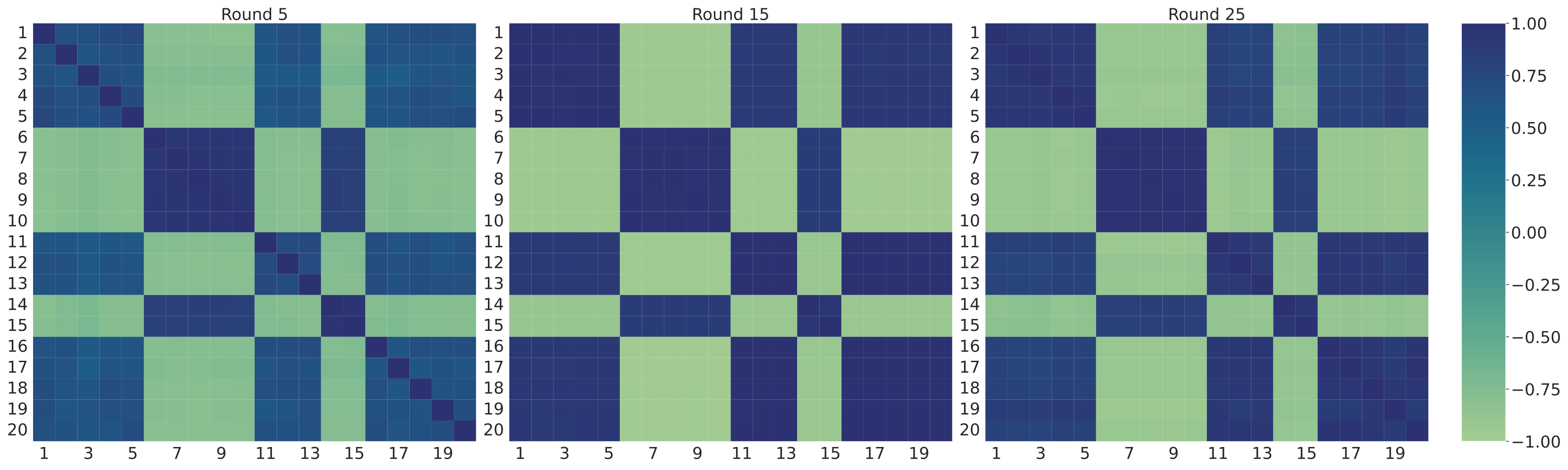}
    \caption{Mutual client similarity at different stages during a single experimental run on the DST environment, with balanced preference distribution. Left to right: client similarities after aggregation round $5, 15$ and $25$ of $28$, respectively.}
    \label{appx:fig:cval-heatmap-dst-eq}
    \Description{Three heatmaps. In the first, five groups are apparent, three matching the assigned preference groups, and two more split from the fourth preference group. These groups remain consistent as training progresses.}
\end{figure}

\begin{figure}[th]
    \centering
    \includegraphics[width=.75\columnwidth]{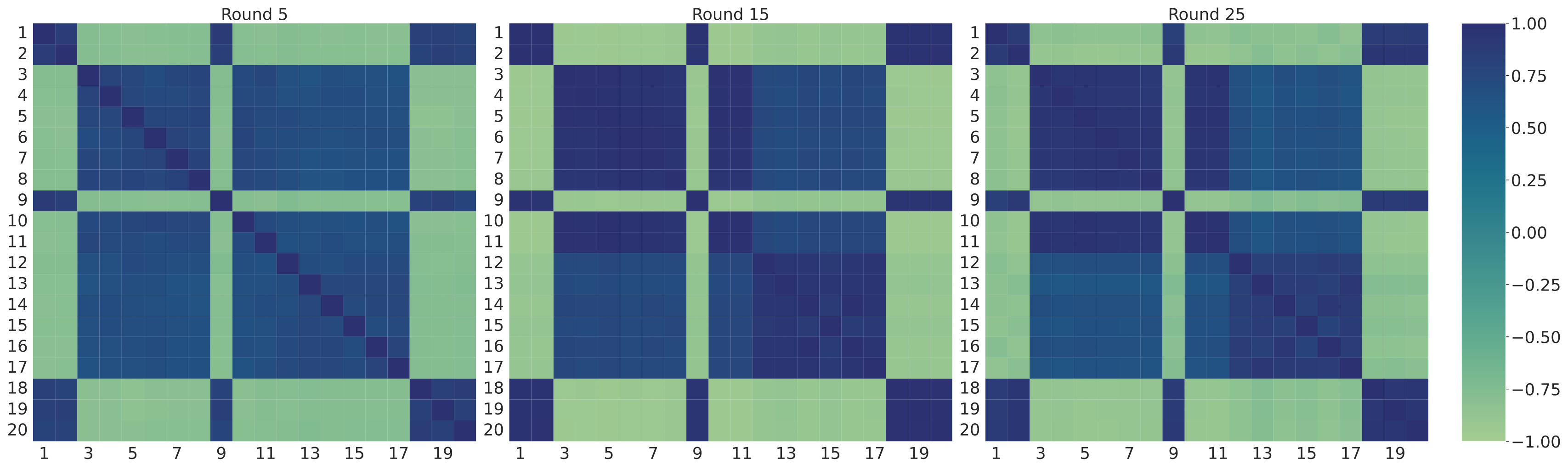}
    \caption{Mutual client similarity at different stages during a single experimental run on the DST environment, with unbalanced preference distribution. Left to right: client similarities after aggregation round $5, 15$ and $25$ of $28$, respectively.}
    \Description{Three heatmaps, showing distinct clusters partially matching the underlying preference assignments.}
    \label{appx:fig:cval-heatmap-dst-uneq}
\end{figure}

\textbf{Deep-Sea Treasure.} Sample results for the development of client similarity during training on the Deep-Sea Treasure environment with balanced and unbalanced preference assignment are shown in Figure~\ref{appx:fig:cval-heatmap-dst-eq} and Figure~\ref{appx:fig:cval-heatmap-dst-uneq}, respectively. In both figures, we observe that a grouping of clients becomes visible quite early in the learning process. Though this grouping is not perfect, it does largely correspond to those sets of clients that have been assigned the same preference. The flaws in the grouping process likely spring from an early clustering step, where preference similarities were not fully reflected in the respective model gradients.


\begin{figure}[bh]
    \centering
    \includegraphics[width=.75\columnwidth]{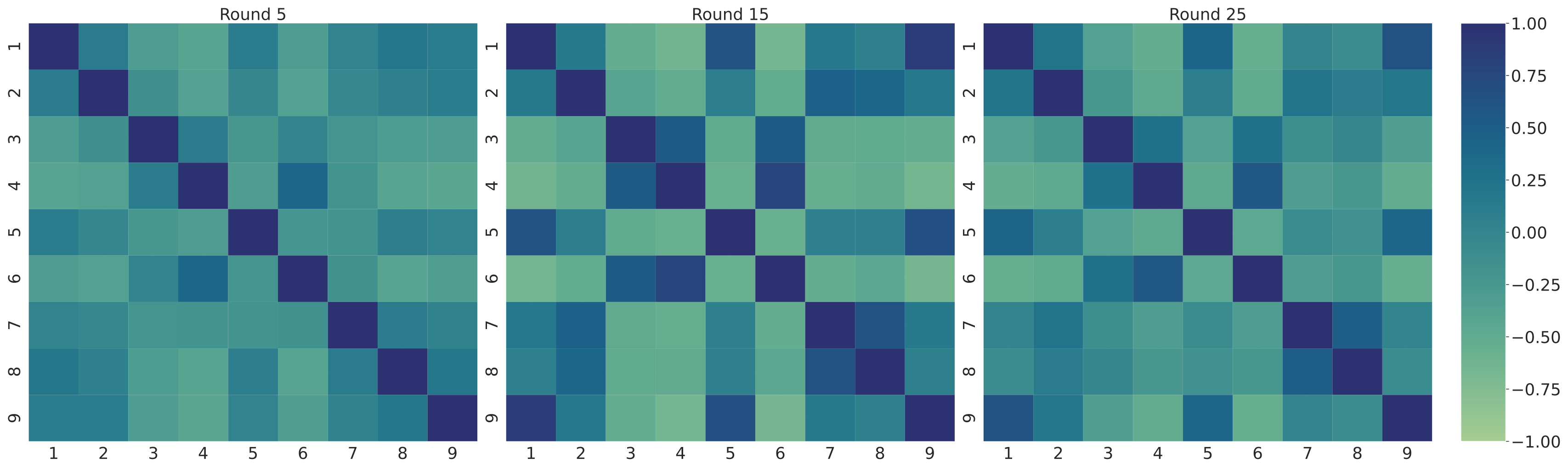}
    \caption{Mutual client similarity at different stages during a single experimental run on the MO-HC environment, with balanced preference distribution. Left to right: client similarities after aggregation round $5, 15$ and $25$ of $30$, respectively.}
    \label{appx:fig:cval-heatmap-hc-eq}
    \Description{Three heatmaps. Some client affinities are visible, but no clear clustering pattern.}
\end{figure}

\begin{figure}[th]
    \centering
    \includegraphics[width=.75\columnwidth]{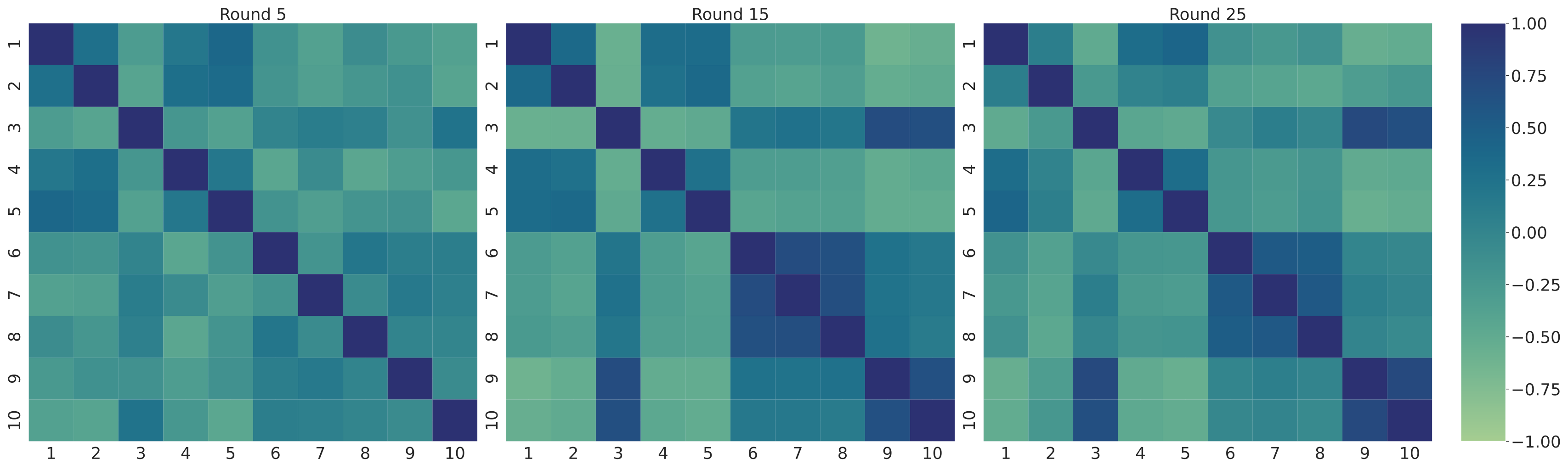}
    \caption{Mutual client similarity at different stages during a single experimental run on the MO-HC environment, with unbalanced preference distribution. Left to right: client similarities after aggregation round $5, 15$ and $25$ of $30$, respectively.}
    \label{appx:fig:cval-heatmap-hc-uneq}
        \Description{Three heatmaps. Some client affinities are visible, but only one clear clustering pattern.}
\end{figure}

\textbf{MO-Halfcheetah.} Uniquely for this environment, experiments were run with a lower number of clients, due to the high computing cost of solving this problem. For the equal distribution, systems with $9$ clients were constructed, with the same preference weights given to $3$ clients each. For the unequal distribution scenario, $10$ clients were run, with $1, 4, 3, $ and $2$ clients receiving the same preferences, respectively. Sample results are shown in Figure~\ref{appx:fig:cval-heatmap-hc-eq} for the balanced distribution, and Figure~\ref{appx:fig:cval-heatmap-hc-uneq} for the unbalanced distribution. For the results of the balanced distribution, we observe a fairly early tendency for dissimilarity between the clients, with some similarity grouping already apparent in the leftmost image, after five aggregation rounds. This early divergence between clients seems to lead in part to counter-intuitive clustering decisions, so that not all clients with the same similarity are grouped together. However, we note that, given the results seen in the main part of the paper, clients do appear to be able to learn together constructively. In the final image, we observe slightly less sharp dissimilarities between clients, indicating that further clustering has taken place, and most clients are likely entirely separated from the rest.\\
For the unbalanced preference distribution, we observe similar results, though interestingly the resulting clusters seem more appropriate to the underlying distribution structure. However, this difference could be a result of the particular experiment instances selected here for visualisation.


\begin{figure}[bh]
    \centering
    \includegraphics[width=.75\columnwidth]{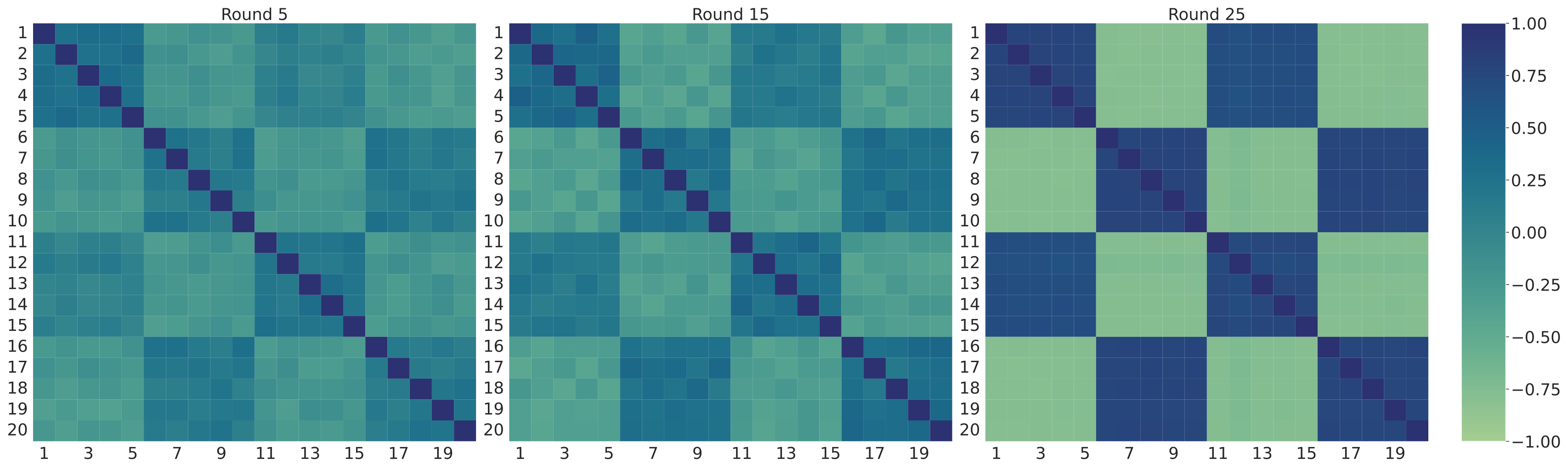}
    \caption{Mutual client similarity at different stages during a single experimental run on the DST environment, with balanced preference distribution. Left to right: client similarities after aggregation round $5, 15$ and $25$ of $30$, respectively.}
    \label{appx:fig:cval-heatmap-llc-eq}
        \Description{Three heatmaps. A clear pattern of four balanced clusters is visible, becoming more clearly delineated as training progresses.}
\end{figure}

\begin{figure}[th]
    \centering
    \includegraphics[width=.75\columnwidth]{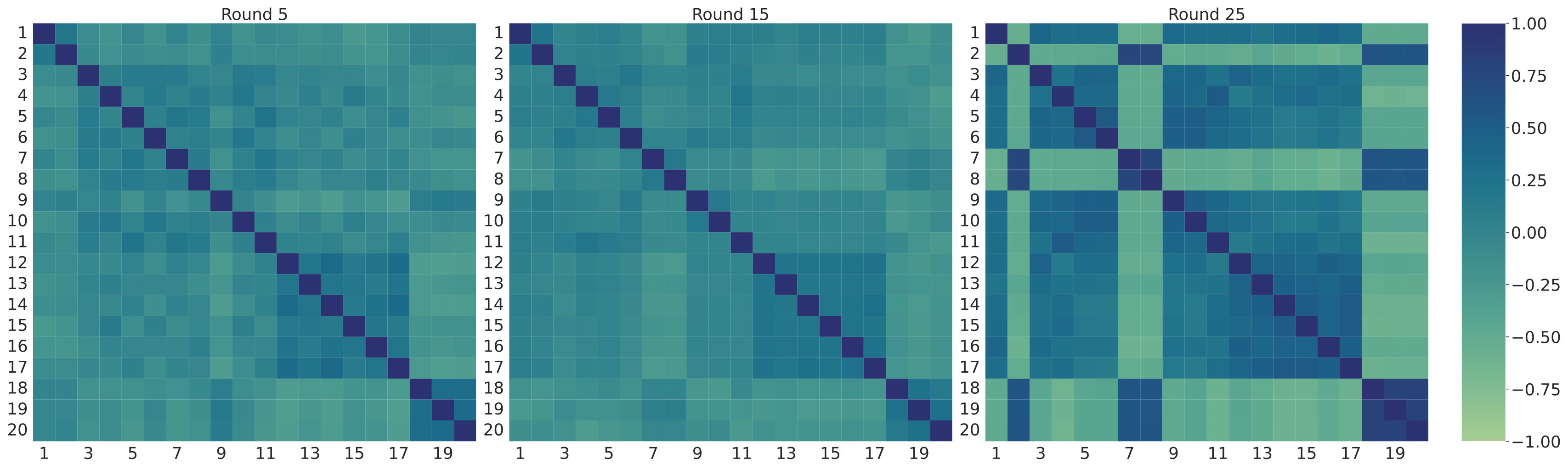}
    \caption{Mutual client similarity at different stages during a single experimental run on the DST environment, with unbalanced preference distribution. Left to right: client similarities after aggregation round $5, 15$ and $25$ of $30$, respectively.}
    \label{appx:fig:cval-heatmap-llc-uneq}
        \Description{Three heatmaps. A pattern emerges slowly, partially matching the underlying preference distribution towards the end of the training process.}
\end{figure}

\textbf{Continuous MO-Lunar Lander.} Figure~\ref{appx:fig:cval-heatmap-llc-eq} and Figure~\ref{appx:fig:cval-heatmap-llc-uneq} show sample similarity results for clients trained on the Continuous MO-Lunar Lander environment under balanced and unbalanced preference distributions, respectively. We observe that these results are quite similar to those of the MO-LL environment, with near-perfect separation of clients into groups with the same preferences. For the visualised instance of the balanced distribution, these groups become visible almost immediately, already clearly recognisable after $5$ aggregation rounds. Indeed, it appears that the clustering process fully separates clients with different preferences almost immediately, with no intermediate step with larger cluster groups discernible.\\
For the visualised instance of training on the unbalanced distribution, it takes markedly longer for clear similarity differences to become visible. This suggests that the clustering process takes longer to separate clients, either because they do indeed benefit from mutual collaboration for an extended time, or perhaps, as speculated earlier, because larger groups of clients with the same preferences dominate the intra-cluster training, skewing the cluster-mean convergence criterion. However, we note that in the final image, after $25$ aggregation rounds, the separation of clients into clusters again matches the underlying distribution structure almost perfectly.


\subsubsection{Investigating CFL clustering}
\begin{figure}[ht]
     \centering
     \begin{subfigure}[b]{0.32\columnwidth}
         \centering
         \includegraphics[width=\columnwidth]{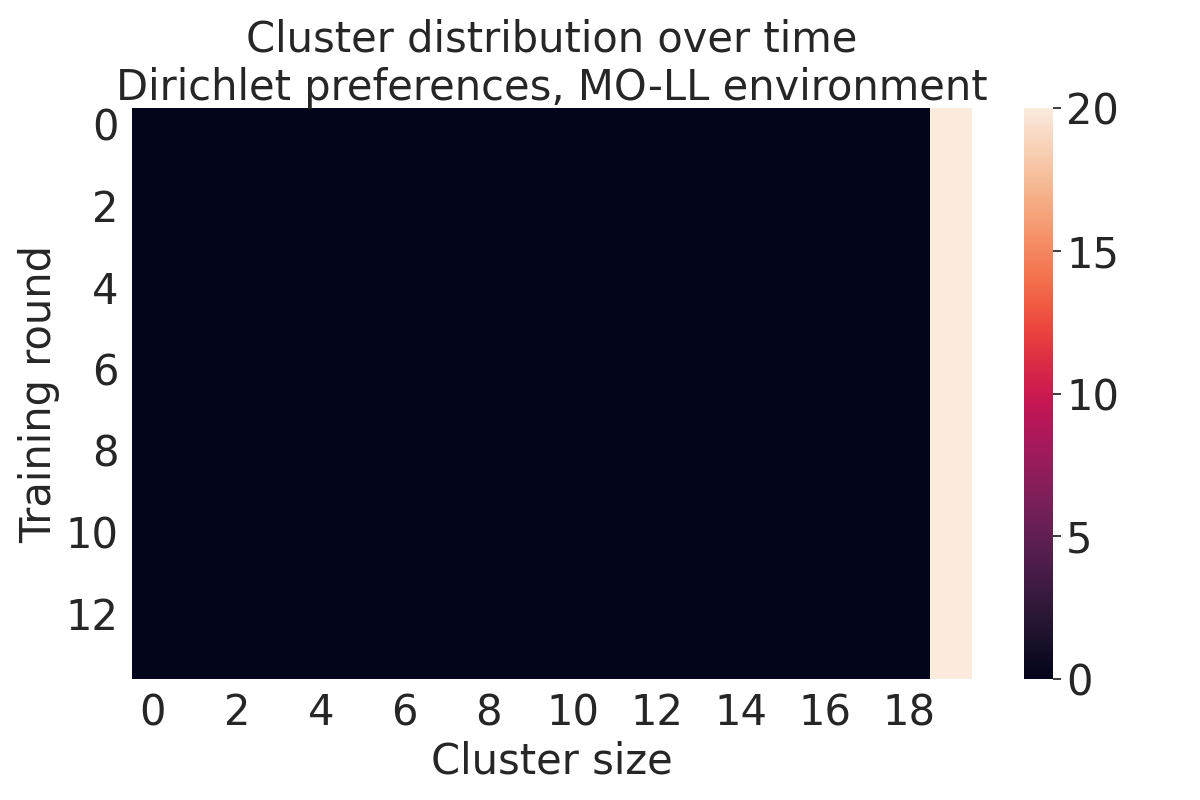}
     \end{subfigure}
     \hfill
     \begin{subfigure}[b]{0.32\columnwidth}
         \centering
         \includegraphics[width=\columnwidth]{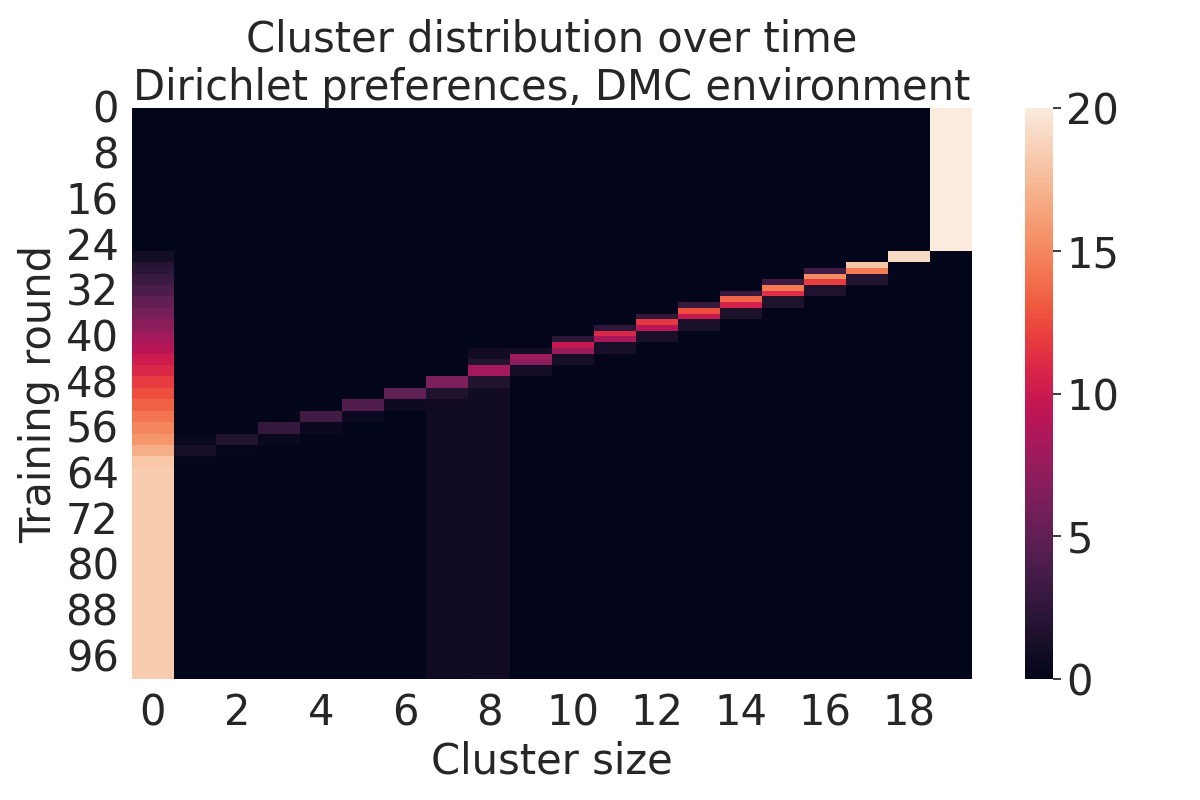}
     \end{subfigure}
     \hfill
     \begin{subfigure}[b]{0.32\columnwidth}
         \centering
         \includegraphics[width=\columnwidth]{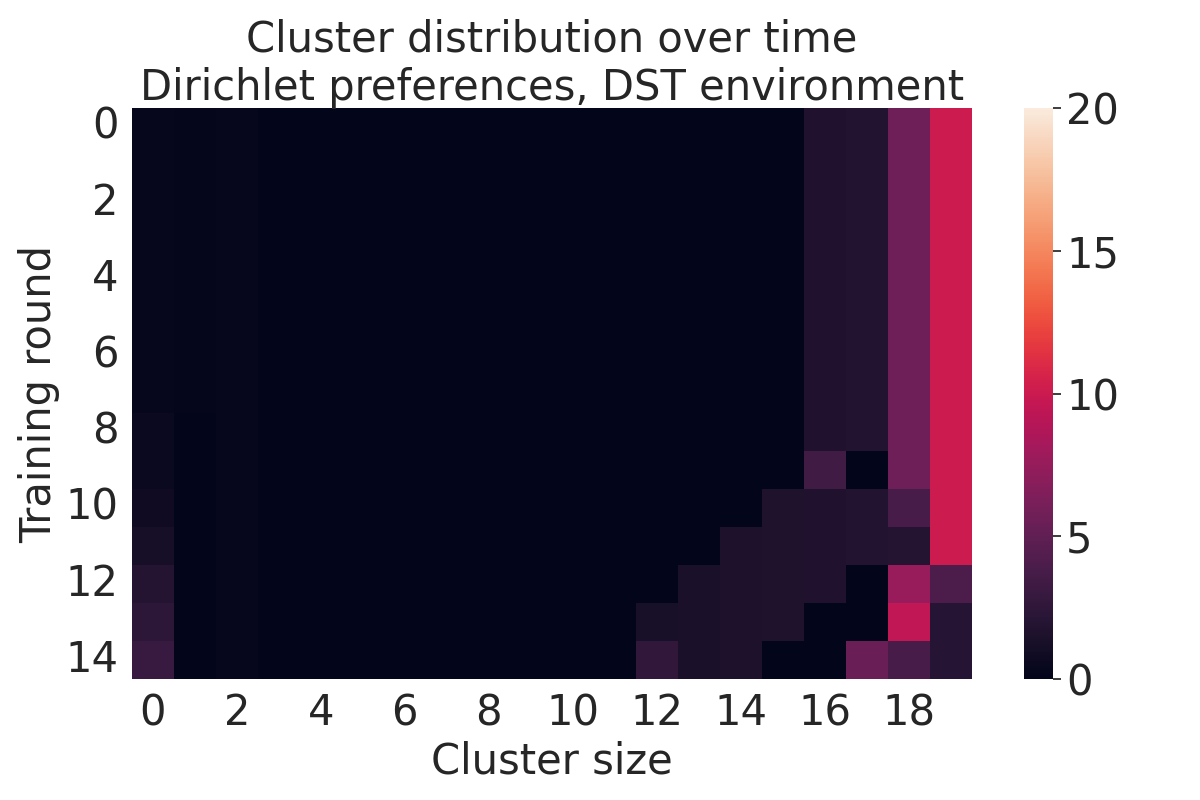}
     \end{subfigure}
        \caption{Clustering behaviour of the CFL algorithm across different environments. Left to right: MO-LL, DMC, DST.}
        \label{appx:fig:cfl-clustering-dqn}
            \Description{Three heatmaps. For the MO-Lunar Lander environment, no clustering occurs. For the Deterministic Minecart, nearly all clustering steps result in a single-client cluster being split off, until only single-client clusters remain. For the Deep-Sea Treasure, clustering occurs rarely, but yields equally imbalanced results.}
\end{figure}

\begin{figure}[ht]
     \centering
     \begin{subfigure}[b]{0.32\columnwidth}
         \centering
         \includegraphics[width=\columnwidth]{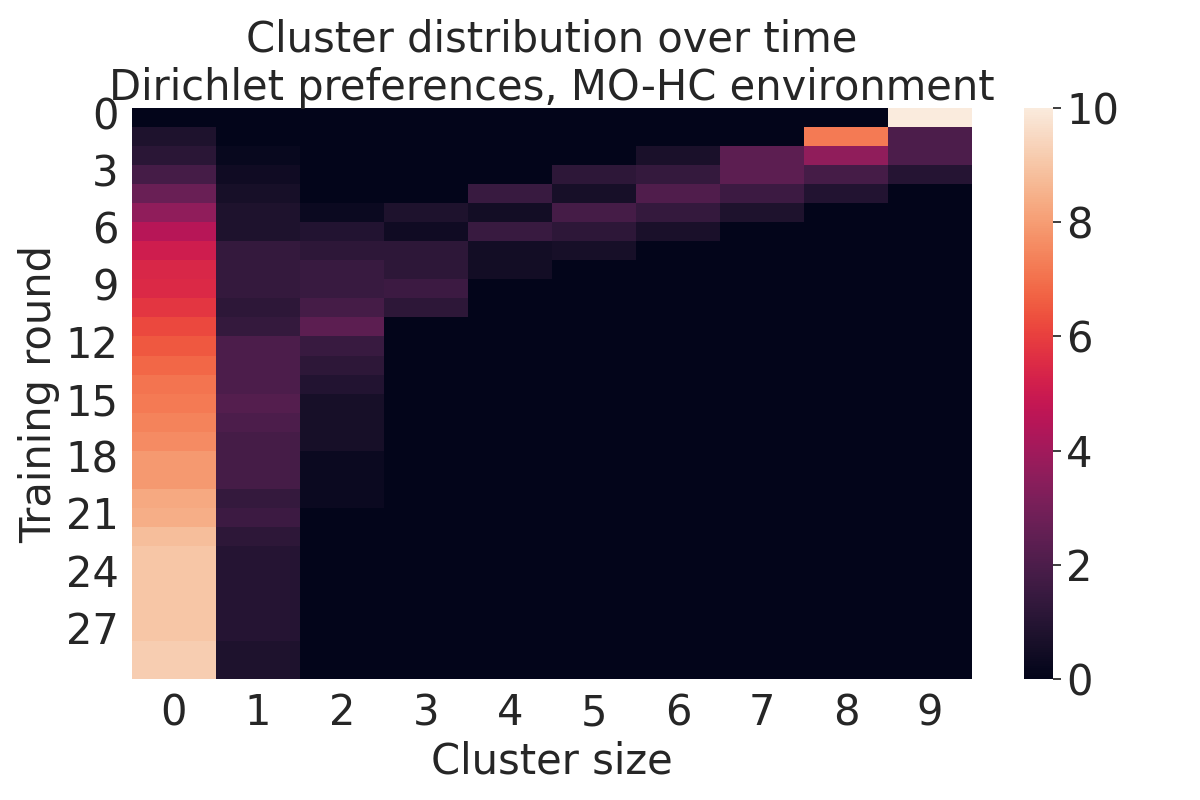}
     \end{subfigure}
     \begin{subfigure}[b]{0.32\columnwidth}
         \centering
         \includegraphics[width=\columnwidth]{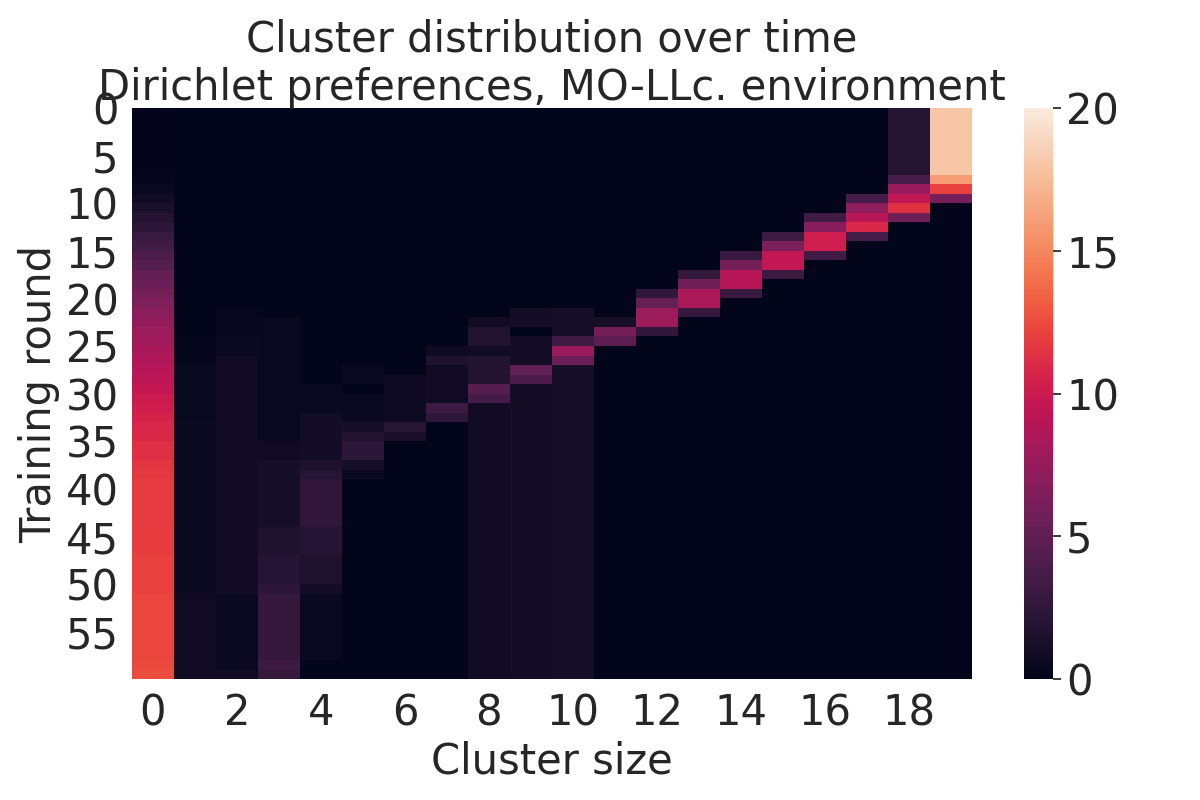}
     \end{subfigure}
        \caption{Clustering behaviour of the CFL algorithm across different environments. Left to right: MO-HC, MO-LLcont.}
        \label{appx:fig:cfl-clustering-ddpg}
            \Description{Two heatmaps. For the MO-Halfcheetah, clustering occurs immediately, producing single-client clusters until no others remain. The MO-Continuous Lunar Lander environment shows similar behaviour.}
\end{figure}

In this subsection, we present a brief exploration of the clustering performance of the CFL algorithm on the experimental problems discussed in this paper. This investigation was sparked by the observation that the CFL algorithm often performed relatively similarly to the non-personalised FedAvg algorithm in our validation experiments. In Figures~\ref{appx:fig:cfl-clustering-dqn} and \ref{appx:fig:cfl-clustering-ddpg}, we trace the evolution in the distribution of cluster sizes across aggregation rounds. In the interest of brevity, we include figures only for preferences generated under a Dirichlet distribution; these are broadly representative of the corresponding results for the other distributions. We make two general observations: \begin{itemize}
    \item For some environments (MO-LL and DST), the clustering process is rarely triggered.
    \item If the clustering process is triggered, it appears to lead to very imbalanced clusters -- often, single-client clusters are created.
\end{itemize}
Given that the clustering threshold parameters for CFL were selected following a hyperparameter search of parameter intervals recommended in \cite{Sattler2019ClusteredFL}, we suspect that the first observation is explained by the second: perhaps the imbalanced clustering occurring here limits the performance of the algorithm sufficiently that the hyperparameter search leads to the selection of a threshold that is rarely triggered, avoiding clustering altogether. To explain the poor clustering performance itself, we propose two hypotheses. First, perhaps the size or training development of the RL models trained in these experiments impacts the success of the similarity metric computed on these models and so impacts the clustering process. Second, the greedy clustering algorithm proposed in \cite{Sattler2019ClusteredFL} may not favour the generation of balanced clusters.
\end{document}